\documentclass[10pt,journal,compsoc]{IEEEtran}
\newif\ifpeerreview

\peerreviewfalse

\usepackage[nocompress]{cite}
\usepackage{url}
\usepackage{amsmath,amssymb,amsthm,mathtools,graphicx}
\usepackage[dvipsnames]{xcolor}
\usepackage{ifthen}
\usepackage{float}
\usepackage{microtype}
\usepackage{subcaption}
\usepackage{booktabs}
\usepackage{multirow}
\usepackage{makecell}
\usepackage{tabularx}
\usepackage{threeparttable}
\usepackage{tablefootnote}
\usepackage{enumitem}
\usepackage{array}
\usepackage{ragged2e}
\usepackage{longtable}
\usepackage{xurl}
\usepackage[T1]{fontenc}
\usepackage[utf8]{inputenc}

\usepackage{listings}
\usepackage{tcolorbox}
\tcbuselibrary{skins, breakable, raster, listings}

\usepackage{tikz}
\usetikzlibrary{shapes.geometric, arrows.meta, positioning, calc}

\usepackage{hyperref}
\usepackage[capitalize,noabbrev]{cleveref}

\usepackage[switch]{lineno}

\usepackage[textsize=tiny]{todonotes}

\DeclareUnicodeCharacter{2082}{$_2$}

\newcolumntype{L}[1]{>{\RaggedRight\arraybackslash}p{#1}}

\newtcolorbox{promptbox}[1]{
    colback=white,
    colframe=black!70!white,
    title=\textbf{#1},
    fonttitle=\bfseries,
    breakable
}

\definecolor{codegreen}{rgb}{0,0.6,0}
\definecolor{codered}{rgb}{0.7,0.1,0.1}
\definecolor{codeblue}{rgb}{0.1,0.1,0.7}
\definecolor{codegray}{rgb}{0.5,0.5,0.5}
\definecolor{backcolour}{rgb}{0.97,0.97,0.97}

\definecolor{rawbg}{rgb}{1.0, 0.96, 0.96}
\definecolor{stdbg}{rgb}{0.96, 1.0, 0.96}
\definecolor{logbg}{rgb}{0.15, 0.15, 0.15}
\definecolor{docbg}{rgb}{1.0, 1.0, 0.92}

\lstdefinestyle{pythonstyle}{
    backgroundcolor=\color{backcolour},
    commentstyle=\color{codegreen}\itshape,
    keywordstyle=\color{codeblue}\bfseries,
    numberstyle=\tiny\color{codegray},
    stringstyle=\color{codered},
    basicstyle=\ttfamily\scriptsize,
    breakatwhitespace=false,
    breaklines=true,
    captionpos=b,
    keepspaces=true,
    numbers=left,
    numbersep=5pt,
    showspaces=false,
    showstringspaces=false,
    showtabs=false,
    tabsize=2
}

\lstdefinestyle{logstyle}{
    backgroundcolor=\color{logbg},
    basicstyle=\ttfamily\scriptsize\color{white},
    keywordstyle=\color{white},
    commentstyle=\color{white},
    stringstyle=\color{white},
    identifierstyle=\color{white},
    breaklines=true,
    numbers=none,
    frame=none,
    showstringspaces=false
}

\lstdefinelanguage{json}{
    basicstyle=\ttfamily\scriptsize,
    numbers=left,
    numberstyle=\tiny\color{codegray},
    stepnumber=1,
    numbersep=8pt,
    showstringspaces=false,
    breaklines=true,
    frame=lines,
    string=[s]{"}{"},
    comment=[l]{:\ "},
    morecomment=[l]{:"},
    literate=
     *{0}{{{\color{blue}0}}}{1}
      {1}{{{\color{blue}1}}}{1}
      {2}{{{\color{blue}2}}}{1}
      {3}{{{\color{blue}3}}}{1}
      {4}{{{\color{blue}4}}}{1}
      {5}{{{\color{blue}5}}}{1}
      {6}{{{\color{blue}6}}}{1}
      {7}{{{\color{blue}7}}}{1}
      {8}{{{\color{blue}8}}}{1}
      {9}{{{\color{blue}9}}}{1}
      {:}{{{\color{red}{:}}}}{1}
      {,}{{{\color{red}{,}}}}{1}
      {\{}{{{\color{blue}{\{}}}}{1}
      {\}}{{{\color{blue}{\}}}}}{1}
      {[}{{{\color{blue}{[}}}}{1}
      {]}{{{\color{blue}{]}}}}{1},
}

\definecolor{figcodekeyword}{HTML}{A020F0}
\definecolor{figcodecomment}{HTML}{008000}
\definecolor{figcodestring}{HTML}{0000FF}
\definecolor{figcodeblack}{HTML}{000000}

\lstdefinestyle{figmystyle}{
    language=Python,
    basicstyle=\ttfamily\scriptsize\color{figcodeblack},
    commentstyle=\color{figcodecomment},
    keywordstyle=\color{figcodekeyword}\bfseries,
    stringstyle=\color{figcodestring},
    numberstyle=\tiny\color{codegray},
    breakatwhitespace=false,
    breaklines=true,
    captionpos=b,
    keepspaces=true,
    showspaces=false,
    showstringspaces=false,
    showtabs=false,
    tabsize=4,
    frame=none,
    columns=fullflexible,
    upquote=true,
    morekeywords={as},
    backgroundcolor=\color{white}
}

\tcbset{mainbox/.style={
    enhanced,
    colframe=black,
    colback=white,
    boxrule=0.8pt,
    arc=0pt,
    outer arc=0pt,
    left=4pt, right=4pt, top=4pt, bottom=4pt,
    boxsep=0pt,
    fonttitle=\bfseries\small,
    fontupper=\small
}}

\tcbset{purposebox/.style={
    enhanced,
    colframe=black,
    colback=white,
    boxrule=0.8pt,
    arc=0pt,
    outer arc=0pt,
    left=4pt, right=4pt, top=2pt, bottom=2pt,
    boxsep=0pt,
    fontupper=\bfseries\small,
    nobeforeafter,
    width=\linewidth
}}

\newcommand{\e}{\begin{equation}}
\newcommand{\ee}{\end{equation}}
\newcommand{\en}{\begin{equation*}}
\newcommand{\een}{\end{equation*}}
\newcommand{\eqn}{\begin{eqnarray}}
\newcommand{\eeqn}{\end{eqnarray}}
\newcommand{\bmat}{\begin{bmatrix}}
\newcommand{\emat}{\end{bmatrix}}

\DeclareMathAlphabet\mathbfcal{OMS}{cmsy}{b}{n}

\newcommand{\vct}[1]{\boldsymbol{#1}}

\newcommand{\calA}{\mathcal{A}}

\newcommand{\calX}{\mathcal{X}}
\newcommand{\calY}{\mathcal{Y}}

\newcommand{\vx}{\vct{x}}
\newcommand{\vy}{\vct{y}}

\newcommand{\veta}{\vct{\eta}}

\graphicspath{{./figs/}}

\newlength{\imgwidth}
\newboolean{twoColVersion}
\setboolean{twoColVersion}{false}
\newcommand{\twoCol}[2]{\ifthenelse{\boolean{twoColVersion}} {#1} {#2} }

\NewDocumentCommand{\yangling}
{ mO{} }{\textcolor{red}{YL: \textsf{\textbf{\small#1}}}}

\newcommand{\crfix}[1]{#1}

\usepackage{etoolbox}
\makeatletter
\patchcmd{\IEEEbiographynophoto}%
  {\vskip 4\baselineskip plus 1fil minus 0\baselineskip}%
  {\vskip 1\baselineskip plus .2\baselineskip minus .2\baselineskip}%
  {}{\typeout{** CR-FIX WARNING: could not patch IEEEbiographynophoto spacing.}}
\makeatother

\theoremstyle{plain}

\theoremstyle{definition}

\theoremstyle{remark}

\newcommand{\paperID}{22}

\title{Imaging-101: Benchmarking LLM \crfix{Coding} Agents \crfix{on} Scientific Computational Imaging}

\author{Siyi~Chen$^{*}$, Jiahe~Ying$^{*}$, Yixuan~Jia$^{*}$, Yuxuan~Gu$^{*}$, Enze~Ye, Weimin~Bai,
Zhijun~Zeng, Shaochi~Ren, Binhong~Gao, Yubing~Li,~\IEEEmembership{Member,~IEEE,} Tianhan~Zhang,
and He~Sun$^{\dagger}$,~\IEEEmembership{Member,~IEEE}%
\IEEEcompsocitemizethanks{
\IEEEcompsocthanksitem This work was supported by the Shanghai Municipal Science and
Technology Major Project (2025SHZDZX026D03), the Natural Science Foundation of Beijing Municipality (Z240010) and the National Natural Science Foundation
of China (32450631 and 62371007).
\IEEEcompsocthanksitem $^{\dagger}$~Corresponding author: He Sun. E-mail: hesun@pku.edu.cn.
\IEEEcompsocthanksitem $^{*}$~Siyi Chen, Jiahe Ying, Yixuan Jia, and Yuxuan Gu contributed equally to this work.
\IEEEcompsocthanksitem Siyi Chen, Jiahe Ying, Yuxuan Gu, Enze Ye, Weimin Bai, Zhijun Zeng,
Shaochi Ren, Binhong Gao, and He Sun are with the College of Future Technology
and the National Biomedical Imaging Center, Peking University, Beijing, 100871, China.
\IEEEcompsocthanksitem Jiahe Ying is also with the AI for Science Institute (AISI), Beijing, 100080, China.
\IEEEcompsocthanksitem Yixuan Jia is with the University of Michigan, Ann Arbor, MI 48105, United States.
\IEEEcompsocthanksitem Yubing Li is with the State Key Laboratory of Acoustics and Marine Information,
Institute of Acoustics, Chinese Academy of Sciences, Beijing, 100190, China, and also with
the University of Chinese Academy of Sciences, Beijing, 100049, China.
\IEEEcompsocthanksitem Tianhan Zhang is with the School of Astronautics, Beihang University, Beijing, 100191, China. Tianhan Zhang is also with the Key Laboratory of Spacecraft Design Optimization and Dynamic Simulation Technologies, Ministry of Education, Beijing, 102206, China.}%
}

\begin{document}

\IEEEtitleabstractindextext{%
\begin{abstract}
Computational imaging, which recovers hidden signals from indirect, noisy
measurements, underpins quantitative discovery across scientific disciplines,
yet building a correct reconstruction pipeline demands deep domain expertise
and remains laborious even for domain scientists.
We introduce \textbf{Imaging-101}, a benchmark of 57 expert-verified
computational imaging tasks spanning six scientific domains, each grounded in
a peer-reviewed paper and canonicalized into a standardized four-stage pipeline
(preprocessing, forward physics modeling, inverse solver, and visualization).
Three evaluation tracks (planning, function-level unit tests, and end-to-end
reconstruction) probe distinct agent capabilities across the full pipeline.
Evaluating seven frontier LLMs uncovers systematic challenges in applying
coding agents to computational imaging that go beyond those exposed by general
coding benchmarks, spanning algorithm selection, physical convention handling,
and pipeline integration.
These findings highlight concrete capability gaps and point toward
skill-augmented, domain-specialized agents as a practical path to reliable
computational imaging \crfix{assistance}.
\end{abstract}

}

\maketitle

\IEEEraisesectionheading{\section{Introduction}\label{sec:introduction}}

\begin{figure*}[!t]
\begin{center}
    \includegraphics[width=\textwidth]{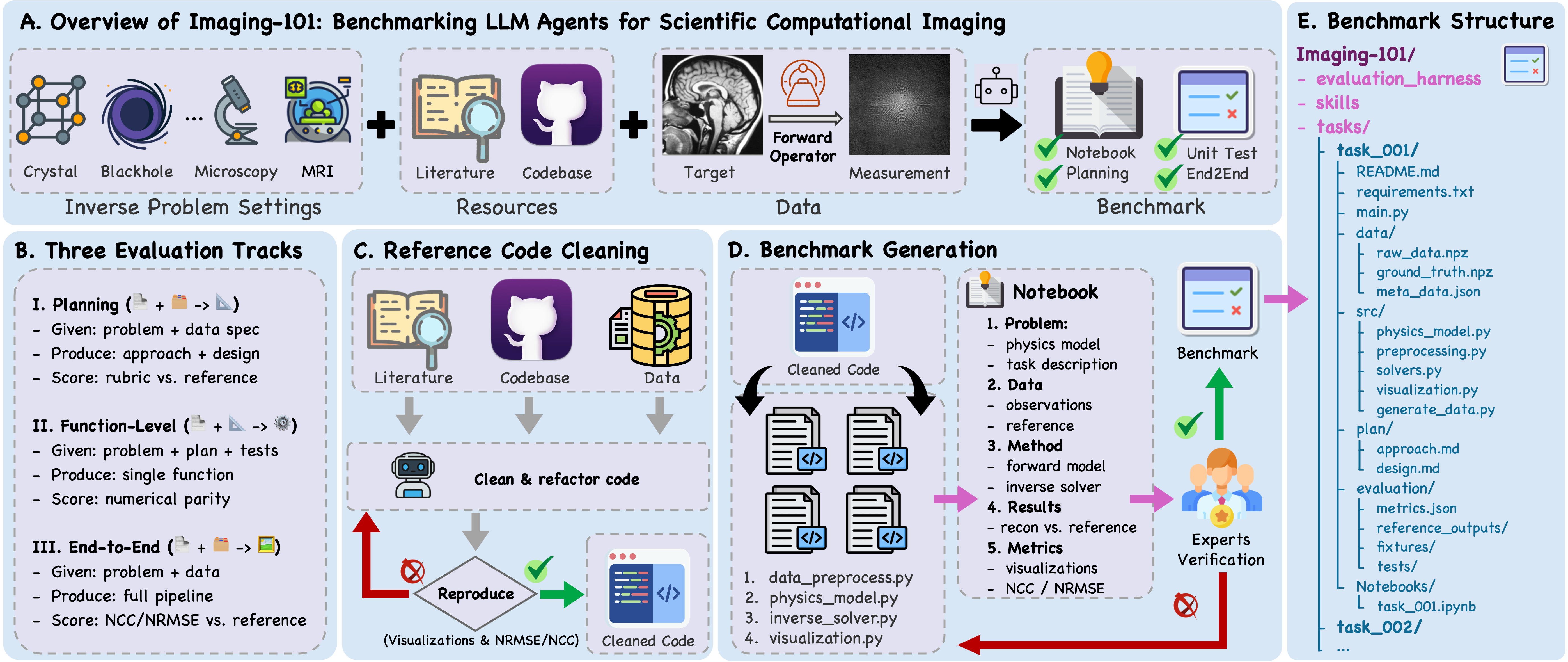}
\end{center}
\caption{\textbf{Overview of Imaging-101.}
\textbf{(A)} Imaging-101 benchmarks LLM agents on scientific
computational imaging tasks across astronomy, biology, chemistry \&
material science, earth science, medicine, and physics.
\textbf{(B)} Each task supports three evaluation tracks:
\emph{Planning} (approach/design vs.\ reference plan),
\emph{Function-level} (per-function numerical unit tests), and
\emph{End-to-end} (full pipeline scored by image quality metrics).
\textbf{(C)} An expert-supervised pipeline (with Claude Code
assistance) reproduces and refactors each reference codebase;
domain experts verify reproduction against the original publication
before the task is accepted.
\textbf{(D)} Accepted code is canonicalized into a standardized
four-stage computational imaging pipeline (preprocessing, forward
physics modeling, inverse solver, and visualization), producing a
reference implementation verified at every stage.
\textbf{(E)} Each released task provides data, reference source,
plan documents, execution source code, evaluation scripts, and unit tests, directly
supporting the three evaluation tracks (see \Cref{app:task_structure} for the full directory layout). An executed Jupyter notebook
accompanies each task, serving as a self-contained tutorial that walks
through the full pipeline with precomputed results for interactive exploration.
}
\vspace{-1em}
\label{fig:main framework of our work}
\end{figure*}

Science begins where measurement begins.
Modern science relies fundamentally on \emph{computational imaging},
the recovery of hidden signals from indirect, noisy, and incomplete
measurements, to infer the unknown states of physical, chemical,
and biological systems~\cite{stuart2010inverse,arridge2019solving}.
Mathematically, each such task is an \emph{inverse problem}: while a
forward model predicts observations from known parameters, the imaging
pipeline operates in the opposite direction, recovering latent
quantities that cannot be measured directly.
This framework underpins quantitative discovery across
scientific disciplines~\cite{king2009automation,stokes2020deep,jumper2021highly,zhou2024hypothesis}, from reconstructing
black hole images~\cite{event2023first}, mapping subsurface
geology~\cite{virieux2009overview} to recovering anatomical structure in
CT~\cite{yang2025local,radon2007determination,shepp2007maximum} and MRI~\cite{lustig2007sparse,fessler2019optimization,lobos2025spatiotemporal}.

Despite their ubiquity, solving a single computational imaging task in practice is laborious~\cite{werling2023addbiomechanics,musslick2025automating}.
Real measurements are noisy, sparse, and often poorly conditioned,
leading to severe ill-posedness~\cite{kaipio2005statistical,tarantola2005inverse}.
Building a correct pipeline requires data cleaning,
forward-model implementation, algorithm selection, code development, and iterative debugging, work that is time-consuming and demands deep domain expertise.
Recent advances in LLMs and agentic systems raise the prospect of
sharing this burden with AI collaborators~\cite{bran2023chemcrow,shen2023hugginggpt}.
LLMs have shown strong performance on parts of the scientific
workflow, including literature retrieval, mathematical reasoning, and
code generation.
However, solving a computational imaging task end-to-end requires
more: the model must jointly reason over physical measurement
processes, mathematical formulations, and numerical
solvers, and translate all into executable code that passes
strict numerical verification.
Whether frontier models can do this reliably remains an open question.

\textbf{Imaging-101.}
To address this gap, we introduce Imaging-101, a benchmark of 57
computational imaging tasks spanning astronomy, biology, chemistry \&
material science, earth science, medicine, and physics.
Every task is grounded in a peer-reviewed paper and its open-source
reference implementation, and graded by executing the agent's code
against calibrated numerical tolerances.
The benchmark is built through an \emph{expert-in-the-loop} pipeline
in which Claude Code accelerates reproduction, refactoring, and
unit-test synthesis, while domain experts retain responsibility for
validating correctness against the original publications.
A central design principle is that every task is canonicalized into a
\emph{standardized four-stage pipeline} (preprocessing, forward
physics modeling, inverse solver, and visualization), so agents can
reason about any task through the same interface.
Imaging-101 supports three complementary evaluation tracks that probe
distinct capabilities: \emph{planning}, \emph{function-level} unit
tests, and \emph{end-to-end} reconstruction.
Full details of the task selection, construction pipeline, and
evaluation protocols are given in~\Cref{sec:overview}.

\textbf{Findings.}
Evaluating seven frontier LLM agents on Imaging-101 reveals systematic
failure modes that are rarely exposed by QA-style science benchmarks
or generic coding benchmarks.
The dominant errors are not surface-level coding mistakes but
scientific reasoning failures, including but not limited to missing domain-specific physical
knowledge (e.g., omitting a required unit conversion between sensor
measurements and algorithm inputs), selecting an inappropriate
solver, and failing to interpret numerical
residuals as signatures of a missing normalization or wrong
convention.
These findings are detailed in~\Cref{sec:results}.

\textbf{Contributions.}
In summary, this work:
\begin{itemize}
    \item introduces \textbf{Imaging-101}, a benchmark of 57
    expert-verified, paper-grounded computational imaging tasks
    supporting planning, function-level, and end-to-end evaluation
    across six scientific domains;
    \item develops an expert-in-the-loop construction pipeline that
    canonicalizes research papers into a standardized four-stage
    computational imaging layout, yielding grounded, reproducible,
    execution-verifiable benchmark tasks and enabling scalable
    community contributions to the benchmark;
    \item provides empirical evidence of current LLM capability gaps
    on scientific computational imaging, characterizing the dominant
    failure modes and laying groundwork for developing domain-specific
    coding agents and research copilots for computational imaging.
\end{itemize}

\vspace{-1.5em}
\section{Overview of Imaging-101}
\label{sec:overview}

Imaging-101 benchmarks whether a large language model can turn a
real experimental description into a numerically correct reconstruction
pipeline.
Every task is grounded in a peer-reviewed paper and its open-source
reference implementation, and graded by executing the agent's code
against calibrated numerical tolerances.
A central design choice is to canonicalize every task into a
\emph{standardized four-stage computational imaging pipeline}
(preprocessing, forward physics modeling, inverse solver, and
visualization) so that an agent can reason about any task through the
same interface.
This section formalizes the class of problems the benchmark targets
(\Cref{sec:problem_scope}), describes the expert-supervised
construction pipeline (\Cref{sec:auto_benchmark_pipeline}), and
specifies the evaluation protocols (\Cref{sec:eval_protocol}).

\subsection{Formulation of Computational Imaging Tasks}
\label{sec:formulation}

A computational imaging task is an \emph{inverse problem}: recover a
latent quantity $\vx \in \calX$ from indirect measurements
$\vy \in \calY$ related through a forward operator $\calA$ and noise
$\veta$:
\begin{equation}
\label{eq:inverse_problem}
    \vy \;=\; \calA(\vx) \;+\; \veta,
\end{equation}
where $\calA\colon \calX \to \calY$ encodes the physics of the
measurement process (e.g., the Radon transform in CT, Fourier
under-sampling in MRI, or wave propagation in seismology) and
$\veta$ captures instrument noise and model mismatch.
Given $\vy$ and a description of $\calA$, the goal is to construct
an estimate $\hat{\vx}$ of the true signal.

Solving~\eqref{eq:inverse_problem} in practice requires
(i)~preprocessing raw measurements into a form compatible with
$\calA$, (ii)~implementing $\calA$ (and often its adjoint
$\calA^{*}$) as executable code, (iii)~choosing and running an
inversion algorithm, and (iv)~visualizing $\hat{\vx}$ with appropriate image display tools.
In Imaging-101, each task exposes exactly this pipeline: the agent
receives a natural-language description of the experimental setup,
the raw measurement data $\vy$, and task metadata, and must produce
executable code that carries out steps~(i)--(iv) and passes
numerical verification against a reference implementation.

\crfix{\textbf{Scope and intended use.}
Imaging-101 evaluates whether an LLM coding agent can turn an
\emph{expert-curated} computational-imaging problem, comprising an
experimental description, the measurement data, and a standardized task
specification, into a numerically correct reconstruction pipeline.
It deliberately does \emph{not} measure (i)~human--AI collaborative
reproduction, (ii)~fully blind paper-to-code reproduction from a raw
publication, or (iii)~general scientific-discovery ability.
Each task is formulated as a well-posed problem, with the experimental description, measurement data and standardized specification provided. Because the task already contains the necessary domain context, the agent does not need to mine the literature for external knowledge, allowing the evaluation to isolate its scientific-coding ability.}

\vspace{-0.5em}
\subsection{Problem Scope and Task Selection}
\label{sec:problem_scope}

\begin{figure}[ht]
    \centering
    \includegraphics[width=\linewidth]{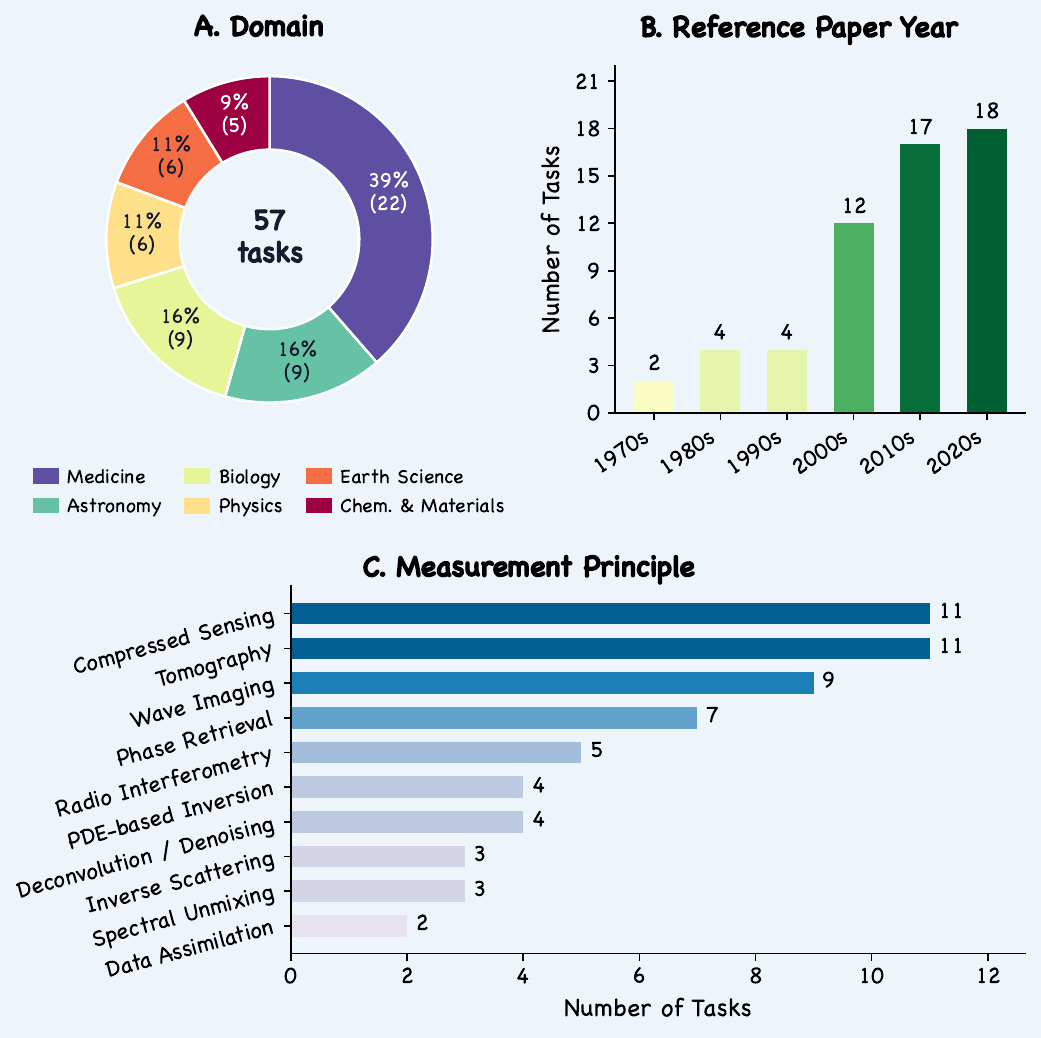}
    \caption{%
      \textbf{Imaging-101 task composition.}
      \textbf{(A)} Distribution of tasks across the six scientific
      domains.
      \textbf{(B)} Publication year of the reference paper for each
      task, showing coverage of work from 1970s to 2020s across multiple decades.
      \textbf{(C)} Distribution by fundamental measurement principle;
      a single task may involve multiple principles.
      }
    \label{fig:task_composition}
\end{figure}


Imaging-101 contains \textbf{57 tasks} spanning six scientific
domains: astronomy, biology, chemistry \& material science, earth
science, medicine, and physics.
Task selection was guided by domain experts from each field, with two
complementary objectives.

\textbf{Domain coverage.}
We include representative imaging modalities from each domain, such as
radio interferometry and exoplanet imaging in astronomy; fluorescence
microscopy, optical diffraction tomography, and Raman spectroscopy
in biology; electron ptychography and hyperspectral unmixing in
chemistry; seismic inversion and InSAR phase unwrapping in earth
science; CT, MRI, PET, ultrasound, and photoacoustic imaging in
medicine; and lensless imaging, ptychography, and deflectometry in
physics.

\textbf{Measurement principle coverage.}
Beyond domain breadth, we deliberately cover a diverse set of
fundamental measurement principles that recur across scientific
imaging: \emph{tomography} (CT, PET, seismic travel-time, optical
diffraction); \emph{phase retrieval} (ptychography, Fourier
ptychography, lensless imaging); \emph{compressed sensing} (sparse-view
CT, accelerated MRI, spectral snapshot imaging); \emph{PDE-based
inversion} (full waveform inversion, photoacoustic, ultrasound);
\emph{inverse scattering} (ODT, seismic RTM); \emph{wave imaging}
(NLOS, plane-wave ultrasound, seismic migration); \emph{spectral
unmixing} (hyperspectral, CARS, Raman); and \emph{data assimilation}
(weather radar, atmospheric reanalysis).
This ensures that benchmark performance reflects general
scientific problem-solving capability rather than proficiency in
a single technique.
Figure~\ref{fig:task_composition} shows the distribution of tasks
across domains and measurement principles.

Tasks were selected to emphasize the characteristics that make
computational imaging challenging for automated solvers.
Each task involves a \emph{long-horizon pipeline} with multiple
coupled stages, where an error in preprocessing propagates silently
through forward modeling into the final reconstruction.
All tasks are grounded in peer-reviewed publications and verified
open-source implementations, ensuring scientific authenticity.

\vspace{-1em}
\subsection{Benchmark Construction Pipeline}
\label{sec:auto_benchmark_pipeline}

Faithful benchmarks for scientific agents require PhD-level labor to
reproduce research code, restructure pipelines, and design reliable
verification~\cite{tian2024scicode,chen2024scienceagentbench,sun2025scienceboard}.
Imaging-101 does not attempt to remove this human effort; it is
essential to make the benchmark scientifically grounded and
execution-verifiable.
To construct a benchmark spanning diverse disciplines efficiently,
we develop an \emph{expert-in-the-loop} pipeline in which domain
experts retain final authority over correctness, while Claude
Code accelerates repetitive engineering tasks: reproducing reference
runs, refactoring code into the standardized layout, synthesizing
unit tests from captured I/O, and drafting tutorials.
Experts verify reproduction against the original publication, inspect
the canonicalized code, review the plan and tutorial, and accept, revise, or
reject each task.
\crfix{The benchmark was built and verified by 12 domain experts (two per
scientific area), including nine senior PhD students and three researchers with PhD degrees. Each task was first constructed by one expert and then
independently verified by a second expert from the same area; when the two
disagreed, they resolved it through discussion and revised the task until they
reached agreement.}
Figure~\ref{fig:main framework of our work}(C)--(E) summarizes the
pipeline; the five stages are as follows.

\textbf{Stage 1. Reproduction of the reference pipeline.}
The coding agent creates an isolated Python environment and iteratively
debugs the original repository until it reproduces one exact
visualization or numerical value reported in the paper or the
upstream code.
This establishes a verified ``ground-truth run'' from which every downstream product is derived.

\textbf{Stage 2. Canonicalization into the Imaging-101 task
template.}
The verified code is refactored into a fixed per-task layout
containing \texttt{main.py}, a \texttt{data/} directory, a modular
\texttt{src/} split along the scientific taxonomy
(\texttt{preprocessing.py}, \texttt{physics\_model.py},
\texttt{solvers.py}, \texttt{visualization.py}), a \texttt{plan/}
directory, an \texttt{evaluation/} directory, and a tutorial
notebook.
Heterogeneous task data are stored as \texttt{data/raw\_data.npz} arrays with
documented key tables, and \texttt{data/meta\_data.json} is restricted to
imaging parameters; solver hyperparameters are excluded to prevent
information leakage to evaluation agents.

\textbf{Stage 3. Reference-output generation and metric boundary
calibration.}
The canonicalized \texttt{main.py} is executed end-to-end and all
intermediate products are written to
\texttt{evaluation/reference\_outputs/}.
Task-level pass criteria are calibrated against baseline
reconstruction quality using normalized cross-correlation (NCC) and
normalized root mean square error (NRMSE), which are
robust to global amplitude scaling and interpretable across imaging
modalities (unlike SSIM or PSNR, which are sensitive to local
image statistics and may not reflect physically meaningful
reconstruction accuracy), and stored in
\texttt{evaluation/metrics.json}.
Specifically, thresholds are set at $0.9\times$ reference NCC and
$1.1\times$ reference NRMSE.
Tasks whose quality is not captured by NCC/NRMSE declare additional
metric fields that the harness discovers automatically.

\textbf{Stage 4. Subtask decomposition and unit-test synthesis.}
Every function in \texttt{src/} receives a \texttt{pytest} in
\texttt{evaluation/tests/} and a fixture in
\texttt{evaluation/fixtures/} captured by instrumenting the reference
run.
Deterministic operators are verified by exact numerical comparison
($\texttt{rtol}\in [10^{-10}, 10^{-2}]$ depending on the scientific task); stochastic components are verified by
statistical properties (shape, dtype, moments), avoiding false
negatives from random-seed replay.
The task is blocked from the benchmark until the full
\texttt{pytest} suite passes against the reference.

\textbf{Stage 5. Tutorial and plan generation.}
The coding agent produces three human-reviewable
products: a \texttt{README.md} with the task description and a mandatory data-key table; a
structured plan comprising \texttt{plan/approach.md} (methodology and
algorithm) and \texttt{plan/design.md} (module layout and data flow);
and an executed tutorial notebook that loads precomputed results and
runs in seconds.
A domain expert reviews the tutorial and plan for completeness,
inspects and runs the source code to verify correctness, and accepts
the task only if all artifacts are consistent with the original
publication.

\subsection{Benchmark Specification and Evaluation Protocol}
\label{sec:eval_protocol}

\subsubsection{Task Instance Structure}
\label{sec:eval_task_instance}
Each accepted task instance contains the following artifacts, which
directly support the three evaluation tracks:

\begin{itemize}
  \item \textbf{\texttt{README.md}}: a structured natural-language
  description of the inverse problem, including background physics, a
  data-key table for all \texttt{.npz} files, and a Method Hints
  section.
  This is the primary input for the \emph{planning} track and the
  \emph{end-to-end} track.

  \item \textbf{\texttt{data/}}: contains \texttt{raw\_data.npz}
  (observations and instrument parameters), \texttt{ground\_truth.npz}
  or \texttt{baseline\_reference.npz} (reference signal for scoring),
  and \texttt{meta\_data.json} (imaging parameters, no solver
  settings).

  \item \textbf{\texttt{plan/approach.md} and
  \texttt{plan/design.md}}: the reference algorithm description and
  module architecture, used as the ground-truth plan in the
  \emph{planning} track and as the input specification in the
  \emph{function-level} track.

  \item \textbf{\texttt{evaluation/tests/} and
  \texttt{evaluation/fixtures/}}: per-function \texttt{pytest} suites
  and captured I/O fixtures, used to grade the \emph{function-level}
  track.

  \item \textbf{\texttt{evaluation/metrics.json}}: the pass/fail
  thresholds (e.g., NCC and NRMSE boundaries) used by all
  execution-based tracks.
\end{itemize}

Imaging-101 supports three complementary evaluation protocols:
\emph{planning} (\Cref{sec:eval_planning}), \emph{function-level}
correctness (\Cref{sec:eval_function}), and \emph{end-to-end}
execution (\Cref{sec:eval_e2e_multiagent}).

\subsubsection{Planning Evaluation}
\label{sec:eval_planning}

In the planning track, the model receives \texttt{README.md} and
\texttt{data/meta\_data.json} and must produce two documents:
\texttt{plan/approach.md} (algorithm description and mathematical
formulation) and \texttt{plan/design.md} (module structure and
function signatures).
Because plan correctness cannot be verified by automated tests, we use a
two-stage human--LLM evaluation pipeline.
First, an LLM judge compares each generated plan against the reference
\texttt{plan/approach.md} and produces a structured difference report
that identifies agreements and discrepancies on three dimensions:
preprocessing, forward physics modeling, and
inverse solver.
Second, domain experts review each report and assign a binary pass/fail
judgment per dimension.
\crfix{Each plan is judged independently by two experts from the same
domain pool used during construction, with model identity blinded; a dimension
is marked correct only when both experts agree, and disagreements are counted as
failures. The full rubric is given in
Appendix~\ref{app:planning_eval}.}
A plan is marked correct if the proposed approach is functionally
equivalent to or a reasonable alternative to the reference, one that an
expert would accept as capable of yielding a correct reconstruction.
Typical errors that trigger a fail judgment are: physically incorrect
model formulas, omission of critical pipeline steps, and
algorithm choices that are fundamentally mismatched to the structure of
the inverse problem.

\subsubsection{Function-Level Evaluation}
\label{sec:eval_function}

In the function-level track, the model receives \texttt{README.md},
the reference plan (\texttt{plan/approach.md} and
\texttt{plan/design.md}), and the target function signature and
docstring\crfix{, together with the paired unit-test file that defines the
acceptance criterion}, while the reference body and fixtures are
withheld. 
The model iterates in a sandboxed environment until it passes the
paired \texttt{pytest} suite or exhausts a fixed budget (10
reflection rounds).
Correctness is judged under the deterministic-vs-stochastic
tolerance regime described in~\Cref{sec:auto_benchmark_pipeline}.
Per-function pass rates are aggregated into three subtask categories (preprocessing, forward physics modeling, and inverse solver)
for structured failure analysis.
Visualization functions are excluded from this track, as there are
multiple valid ways to display a reconstruction and visual
presentation is not the core scientific deliverable.

\subsubsection{End-to-End Evaluation with a Multi-Agent Solver}
\label{sec:eval_e2e_multiagent}
For both open-weight and API-based LLMs, we evaluate end-to-end performance using a self-developed multi-agent coding pipeline with a simple reflection loop, ensuring that any performance differences are attributable to the models themselves rather than to variations in the agent framework.
The agent receives \texttt{README.md}, \texttt{data/}, and
\texttt{requirements.txt}, and must produce the complete
\texttt{src/} pipeline, execute it, and deliver a reconstruction
satisfying the thresholds in \texttt{metrics.json}.
After each execution the agent receives the full stdout/stderr and a
running error history, and may revise its implementation up to \crfix{five }
rounds.
The role prompts, context-window management strategy, and other implementation details of our self-developed coding agent are described in~\Cref{app:e2e_eval}.

\subsubsection{End-to-End Evaluation of Black-Box Coding Agents}
\label{sec:eval_e2e_blackbox}

To evaluate commercial agentic coding tools that cannot be driven by
a raw API call (e.g., Claude Code), the harness provides an identical
workspace and the same pass criterion.
These results establish an upper bound on the current state-of-the-art
agent capability on scientific computational imaging tasks, and are
reported separately in \Cref{sec:cc_evaluation}

\section{Results}
\label{sec:results}

We evaluate \textbf{seven} LLMs on Imaging-101 across three tracks:
\emph{planning}, \emph{function-level coding}, and \emph{end-to-end execution}.
The seven models are Claude-4.6-Opus, GPT-5.4, Gemini-3.1-Pro,
DeepSeek-V3, Qwen3.6-Plus, Kimi-k2.5, and GLM-5.
Table~\ref{tab:main_results} summarizes all results across the three tracks.
\begin{table*}[p]
    \centering
    \footnotesize
    \setlength{\tabcolsep}{3pt}
    \caption{Imaging-101 results across three evaluation tracks for seven LLMs
    (57 tasks each).
    \textbf{Planning:} pass rates assessed by domain experts with
    LLM-judge assistance; \emph{Overall} requires all three components correct;
    \emph{Avg.\ Plan Length} is the mean word count.
    \textbf{Function-level:} \emph{Module \%} = fraction of modules where
    every function passes (all-or-nothing); \emph{Function \%} = fraction
    of individual functions that pass; \emph{Avg.\ Rounds to Success} = mean
    reflection rounds to success.
    \textbf{End-to-end:} \emph{Success Rate} = fraction of tasks meeting
    image-quality thresholds.
    \textbf{Bold}: best; \underline{underline}: second best.
    }
    \label{tab:main_results}

  \newcommand{\V}[3]{#1\kern0.07em/\kern0.07em#2\,{\scriptsize(#3\%)}}
  \setlength{\tabcolsep}{2pt}
\begin{tabularx}{\textwidth}{@{} X *{7}{c} @{}}
          \toprule
          & \textbf{Claude-4.6-Opus} & \textbf{DeepSeek-V3} & \textbf{Gemini-3.1-Pro}
          & \textbf{GPT-5.4} & \textbf{Qwen3.6-Plus} & \textbf{Kimi-k2.5} & \textbf{GLM-5} \\
          \midrule
          \multicolumn{8}{@{}l}{\textit{Planning --- Pass Rate}} \\
          \quad Preprocessing
            & \underline{\V{39}{57}{68.4}} & \textbf{\V{40}{57}{70.2}} & \V{35}{57}{61.4}
            & \textbf{\V{40}{57}{70.2}} & \underline{\V{39}{57}{68.4}} & \textbf{\V{40}{57}{70.2}} & \V{38}{57}{66.7} \\
          \quad Forward Physics
            & \underline{\V{45}{57}{78.9}} & \V{44}{57}{77.2} & \V{43}{57}{75.4}
            & \textbf{\V{46}{57}{80.7}} & \V{43}{57}{75.4} & \V{43}{57}{75.4} & \V{41}{57}{71.9} \\
          \quad Inverse Solver
            & \underline{\V{52}{57}{91.2}} & \underline{\V{52}{57}{91.2}} & \V{43}{57}{75.4}
            & \textbf{\V{53}{57}{93.0}} & \V{50}{57}{87.7} & \textbf{\V{53}{57}{93.0}} & \V{50}{57}{87.7} \\
          \quad Overall
            & \underline{\V{29}{57}{50.9}} & \V{28}{57}{49.1} & \V{23}{57}{40.4}
            & \textbf{\V{30}{57}{52.6}} & \V{26}{57}{45.6} & \textbf{\V{30}{57}{52.6}} & \V{22}{57}{38.6} \\
          \quad Avg.\ Plan Length (words)
            & 1,125 & 1,175 & 664 & 1,401 & 1,116 & 866 & 992 \\
          \midrule
           \multicolumn{8}{@{}l}{\textit{Function-Level --- Module Pass Rate}} \\
          \quad Preprocessing
            & \underline{\V{43}{57}{75.4}} & \underline{\V{43}{57}{75.4}} & \textbf{\V{45}{57}{78.9}} & \V{42}{57}{73.7} & \V{42}{57}{73.7} & \V{35}{57}{61.4} & \V{26}{57}{45.6} \\
          \quad Forward Physics
            & \textbf{\V{35}{57}{61.4}} & \V{25}{57}{43.9} & \V{30}{57}{52.6} & \underline{\V{32}{57}{56.1}} & \V{24}{57}{42.1} & \V{20}{57}{35.1} & \V{15}{57}{26.3} \\
          \quad Inverse Solver
            & \textbf{\V{28}{57}{49.1}} & \V{18}{57}{31.6} & \V{23}{57}{40.4} & \underline{\V{24}{57}{42.1}} & \V{14}{57}{24.6} & \V{9}{57}{15.8} & \V{6}{57}{10.5} \\
          \quad Overall
            & \textbf{\V{13}{57}{22.8}} & \V{9}{57}{15.8} & \textbf{\V{13}{57}{22.8}} & \underline{\V{12}{57}{21.1}} & \V{7}{57}{12.3} & \V{1}{57}{1.8} & \V{1}{57}{1.8} \\
          \multicolumn{8}{@{}l}{\textit{Function-Level --- Function Pass Rate}} \\
          \quad Preprocessing
            & \V{386}{461}{83.7} & \V{380}{461}{82.4} & \V{388}{461}{84.2} & \textbf{\V{409}{461}{88.7}} & \underline{\V{402}{461}{87.2}} & \V{353}{461}{76.6} & \V{316}{461}{68.5} \\
          \quad Forward Physics
            & \textbf{\V{517}{567}{91.2}} & \V{477}{567}{84.1} & \V{495}{567}{87.3} & \underline{\V{510}{567}{89.9}} & \V{457}{567}{80.6} & \V{426}{567}{75.1} & \V{399}{567}{70.4} \\
          \quad Inverse Solver
            & \underline{\V{337}{409}{82.4}} & \V{318}{409}{77.8} & \V{316}{409}{77.3} & \textbf{\V{338}{409}{82.6}} & \V{294}{409}{71.9} & \V{242}{409}{59.2} & \V{146}{409}{35.7} \\
          \quad Avg.\ Rounds to Success
            & \underline{4.24} & 4.77 & 6.63 & \textbf{2.60} & 8.50 &6.70& 9.28 \\
          \midrule
          \multicolumn{8}{@{}l}{\textit{End-to-End}} \\
          \quad Success Rate
            & \textbf{\V{17}{57}{29.8}} & \V{6}{57}{10.5} & \V{8}{57}{14.0} & \underline{\V{11}{57}{19.3}} & \V{4}{57}{7.0} & \V{4}{57}{7.0} & \V{7}{57}{12.3} \\
          \quad Avg.\ Rounds to Success
            & 2.25 & \underline{1.80} & 2.12 & \textbf{1.45} & 2.25 & 2.75 & 2.14 \\
          \bottomrule
      \end{tabularx}

\end{table*}

\begin{figure*}[p]
    \centering
    \includegraphics[width=\linewidth]{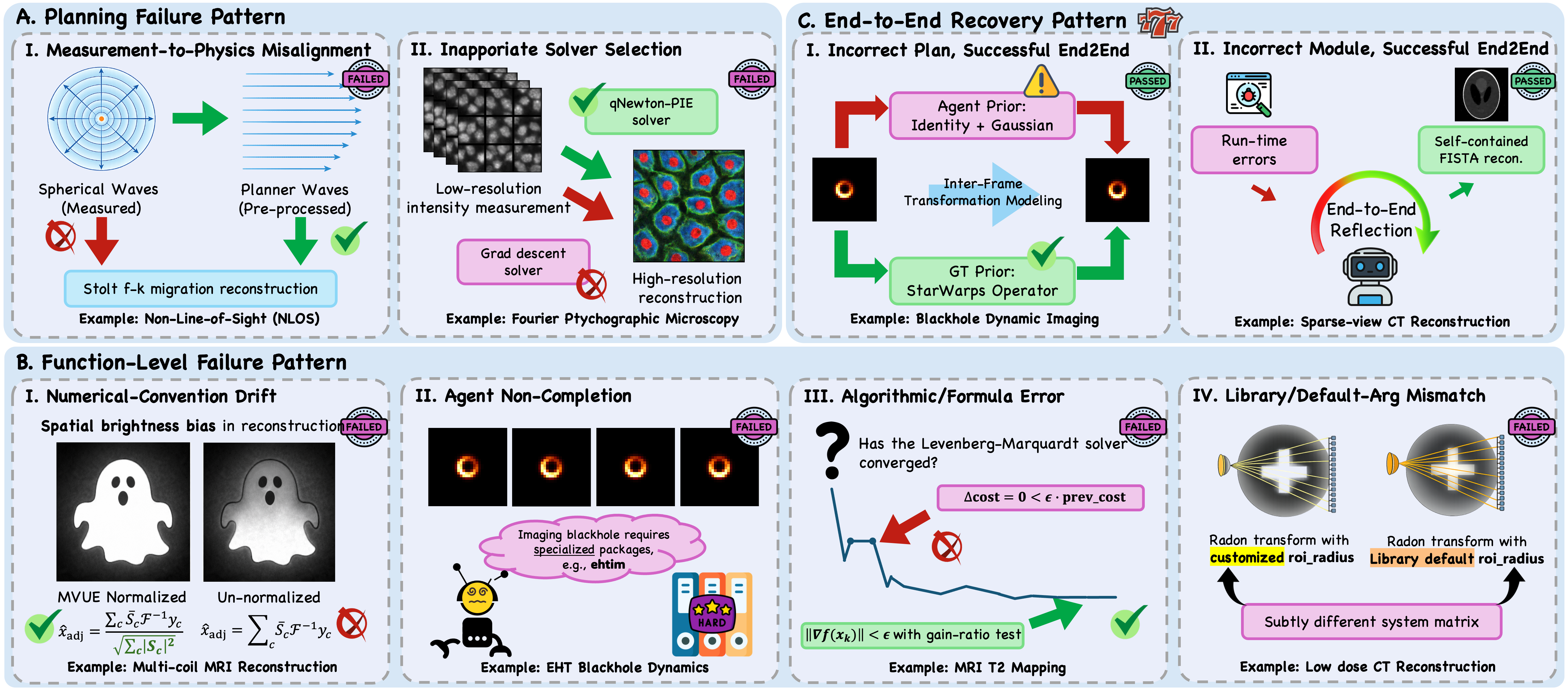}
    \caption{\textbf{Case studies across the three evaluation tracks.}
    \textbf{(A)}~Planning failures fall into two categories:
    (I)~\emph{measurement-to-physics misalignment}, where the agent
    skips a required preprocessing step (e.g., intensity-to-wavefield
    conversion in confocal NLOS), and
    (II)~\emph{inappropriate solver selection}, where the agent
    substitutes a generic optimizer for a physics-coupled algorithm
    (e.g., gradient descent instead of ptychographic phase retrieval).
    \textbf{(B)}~Function-level code bugs fall into four categories:
    (I)~\emph{numerical-convention drift} (e.g., omitting MVUE coil
    normalization in multi-coil MRI, producing spatially varying
    brightness bias),
    (II)~\emph{agent non-completion} (exhausting the reflection budget
    without producing code, common on tasks with exotic libraries),
    (III)~\emph{algorithmic/formula error} (e.g., a Levenberg--Marquardt
    solver with a flawed convergence criterion in MRI T2 mapping), and
    (IV)~\emph{library/default-argument mismatch} (e.g., passing an
    explicit \texttt{roi\_radius} to SVMBIR that differs from the
    library's internal default in low-dose CT).
    \textbf{(C)}~End-to-end recovery patterns show two mechanisms by
    which agents pass despite upstream errors:
    (I)~\emph{incorrect plan, successful end-to-end}, where the agent
    uses a simpler prior (identity + Gaussian noise instead of
    StarWarps' warp operator) that still produces an acceptable
    reconstruction when the test data does not stress the missing
    prior (EHT black hole dynamic imaging), and
    (II)~\emph{incorrect modules, successful recovery via reflection},
    where unit tests fail on the Radon convention and back-projection
    scaling, but runtime errors (shape mismatches, large NRMSE) provide
    actionable feedback that guides the agent to a self-consistent
    FISTA implementation within 5 reflection rounds
    (sparse-view CT reconstruction).}
    \label{fig:failure_mode}
\end{figure*}

\subsection{Planning Evaluation}
\label{sec:planning}

\subsubsection{Quantitative Results}
\label{sec:planning_quant}

The top rows of Table~\ref{tab:main_results} report pass rates for preprocessing, forward physics modeling, and inverse algorithm design across all 57 tasks in the planning-track evaluation of LLMs.
GPT-5.4 and Kimi-k2.5 lead overall (both 52.6\%), with Claude-4.6-Opus close
behind (50.9\%); GPT-5.4 in particular tops or ties for the top on all three
dimensions.
DeepSeek-V3 (49.1\%) trails this leading group only narrowly, while
Qwen3.6-Plus (45.6\%), Gemini-3.1-Pro (40.4\%), and GLM-5 (38.6\%) fall
further back.
Gemini-3.1-Pro underperforms relative to its general capability, due to a tendency toward overly terse plans (\crfix{around 60\% of} the length of other models' on average) that fail to capture domain-specific details.
Consequently, it defaults to generic optimizers (e.g., Adam) on various specific tasks, such as black hole dynamic imaging and compressed-sensing MRI, uses L2 loss where phase-constrained objectives require cosine distance, and omits several required preprocessing steps.

Across all models, preprocessing is the hardest planning dimension
(cross-model mean 67.9\%), trailing both forward physics (76.4\%) and
inverse algorithm selection (88.5\%).
Preprocessing is error-prone because it hinges on recognizing implicit,
task-specific steps that convert raw measurements into the physical quantity
the reconstruction expects; such steps are easy to overlook when the problem
description does not state them explicitly.
Inverse algorithm selection, by contrast, is comparatively well-constrained at
the planning stage: the appropriate solver is usually identifiable from the
problem class, so models can name a suitable algorithm even on tasks where they
later struggle to implement it correctly.

\textbf{Common failure patterns.}
In preprocessing, the dominant failure is \emph{measurement-to-model misalignment}: models overlook
the necessary steps that convert raw sensor measurements into
the physical quantity the reconstruction algorithm expects, or they
confuse the physical regime and select an algorithm that is broadly
appropriate for the imaging modality but wrong for the specific task.
In inverse solver selection, which models handle correctly far more often, the characteristic failure is \emph{inappropriate solver selection}: models correctly name a physics-coupled reconstruction algorithm in the plan but then replace
it with a generic gradient-descent or auto-differentiation scheme,
citing general optimization heuristics that do not account for the
structure of the inverse problem.
Section~\ref{sec:planning_cases} illustrates each pattern with a
concrete example.

\subsubsection{Case Studies}
\label{sec:planning_cases}

We illustrate the two planning failure patterns with concrete examples
(Figure~\ref{fig:failure_mode}A).

\textbf{Case 1: Failure to align measurement data with the physics model.}
In confocal non-line-of-sight (NLOS) imaging, a pulsed laser
illuminates a relay wall and a time-resolved sensor records photons
that scatter off a hidden scene.
The standard reconstruction algorithm, Stolt $f$-$k$
migration~\cite{lindell2019wave}, is derived for a scalar wavefield
amplitude that decays as $1/r$ with distance.
The sensor, however, records intensity, which decays as $1/r^4$
(two independent $1/r^2$ legs).
Bridging the two requires a preprocessing step,
\begin{equation}
  \Psi(x', y', t) = \sqrt{|\tau_c(x', y', t)|\cdot t^{2}},
\label{eq:nlos_psi}
\end{equation}
where the $t^2$ factor compensates for one $1/r^2$ leg and the square
root converts intensity to amplitude.
Skipping this step passes a physically inconsistent quantity to the
algorithm, analogous to supplying a power spectrum where a Fourier
transform is required.

Across all seven models, \emph{no plan} reproduces both operations.
Two models omit preprocessing entirely; three apply a time weighting
with the wrong exponent ($t^1$, $t^3$, or $t^4$); and two apply the
correct $t^2$ factor but omit the square root.
One model explicitly justifies the omission by claiming that the Stolt
mapping's Jacobian handles the amplitude correction, a confident but
incorrect statement, since the Stolt Jacobian is a coordinate-change
factor unrelated to the radial intensity decay.
The failure is hard to detect because the pipeline runs without error
and returns a plausible-looking (but incorrect) reconstruction
regardless of whether the step is applied.

\textbf{Case 2: Inappropriate solver selection.}
Fourier ptychographic microscopy (FPM) reconstructs a high-resolution
complex image from a sequence of low-resolution intensity images
captured under different illumination angles~\cite{Loetgering:23}.
The standard solver is the quasi-Newton ptychographic iteration
(qNewton-PIE)~\cite{yeh2015experimental}, whose update rule uses the
pupil power spectrum as a closed-form Hessian approximation, making
it specifically suited to the phase-retrieval structure of the FPM
forward model.

Three of the seven models plan to use qNewton-PIE or an equivalent
PIE-family update; the remaining four substitute vanilla gradient
descent or Adam.
The substitution takes two forms.
In the \emph{silent} form, the plan never considers the specialized
solver and goes directly to minimizing a data-fidelity loss with Adam.
In the \emph{rationalized} form, the plan correctly identifies
qNewton-PIE by name and then argues against it using generic
optimization heuristics, for example, claiming that ``gradient-based
optimization is the industry standard'' or that the specialized solver
``is overkill here.''
Both arguments are wrong: qNewton-PIE is the domain standard, and
its Hessian preconditioning is precisely what stabilizes optimization
on ill-conditioned phase-retrieval landscapes.
The rationalized form is the more problematic of the two: the agent
has already identified the correct approach and actively discards it,
suggesting that longer reasoning chains can compound rather than
correct the error.

\subsection{Function-Level Evaluation}
\label{sec:function}

\subsubsection{Quantitative Results}
\label{sec:function_quant}

Table~\ref{tab:main_results} reports two granularities of pass rate for each
pipeline stage.
\emph{Module \%} applies a strict all-or-nothing criterion: a module passes
only if every function in it passes its unit test.
\emph{Function \%} measures the fraction of individual functions that pass,
regardless of whether sibling functions in the same module pass.
The gap between the two metrics indicates how often a model gets most of a
module right but fails on one or two critical functions.
A large gap signals that failures are concentrated in a small number of
hard functions rather than spread uniformly across a module.

Preprocessing is the easiest stage by module pass rate (cross-model mean:
69.2\% module, 81.6\% function); inverse solvers are the hardest
(30.6\% module, 69.6\% function).
Forward physics shows a large module--function gap (37.3\,pp, comparable to the
inverse solver's 39.0\,pp),
indicating that most physics functions pass individually but one or
two critical ones (typically the core forward operator or its
adjoint) fail and drag the whole module down.
Overall, only 1.8--22.8\% of pairs pass all three modules simultaneously.

The function-level track also records the number of reflection rounds each
model uses before passing; \emph{Avg.\ Rounds to Success} in
Table~\ref{tab:main_results} reports this quantity.
Models with lower average rounds (e.g., GPT-5.4 at 2.60, Claude-4.6-Opus
at 4.24) converge faster on solvable functions, whereas models with higher
averages (e.g., GLM-5 at 9.28, Qwen3.6-Plus at 8.50) spend more rounds on
trial-and-error debugging before reaching a passing implementation.

\textbf{Failure mode breakdown.}
\crfix{To understand \emph{why} functions fail, not just how often, we
inspected every failing (task, model) pair and classified each module's root
cause into one of four categories (Figure~\ref{fig:failure_mode}B); the
percentages below are taken over these failing pairs.}
Of all failure instances, three categories are genuine code
bugs: numerical-convention drift (42.1\%), algorithmic / formula
error (19.0\%), and library / default-argument mismatch (7.9\%), 
while the fourth, agent non-completion (28.2\%), reflects the agent
exhausting its reflection budget before producing any code.
We describe each category below and illustrate them with case studies
in Section~\ref{sec:function_cases}.

\textbf{Numerical-convention drift (42.1\%).}
This is the dominant code-bug category.
The model's implementation is mathematically self-consistent but uses
a different sign, scale, axis-ordering, or normalization convention
than the reference.
A defining property is its invisibility to standard sanity checks:
the code produces valid output shapes, passes round-trip tests, and
often looks correct visually, but fails tight numerical-parity tests
against the benchmark fixture.

\textbf{Agent non-completion (28.2\%).}
These are not code bugs; the agent exhausts its reflection budget
without producing a compilable module.
Non-completion correlates with library exoticism: tasks built on
specialized packages (\texttt{ehtim}, \texttt{deepwave},
\texttt{fastmri}) time out far more often than tasks using standard
NumPy/SciPy stacks, suggesting that the bottleneck is API
familiarization rather than mathematical reasoning. A second, model-specific driver is over-reasoning: Kimi-k2.5 and GLM-5 in particular spend entire turns on unproductive deliberation, exhausting their token and iteration budgets before emitting a runnable artifact (Appendix~\ref{app:overthinking}).

\textbf{Algorithmic / formula error (19.0\%).}
These cluster at mathematical edge cases: a convergence criterion that
misbehaves on rejected steps, a parametric lens model that gives the
wrong answer in a degenerate limit, or using \texttt{.real} instead
of \texttt{np.abs} on a complex reconstruction.
Unlike convention drift, these errors implement a genuinely different
algorithm: there is no re-parameterization under which the code
reduces to the reference.

\textbf{Library / default-argument mismatch (7.9\%).}
Models substitute a third-party library call (e.g.,
\texttt{scikit-image}, \texttt{SVMBIR}, \texttt{fastMRI}) for the
reference's custom implementation, but a default argument differs or a
reimplemented class is incompatible with pretrained weights.
The call runs without error, so the bug surfaces only when
numerical-parity tests compare against reference fixtures generated
under the original defaults.

\subsubsection{Case Studies}
\label{sec:function_cases}

We illustrate the four failure categories with representative examples
(Figure~\ref{fig:failure_mode}B).

\textbf{Case 1: Numerical-convention drift.}
In compressed-sensing MRI with parallel imaging, multi-coil
measurements follow $y_c = \mathcal{M}\mathcal{F}S_c\,x$, 
where $S_c$ are coil sensitivity maps, $\mathcal{F}$ is the 2D DFT, and $\mathcal{M}$ is a binary undersampling mask. 
The minimum-variance unbiased estimate (MVUE) adjoint combines coils as 

\begin{equation}
\hat{x}_{\mathrm{adj}}=\frac{\sum_c \bar{S}_c\,\mathcal{F}^{-1}\,y_c}{\sqrt{\sum_c \lvert S_c \rvert^{2}}},
\end{equation}

where the denominator compensates for the spatially varying total
sensitivity so that regions with fewer active coils are not
systematically attenuated.
In the \texttt{mri\_l1\_wavelet} task, every model omits the
$1/\sqrt{\sum_c\lvert S_c\rvert^{2}}$ normalization and returns the
unnormalized sum $\sum_c \bar S_c \mathcal{F}^{-1}y_c$ instead.
Because this omission preserves the output shape, dtype, and
FFT round-trip identity, all structural and determinism tests pass.
Only the per-fixture numerical comparison (at \texttt{rtol=1e-10})
against the MVUE-normalized reference reveals the discrepancy.
The resulting image exhibits spatially varying brightness bias, i.e., 
dark in low-sensitivity regions and bright in high-sensitivity
regions, which propagates into the downstream L1-wavelet
reconstruction and degrades both NCC and NRMSE.
Unlike an arbitrary sign or axis convention, the MVUE
normalization is uniquely determined by the physics: it is the
minimum-variance combination for Gaussian noise, and omitting it
changes the reconstruction quality, not merely its numerical
representation.

More broadly, this failure exemplifies a recurring pattern across the
benchmark: the dominant source of function-level errors is not a lack
of general coding ability but a lack of \emph{domain-specific
scientific skills}, i.e., the precise operational knowledge of how
quantities are represented, combined, and normalized within a
particular imaging modality.
MVUE coil combination in MRI, centered-FFT conventions in
$k$-space processing, Parker weighting in fan-beam CT, and
closure-phase sign conventions in radio interferometry are all
instances of such skills: each is a compact, well-defined piece of
domain expertise that recurs across multiple tasks within its modality
and that, once missed, produces code that is structurally correct but
numerically wrong.

\textbf{Case 2: Agent non-completion.}
In the EHT black-hole uncertainty-quantification task, which requires
interfacing with \texttt{ehtim} for NUFFT-based visibility modeling
and variational inference, one model exhausts its full \crfix{10}
-\crfix{round}\ 
reflection budget without producing a \texttt{physics\_model.py} file.
The run log shows only diagnostic scripts in \texttt{files\_created}.
This pattern recurs on tasks built atop specialized packages
(\texttt{ehtim}, \texttt{deepwave}, \texttt{fastmri}): the agent
spends its iteration budget learning the unfamiliar API rather than
writing the target function, suggesting that the bottleneck is API
familiarization rather than mathematical reasoning.

\textbf{Case 3: Algorithmic / formula error.}
T2 mapping fits a mono-exponential decay model to a series of MRI
echo images.
One model's Levenberg-Marquardt solver tests convergence by
$|\Delta\texttt{cost}| < \epsilon \cdot \texttt{prev\_cost}$ and updates
\texttt{prev\_cost} unconditionally on every iteration,
\emph{including rejected trial steps}.
Two consecutive rejections yield $|\Delta\text{cost}| = 0$, triggering
premature convergence at a stuck, non-optimal point.
The reference implementation uses a gradient-norm stopping criterion
($\max|J^T r| < \epsilon_1$) combined with a gain-ratio test for step
rejection, which correctly distinguishes convergence from stalling.
On clean-data and poor-initialization regression tests, the flawed
solver returns \texttt{T2\_fit} values far from the ground truth
with \texttt{converged=True}.

\textbf{Case 4: Library / default-argument mismatch.}
In low-dose CT reconstruction, SVMBIR's projector uses a
region-of-interest radius (\texttt{roi\_radius}) that defaults to
\texttt{None}, letting the library compute it internally.
One model guesses the default formula as
\texttt{0.5 * num\_channels} and passes it explicitly, but SVMBIR's
actual default uses \texttt{num\_channels - 1}, producing a
subtly different system matrix.
Shape and non-negativity tests pass; only \texttt{rtol=1e-10}
fixture parity exposes the mismatch.
This illustrates a general trap: the LLM reaches for the most
syntactically economical library call without verifying that its
defaults match the reference implementation's explicit parameters.

\subsection{End-to-End Evaluation with A Multi-Agent Solver}
\label{sec:end2end}

\subsubsection{Quantitative Results}
\label{sec:end2end_quant}

Table~\ref{tab:main_results} reports end-to-end pass rates across all seven models.
Claude-4.6-Opus leads at 29.8\%, well ahead of GPT-5.4 (19.3\%);
Gemini-3.1-Pro (14.0\%), GLM-5 (12.3\%), and DeepSeek-V3 (10.5\%) form a middle
group, while Kimi-k2.5 and Qwen3.6-Plus trail at 7.0\%.
The top success rate is over four times the lowest, a wide spread that reflects
how end-to-end execution compounds the per-stage gaps seen above.

\textbf{Relationship between planning, function-level, and end-to-end success.}
Table~\ref{tab:e2e_breakdown} breaks down end-to-end pass rates for Claude-4.6-Opus
across the four combinations of planning and function-level outcomes, showing
that both dimensions contribute but neither alone is sufficient.

The joint condition (correct plan \emph{and} correct function-level modules) yields
the highest pass rate (71.4\%, 5/7), the only stratum in which a majority of
tasks succeed end-to-end.
The remaining failures in this best-case stratum reflect \emph{pipeline integration
errors}: assembling a complete solution requires maintaining consistent data shapes,
variable names, and interface contracts across multiple source files within a long
context window, a global coherence requirement that neither per-dimension planning
review nor per-function unit tests can capture.

Comparing the marginals reveals an asymmetry between the two dimensions.
Planning correctness has a larger effect on end-to-end success: the pass rate drops
from 44.8\% to 14.3\% between plan-correct and plan-incorrect tasks, whereas the
drop for function-level correctness is smaller (46.2\% to 25.0\%).
Getting the overall algorithmic direction right matters more for the final outcome
than getting every individual function exactly right.

The non-zero pass rates in the off-diagonal cells reflect two distinct recovery
mechanisms.
When function-level modules fail but the plan is sound (36.4\%, 8/22), the
end-to-end reflection loop acts as a secondary correction channel: errors that
surface as explicit runtime signals (shape mismatches, diverging losses, or poor
reconstruction metrics) provide targeted feedback that allows the agent to
self-correct across rounds, even without passing unit tests.
When the plan is incorrect but modules are implemented correctly (16.7\%, 1/6),
the agent has executed the wrong algorithm faithfully; recovery is possible when
the alternative approach still yields a reconstruction within the evaluation margin.
Even when both plan and modules are incorrect, end-to-end recovery occasionally
occurs (13.6\%, 3/22), driven by the same reflection mechanism but at the lowest rate.

\begin{table}[h]
\centering
\caption{%
  Cross-tabulation of end-to-end pass rates for Claude-4.6-Opus (57 tasks)
  stratified by planning correctness and function-level module correctness.
  Each cell gives the end-to-end pass rate as a percentage with the
  corresponding count (pass\,/\,stratum size) in parentheses.
  Row and column marginals show the aggregate pass rate for each
  function-level and planning stratum, respectively.
  \emph{Plan correct}: all three planning dimensions (preprocessing, forward
  physics, inverse solver) pass expert review.
  \emph{Function-level correct}: every function in every module passes its
  unit test.
}
\label{tab:e2e_breakdown}
\small
\setlength{\tabcolsep}{4pt}
\begin{tabular}{l*{3}{c}}
\toprule
 & Plan \checkmark & Plan \texttimes & Total\\
\midrule
Func.\ \checkmark & 5/7\,{\scriptsize(71.4\%)}  & 1/6\,{\scriptsize(16.7\%)}  & 6/13\,{\scriptsize(46.2\%)}  \\[2pt]
Func.\ \texttimes & 8/22\,{\scriptsize(36.4\%)} & 3/22\,{\scriptsize(13.6\%)}  & 11/44\,{\scriptsize(25.0\%)} \\
\midrule
Total              & 13/29\,{\scriptsize(44.8\%)} & 4/28\,{\scriptsize(14.3\%)} & 17/57\,{\scriptsize(29.8\%)} \\
\bottomrule
\vspace{-2.4em}
\end{tabular}
\end{table}





\subsubsection{Case Studies}
\label{sec:e2e_cases}

The following two cases illustrate the off-diagonal recovery mechanisms
identified in Table~\ref{tab:e2e_breakdown}.

\textbf{Case 1: Incorrect plan, successful end-to-end.}
The black hole dynamic imaging task reconstructs a time-varying black-hole image
sequence from sparse radio interferometric measurements. The reference method,
StarWarps~\cite{bouman2018reconstructing}, models frame-to-frame evolution through a warp
operator, an optical-flow-like transformation that enforces smooth, structured
motion between consecutive frames, and the task README explicitly hints at this
prior. Claude's plan discarded the warp prior and instead modeled inter-frame
dynamics as an identity transformation plus white Gaussian noise, treating the
sequence as a near-static image with independent per-frame perturbations.
Despite the planning error, the end-to-end run passes.
The test sequence involves slow, smooth motion, so the simplified model produces
a reconstruction whose NCC and NRMSE fall within the evaluation margin
(0.9$\times$ reference NCC, 1.1$\times$ reference NRMSE); the metrics are
noticeably worse than the reference but above the pass threshold.
This illustrates the recovery mechanism noted above: when a simpler alternative
is physically plausible for the test data, an incorrect plan can still yield a
passing reconstruction.
In practice, a simple working prototype is often a legitimate starting point in
scientific imaging, and more complex priors are justified only when the data
demand them.

\textbf{Case 2: Incorrect modules, successful end-to-end recovery.}
The sparse-view CT task reconstructs a 2D attenuation image from
a 30-view sinogram using TV-regularized reconstruction.
Module-level evaluation exposed clear failures: the Radon operator used
\texttt{skimage} interpolation conventions that did not match the reference,
and the solver's back-projection scaling was inconsistent with the reference
PDHG implementation, causing both unit tests to fail.
Despite these failures, the end-to-end reflection loop provided richer
feedback than unit tests alone: failed reconstruction in early rounds prompted
the agent to revise both the forward model and the solver, and by round~5
it had converged to a self-consistent FISTA implementation that achieved
$\mathrm{NCC}=0.965$ and $\mathrm{NRMSE}=0.071$.
Recovery was possible because each error produced an explicit runtime
signal (shape mismatches and large metric values) that guided targeted fixes.
This contrasts with numerical-convention drift (Section~\ref{sec:function_quant}),
where the implementation runs without error and end-to-end metric feedback
alone is insufficient to identify the source of the discrepancy.

\crfix{\textbf{When image metrics are not enough.}
The end-to-end track scores reconstructions with NCC and NRMSE, which capture
complementary aspects of image quality (structural agreement and numerical
accuracy) and are jointly hard to game. They cannot, however, certify that a
solution is \emph{physically} correct: as the convention-drift and black hole dynamic imaging task
cases above show, a pipeline can be physically wrong yet visually plausible.
This is exactly why Imaging-101 pairs end-to-end scoring with the planning and
function-level tracks, whose operator- and function-level parity checks expose
physics errors that image metrics miss. We deliberately do not impose these
intermediate checks inside the end-to-end track, so as not to over-constrain the
implementation choices we aim to evaluate. Appendix~\ref{app:additional_analyses}
analyzes metric choice (including a comparison with PSNR/SSIM) and the use of
multiple test instances per task.}

\vspace{-0.7em}

\subsection{End-to-End Evaluation of Black-Box Coding Agents}                                                 
  \label{sec:cc_evaluation}                         To complement the sandboxed multi-agent evaluation above, we                                                  
  additionally benchmark \textbf{Claude Code}~(Anthropic, 2025),
  a black-box coding agent with access to a local shell, file system,
  and package manager, on the same 57-task suite.
  Unlike the multi-agent pipeline, Claude Code operates as a single
  opaque agent: it receives only \texttt{README.md}, raw observation
  data, and a standardized \texttt{INSTRUCTIONS.md} prompt, then
  works autonomously until it produces a reconstruction artifact.
  The same NCC/NRMSE evaluation protocol is applied.

  Claude Code achieves a pass rate of \textbf{56.1\%} (32/57 tasks),
  substantially outperforming the best sandboxed model
  (\crfix{Claude-4.6-Opus}, 29.8\%).
  The performance gap highlights the benefit of richer execution
  affordances, including unrestricted package installation, iterative
  debugging with shell feedback, and flexible file
  exploration, capabilities that the sandboxed agents lack.
  As Claude Code operates under a different execution environment,
  we report its results separately as a reference point rather than
  including it in the controlled comparison in
  Table~\ref{tab:main_results}.

\vspace{-0.8em}

\section{Related Work}
\label{sec:related_work}

\subsection{Scientific Reasoning and Coding Benchmarks}
LLM evaluations have progressed from knowledge-oriented QA~\cite{wang2023scibench} through execution-graded coding benchmarks~\cite{chen2021evaluating,jimenez2023swe,zhuo2024bigcodebench} to scientific coding benchmarks that test runnable solutions~\cite{tian2024scicode}, data-driven workflows~\cite{chen2024scienceagentbench}, ML pipelines~\cite{chan2024mle}, and multi-step analysis~\cite{gu2024blade}. These benchmarks motivate Imaging-101's execution-based evaluation paradigm. However, they predominantly operate under a \textbf{clean data and clean interface} assumption: agents receive structured inputs and act as analysts on prepared artifacts. In scientific imaging, the upstream bottleneck is \emph{computational perception}: measurements are indirect, noisy, and shaped by a physical measurement operator, and correctness depends on faithfully modeling that operator.

Domain-specific benchmarks have recently emerged for astronomy~\cite{joseph2025astrovisbench}, biomedical data science~\cite{wang2025biodsa}, CFD~\cite{somasekharan2025cfdllmbench}, and multimodal scientific workflows~\cite{sun2025scienceboard}. None center on \emph{scientific computational imaging}, the process of recovering latent physical quantities from indirect measurements through a forward model, posed as an inverse problem. Imaging-101 fills this gap.

\subsection{LLM Reasoning Limitations in Scientific Problem-Solving}
GPQA~\cite{rein2024gpqa} demonstrates that frontier models achieve substantially lower accuracy than domain experts on graduate-level science questions, with failures concentrated on multi-step application rather than knowledge recall. Schmied et al.~\cite{schmied2025llms} formalize a ``knowing-doing'' gap: models possess knowledge but fail to deploy it under implicit constraints. PhyBench~\cite{qiu2025phybench} corroborates this, showing that models struggle to apply physical laws when problems require integrating multiple principles or reasoning about non-standard configurations.

Our evaluation reveals an analogous phenomenon in computational imaging, which we characterize as \emph{probabilistic mode collapse}: models overwhelmingly default to the statistically dominant convention in their training distribution even when the task context requires a different one. This extends the knowing-doing gap from QA settings to executable scientific computation, where defaulting to the wrong convention produces code that is structurally correct but numerically wrong, the dominant failure mode in our function-level analysis.
\crfix{A controlled repetition study makes this concrete: across 50 independent
runs on the \texttt{mri\_l1\_wavelet} task, Claude-4.6-Opus reproduced the
convention-drift failure in 47 and succeeded in only 3, indicating that the
model possesses but fails to reliably invoke the required knowledge
(Appendix~\ref{app:modecollapse}).}

\section{Conclusion}
\label{sec:conclusion}

We introduced Imaging-101, a benchmark of 57 expert-verified computational
imaging tasks across six scientific domains, supporting planning, function-level,
and end-to-end evaluation through a standardized four-stage pipeline
(preprocessing, forward physics, inverse solver, and visualization).
Every task is grounded in a peer-reviewed publication and released with reference
implementations, evaluation fixtures, and executed Jupyter notebooks.

Evaluating seven frontier LLMs reveals two layers of challenge.
The first is shared with general coding benchmarks: convention drift, API
mismatches, and pipeline integration errors dominate failures.
The second is specific to computational imaging: models systematically struggle
to select physically correct inverse solvers, and routinely miss the unit
conventions, normalization factors, and measurement-model coupling that
practitioners treat as tacit knowledge but that never appear in papers.
These invisible details are the dark matter of computational imaging
pipelines, decisive for correctness yet absent from any written specification. Looking forward, skill libraries that codify this tacit knowledge as reusable,
verified modules offer a natural path toward domain-specialized imaging copilots
and, ultimately, \crfix{progress toward} automated solvers for scientific inverse problems across domains.

\textbf{Limitations.}
Imaging-101 covers only tasks with publicly available code and data, which may
underrepresent proprietary or emerging domains.
\crfix{Because every task derives from a published paper and open-source code, we
cannot fully rule out pretraining leakage; we treat this as an unavoidable but
measurable limitation and quantify it through a decade-wise analysis and a Codex
re-cleaning control in Appendix~\ref{app:additional_analyses}.}
All models are evaluated with a single agent framework and fixed reflection budget, so results reflect joint model-harness capability rather than intrinsic model ability in isolation.
\crfix{Evaluation is also fully autonomous, which yields a deterministic and
reproducible protocol; human-in-the-loop assessment is a valuable complementary
direction, and prior work reports that autonomous and semi-autonomous evaluation
correlate strongly~\cite{pmlr-v235-chiang24b}.}
We welcome community contributions of new tasks following the standardized pipeline to broaden domain and modality coverage.

\vspace{-3em}

\vskip 4em


\bibliographystyle{IEEEtran}
\bibliography{references}

\ifpeerreview \else
\begin{IEEEbiographynophoto}{Siyi Chen} received the B.S. degree in physics from Peking University, Beijing, China. He is currently pursuing the Ph.D. degree with the Department of Electrical Engineering and Computer Science (EECS), University of Michigan, Ann Arbor, MI, USA. His research interests include probabilistic machine learning, diffusion models, and AI for science, such as data assimilation.
\end{IEEEbiographynophoto}

\begin{IEEEbiographynophoto}{Jiahe Ying}
is currently working toward the B.S. degree with the School of Physics, Peking University. His research lies at the intersection of artificial intelligence and natural sciences, focusing on building AI systems to accelerate scientific discovery across disciplines.
\end{IEEEbiographynophoto}

\begin{IEEEbiographynophoto}{Yixuan Jia}
received the B.S. degree in Department of Precision Instrument from Tsinghua University, Beijing, China. He is currently pursuing the Ph.D. degree with University of Michigan, Ann Arbor, MI, United States. His research interests include generative models, representation learning and agentic AI, with applications to scientific domains, e.g., data assimilation and computational imaging.
\end{IEEEbiographynophoto}

\begin{IEEEbiographynophoto}{Yuxuan Gu}
is currently persuing the master degree at the School of Software and Microelectronics, Peking University, where his research interests primarily focus on the distillation of diffusion models, inverse problem solving, and the development of intelligent agents.
\end{IEEEbiographynophoto}

\begin{IEEEbiographynophoto}{Enze Ye}
received his B.S. in Electronic Engineering from Tsinghua University. He is currently pursuing a Ph.D. in Biomedical Engineering at the College of Future Technology, Peking University, with research interests in computational imaging.
\end{IEEEbiographynophoto}

\begin{IEEEbiographynophoto}{Weimin Bai}
received the B.S. degree from China Agricultural University and the M.E. degree from Southeast University. He is currently pursuing a PhD degree with the Academy for Advanced Interdisciplinary Studies, Peking University, China. His main research interests include generative models and inverse problems.
\end{IEEEbiographynophoto}

\begin{IEEEbiographynophoto}{Zhijun Zeng}
received the B.S. degree in mathematics from Shanghai University of Finance and Economics, Shanghai, China. He is currently pursuing the Ph.D. degree with the Yau Mathematical Sciences Center, Tsinghua University, Beijing, China. His research interests include computational imaging, inverse problems, numerical methods for partial differential equations, and machine learning for scientific computing.
\end{IEEEbiographynophoto}

\begin{IEEEbiographynophoto}{Shaochi Ren}
received his B.S. in Mechanical Engineering from Peking University. He is currently pursuing a Ph.D. in Biomedical Engineering at the College of Future Technology, Peking University, with research interests in computational imaging.
\end{IEEEbiographynophoto}

\begin{IEEEbiographynophoto}{Binhong Gao}
received the B.S. degree from Tianjin University, Tianjin, China, in 2026. He is currently pursuing the Ph.D. degree with the College of Future Technology, Peking University, Beijing, China, under the supervision of Prof. He Sun. His research interests include ptychography, computational imaging, and coherent diffractive imaging.
\end{IEEEbiographynophoto}

\begin{IEEEbiographynophoto}{Yubing Li}
(Member, IEEE) received the B.Sc. degree from Tongji University, Shanghai, China, in 2011, the M.Sc. degree from Universit\'e Paris Diderot--Paris 7, Paris, France, in 2014, and the Ph.D. degree from Paris Sciences \& Lettres (PSL) University, Paris, France, in 2018. He is currently a Researcher with the Institute of Acoustics, Chinese Academy of Sciences, Beijing, China. His main research interests include signal processing, computational imaging, and artificial intelligence research in medical and detection acoustics.
\end{IEEEbiographynophoto}

\begin{IEEEbiographynophoto}{Tianhan Zhang}
received his B.S. from Peking University and his Ph.D. from Princeton University. He is currently a Professor in the School of Astronautics at Beihang University focusing on AI for Science.
\end{IEEEbiographynophoto}

\begin{IEEEbiographynophoto}{He Sun} (Member, IEEE)
 is an Assistant Professor at the College of Future Technology and the National Biomedical Imaging Center, Peking University, China. Prior to joining Peking University, he was a Postdoctoral Researcher in the Department of Computing and Mathematical Sciences at the California Institute of Technology. He received the Ph.D. degree in Mechanical and Aerospace Engineering from Princeton University in 2019 and the bachelor's degree in Engineering Mechanics and Economics from Peking University in 2014. His research combines physics-grounded AI algorithms with hardware innovations for imaging at extreme scales. His past work has supported various challenging science missions, such as black hole imaging and the search for Earth-like exoplanets, as well as biomedical and industrial imaging modalities including ultrasound tomography and computational microscopy.
\end{IEEEbiographynophoto}
\fi


\onecolumn
\appendices
\setcounter{figure}{0}
\renewcommand{\thefigure}{S\arabic{figure}}

\newcounter{reptaskcnt}
\setcounter{reptaskcnt}{0}
\newcommand{\reptaskcap}[3][]{%
  \stepcounter{reptaskcnt}%
  \edef\rtasknum{\ifnum\value{reptaskcnt}<10 0\arabic{reptaskcnt}\else\arabic{reptaskcnt}\fi}%
  \ifx&#1&%
    \caption[\textbf{Rep.\ Task \rtasknum}: #2]{\textbf{Representative Task \rtasknum:} #2 #3}%
  \else
    \caption[#1]{\textbf{Representative Task \rtasknum:} #2 #3}%
  \fi
}

\section{Task List and Dataset}
\label{app:dataset}


\subsection{Full Task List}
\label{app:task_list}

Imaging-101 contains 57 tasks spanning six scientific domains.
Table~\ref{tab:task_list} lists every task together with its domain,
one-sentence description, primary reference, and the public repository
or project page used as the reference implementation. \crfix{Our website is in \url{https://starpacker.github.io/agent-imaging-website/} and the implementation is available at \url{https://github.com/AI4ImagingLab/imaging-101-release}}


%

%

%

\newcommand{\tsk}[1]{{\small\ttfamily\def\_{\textunderscore\penalty0 }#1}}

\begin{longtable}{>{\raggedright\arraybackslash}p{4.0cm} p{6.5cm} p{3.5cm}}
\caption{Complete task list for Imaging-101.
Each row lists the task name, a one-sentence description,
and the primary reference.}
\label{tab:task_list} \\
\toprule
Task & Description & Reference \\
\midrule
\endfirsthead
\multicolumn{3}{c}{\tablename\ \thetable{} (continued)} \\
\toprule
Task & Description & Reference \\
\midrule
\endhead
\midrule
\multicolumn{3}{r}{\emph{Continued on next page}} \\
\endfoot
\bottomrule
\endlastfoot

\multicolumn{3}{l}{\textbf{Astronomy}} \\[2pt]

\tsk{eht\_black\_hole\_original}
  & Recover a black hole radio image from gain-corrupted visibilities using closure-quantity RML.
  & Chael et al.\ (2018); EHT Collab.\ (2019) \\

\tsk{eht\_black\_hole\_dynamic}
  & Reconstruct a time-varying black hole video from sparse per-frame interferometric measurements.
  & Bouman et al.\ (2017); EHT Collab.\ (2019) \\

\tsk{eht\_black\_hole\_UQ}
  & Learn the posterior distribution over black hole images from sparse interferometric data via normalizing flows.
  & Sun \& Bouman (2020); Dinh et al.\ (2017) \\

\tsk{eht\_black\_hole\_feature\_extraction\_dynamic}
  & Infer posterior distributions of black hole crescent geometry from time-varying closure quantities.
  & Sun et al.\ (2022); EHT Collab.\ (2022) \\

\tsk{eht\_black\_hole\_tomography}
  & Recover 3D black hole accretion emission from time-series images via gravitational tomography.
  & Levis et al.\ (2022); Levis et al.\ (2024) \\

\tsk{lucky\_imaging}
  & Reconstruct a high-resolution lunar image from a short-exposure video using lucky-frame selection.
  & Fried (1978); Law et al.\ (2006) \\

\tsk{exoplanet\_imaging}
  & Detect a faint exoplanet from a coronagraphic image sequence using KLIP-ADI PSF subtraction.
  & Soummer et al.\ (2012); Ko et al.\ (2024) \\

\tsk{shack\_hartmann}
  & Reconstruct the pupil wavefront phase from Shack-Hartmann detector spot images.
  & Por et al.\ (2018); Hardy (1998) \\

\tsk{shapelet\_source\_reconstruction}
  & Reconstruct an unlensed source galaxy from a gravitationally lensed image via shapelet decomposition.
  & Refregier (2003); Birrer et al.\ (2021) \\

\midrule
\multicolumn{3}{l}{\textbf{Biology}} \\[2pt]

\tsk{SSNP\_ODT}
  & Recover the 3D refractive index distribution of a biological sample from intensity-only diffraction tomography measurements.
  & Zhu et al.\ (2022); Lim et al.\ (2019) \\

\tsk{reflection\_ODT}
  & Reconstruct the 3D refractive index of a sample from multi-angle reflected intensity measurements.
  & Zhu et al.\ (2025) \\

\tsk{fourier\_ptychography}
  & Synthesize a wide-field high-resolution complex image from LED-illuminated low-resolution captures.
  & Loetgering et al.\ (2023) \\

\tsk{microscope\_denoising}
  & Restore low-SNR fluorescence microscopy images using zero-shot self-supervised deconvolution-denoising.
  & Qiao et al.\ (2024) \\

\tsk{light\_field\_microscope}
  & Reconstruct a 3D fluorescent volume from a single 2D light-field microscope capture via deconvolution.
  & Broxton et al.\ (2013) \\

\tsk{single\_molecule\_light\_field}
  & Localize single fluorescent molecules in 3D from multi-view Fourier light-field detections.
  & Sims et al.\ (2020) \\

\tsk{fpm\_inr\_reconstruction}
  & Reconstruct a high-resolution 3D complex field from multiplexed FPM data using implicit neural representations.
  & Zhou et al.\ (2023) \\

\tsk{s2ism}
  & Reconstruct a super-resolved fluorescence image from multi-channel SPAD image-scanning microscopy data.
  & Castello et al.\ (2019) \\

\tsk{hessian\_sim}
  & Reconstruct a $2\times$ super-resolved fluorescence image from structured illumination microscopy raw frames using Hessian regularization.
  & Huang et al.\ (2018) \\

\midrule
\multicolumn{3}{l}{\textbf{Physics}} \\[2pt]

\tsk{conventional\_ptychography}
  & Recover the complex amplitude of a sample from overlapping far-field diffraction patterns.
  & Loetgering et al.\ (2023) \\

\tsk{spectral\_snapshot\_compressive\_imaging}
  & Reconstruct a 3D hyperspectral cube from a single 2D coded-aperture snapshot measurement.
  & Zheng et al.\ (2021); Yuan (2016) \\

\tsk{electron\_ptychography}
  & Reconstruct the complex transmission function of nanoparticles from 4D-STEM electron diffraction patterns.
  & Rodenburg \& Faulkner (2004); Ophus (2019) \\

\tsk{confocal\_nlos\_fk}
  & Recover a hidden scene from time-resolved confocal relay-wall measurements using Stolt $f$-$k$ migration.
  & Lindell et al.\ (2019); O'Toole et al.\ (2018) \\

\tsk{lensless\_imaging}
  & Reconstruct a scene from a single DiffuserCam raw capture by inverting a known diffuser PSF.
  & Bezzam et al.\ (2023); Boyd et al.\ (2011) \\

\tsk{differentiable\_deflectometry}
  & Recover surface curvatures of a refractive optical element from phase-shifted fringe images.
  & Wang et al.\ (2021) \\

\midrule
\multicolumn{3}{l}{\textbf{Chemistry \& Material Science}} \\[2pt]

\tsk{mcr\_hyperspectral}
  & Recover pure spectral components and spatial concentration maps from a noisy hyperspectral image via MCR-ALS.
  & Camp (2019) \\

\tsk{raman\_cell\_phenotyping}
  & Unmix volumetric Raman spectroscopic data of cells into biomolecular endmember abundance maps.
  & Kallepitis et al.\ (2017); Winter (1999) \\

\tsk{cars\_spectroscopy}
  & Recover gas temperature from a synthetic N$_2$ CARS spectrum under high-pressure combustion conditions.
  & Kataoka et al.\ (1982); Palmer (1989) \\

\tsk{xray\_ptychography\_tike}
  & Reconstruct a complex transmission function from coherent X-ray far-field diffraction patterns.
  & Thibault et al.\ (2008); Ching et al.\ (2024) \\

\tsk{xray\_laminography\_tike}
  & Reconstruct a 3D complex-valued volume from parallel-beam X-ray projections in laminography geometry.
  & Ching \& Gursoy (2020) \\

\midrule
\multicolumn{3}{l}{\textbf{Earth Science}} \\[2pt]

\tsk{seismic\_FWI\_original}
  & Recover a subsurface P-wave velocity model from surface seismograms using differentiable full waveform inversion.
  & Pasalic \& McGarry (2010) \\

\tsk{seismic\_traveltime\_tomography}
  & Recover a 2D P-wave velocity model from first-arrival traveltimes using adjoint-state eikonal tomography.
  & Chen et al.\ (2024); Sethian (1996) \\

\tsk{seismic\_lsrtm\_original}
  & Recover a 2D subsurface reflectivity image from surface seismograms using least-squares reverse-time migration.
  & Pasalic \& McGarry (2010) \\

\tsk{insar\_phase\_unwrapping}
  & Recover absolute surface deformation phase from a $2\pi$-wrapped InSAR interferogram.
  & Chen \& Zebker (2001); Chartrand et al.\ (2019) \\

\tsk{era5\_tensorvar}
  & Reconstruct ERA5 atmospheric state fields from sparse observations using tensor-decomposed 4D-Var assimilation.
  & Yang et al.\ (2025); Hersbach et al.\ (2020) \\

\tsk{weather\_radar\_data\_assimilation}
  & Reconstruct gridded precipitation fields from sparse weather radar observations using flow-based data assimilation.
  & Ravuri et al.\ (2021) \\

\midrule
\multicolumn{3}{l}{\textbf{Medicine}} \\[2pt]

\tsk{ct\_fan\_beam}
  & Reconstruct a 2D attenuation image from fan-beam CT sinograms using Parker-weighted filtered back-projection.
  & Kak \& Slaney (1988); Parker (1982) \\

\tsk{ct\_sparse\_view}
  & Reconstruct a 2D phantom from 30 sparse-view Radon projections via TV-minimization with PDHG.
  & Sidky \& Pan (2008); Chambolle \& Pock (2011) \\

\tsk{ct\_poisson\_lowdose}
  & Reconstruct an attenuation image from Poisson-noise low-dose sinograms using model-based statistical reconstruction.
  & Bouman \& Sauer (1993); Thibault et al.\ (2007) \\

\tsk{ct\_dual\_energy}
  & Decompose dual-energy CT sinograms into basis material density maps for tissue characterization.
  & Alvarez \& Macovski (1976); Hubbell \& Seltzer (1995) \\

\tsk{xray\_tooth\_gridrec}
  & Reconstruct a 2D cross-section of a tooth from parallel-beam X-ray sinograms using gridrec.
  & Dowd et al.\ (1999); Gursoy et al.\ (2014) \\

\tsk{mri\_l1\_wavelet}
  & Reconstruct a phantom from 8$\times$-undersampled multi-coil k-space using $\ell_1$-wavelet compressed sensing.
  & Lustig et al.\ (2007); Beck \& Teboulle (2009) \\

\tsk{mri\_tv}
  & Reconstruct knee MRI from 8$\times$-undersampled k-space using total-variation regularization.
  & Lustig et al.\ (2007); Block et al.\ (2007) \\

\tsk{mri\_t2\_mapping}
  & Estimate per-pixel T2 relaxation maps from a multi-echo spin-echo MRI magnitude series.
  & Hennig (1988); Prasloski et al.\ (2012) \\

\tsk{mri\_sense}
  & Reconstruct a 4$\times$-accelerated multi-coil MRI image using conjugate-gradient SENSE.
  & Pruessmann et al.\ (1999, 2001) \\

\tsk{mri\_grappa}
  & Reconstruct a 2$\times$-accelerated multi-coil brain MRI using auto-calibrated GRAPPA k-space interpolation.
  & Griswold et al.\ (2002) \\

\tsk{mri\_noncartesian\_cs}
  & Reconstruct a complex MRI image from radially sampled multi-coil k-space via NUFFT-based compressed sensing.
  & Lustig et al.\ (2007); Pipe \& Menon (1999) \\

\tsk{diffusion\_mri\_dti}
  & Estimate per-voxel diffusion tensors and fractional-anisotropy maps from multi-direction diffusion-weighted MRI.
  & Basser et al.\ (1994); Salvador et al.\ (2005) \\

\tsk{mri\_dynamic\_dce}
  & Reconstruct a DCE-MRI time series from per-frame undersampled k-space via temporal total-variation regularization.
  & Lustig et al.\ (2007); Feng et al.\ (2014) \\

\tsk{mri\_pnp\_admm}
  & Reconstruct a brain MRI from 30\% k-space using Plug-and-Play ADMM with a learned CNN denoiser prior.
  & Ryu et al.\ (2019); Zhang et al.\ (2017) \\

\tsk{mri\_varnet}
  & Reconstruct 320$\times$320 knee MRI from 4$\times$-accelerated k-space using an end-to-end unrolled VarNet.
  & Sriram et al.\ (2020); Zbontar et al.\ (2018) \\

\tsk{pnp\_mri\_reconstruction}
  & Reconstruct undersampled MRI using Plug-and-Play priors with a multi-scale self-supervised sparsifying network.
  & Song et al.\ (2020) \\

\tsk{plane\_wave\_ultrasound}
  & Reconstruct focused B-mode ultrasound images from multi-angle plane-wave RF data via Stolt f-k migration.
  & Garcia et al.\ (2013) \\

\tsk{ultrasound\_sos\_tomography}
  & Reconstruct a 2D speed-of-sound map from ultrasound transmission travel-time measurements.
  & Duric et al.\ (2007); Li et al.\ (2009) \\

\tsk{usct\_FWI}
  & Reconstruct breast tissue sound-speed distribution from ring-array ultrasound transmission data using FWI.
  & Osnabrugge et al.\ (2016); Wiskin et al.\ (2013) \\

\tsk{pet\_mlem}
  & Reconstruct a 2D PET activity distribution from Poisson-noisy sinograms via MLEM / OS-EM.
  & Shepp \& Vardi (1982); Hudson \& Larkin (1994) \\

\tsk{photoacoustic\_tomography}
  & Recover an initial pressure distribution from circular-array photoacoustic time-series using back-projection.
  & Xu \& Wang (2005) \\

\tsk{eit\_conductivity\_reconstruction}
  & Reconstruct 2D internal conductivity from boundary voltage measurements using regularized EIT inversion.
  & Adler \& Lionheart (2006); Adler et al.\ (2009) \\

\end{longtable}
\twocolumn

%

\definecolor{readmebg}{HTML}{EEF6FF}
\definecolor{readmeframe}{HTML}{4A90D9}
\definecolor{approachbg}{HTML}{F0FFF0}
\definecolor{approachframe}{HTML}{5AAF5A}
\definecolor{designbg}{HTML}{FFF8EE}
\definecolor{designframe}{HTML}{D4A855}
\definecolor{dirtreetxt}{HTML}{2C3E50}
\definecolor{dirtreebg}{HTML}{F7F7F7}
\definecolor{promptbg}{HTML}{F5F0FA}   
\definecolor{promptframe}{HTML}{7E57C2} 

\newtcolorbox{readmebox}[1][]{%
  enhanced, breakable,
  colback=readmebg, colframe=readmeframe,
  fonttitle=\bfseries\ttfamily,
  title={README.md}, #1}

\newtcolorbox{approachbox}[1][]{%
  enhanced, breakable,
  colback=approachbg, colframe=approachframe,
  fonttitle=\bfseries\ttfamily,
  title={plan/approach.md}, #1}

\newtcolorbox{designbox}[1][]{%
  enhanced, breakable,
  colback=designbg, colframe=designframe,
  fonttitle=\bfseries\ttfamily,
  title={plan/design.md}, #1}

\renewtcolorbox{promptbox}[2][]{%
  enhanced,
  unbreakable,
  colback=promptbg,
  colframe=promptframe,
  fonttitle=\bfseries\ttfamily,
  title={#2},
  #1
}

\onecolumn
\subsection{Task Structure}
\label{app:task_structure}

\begin{figure}[htbp]
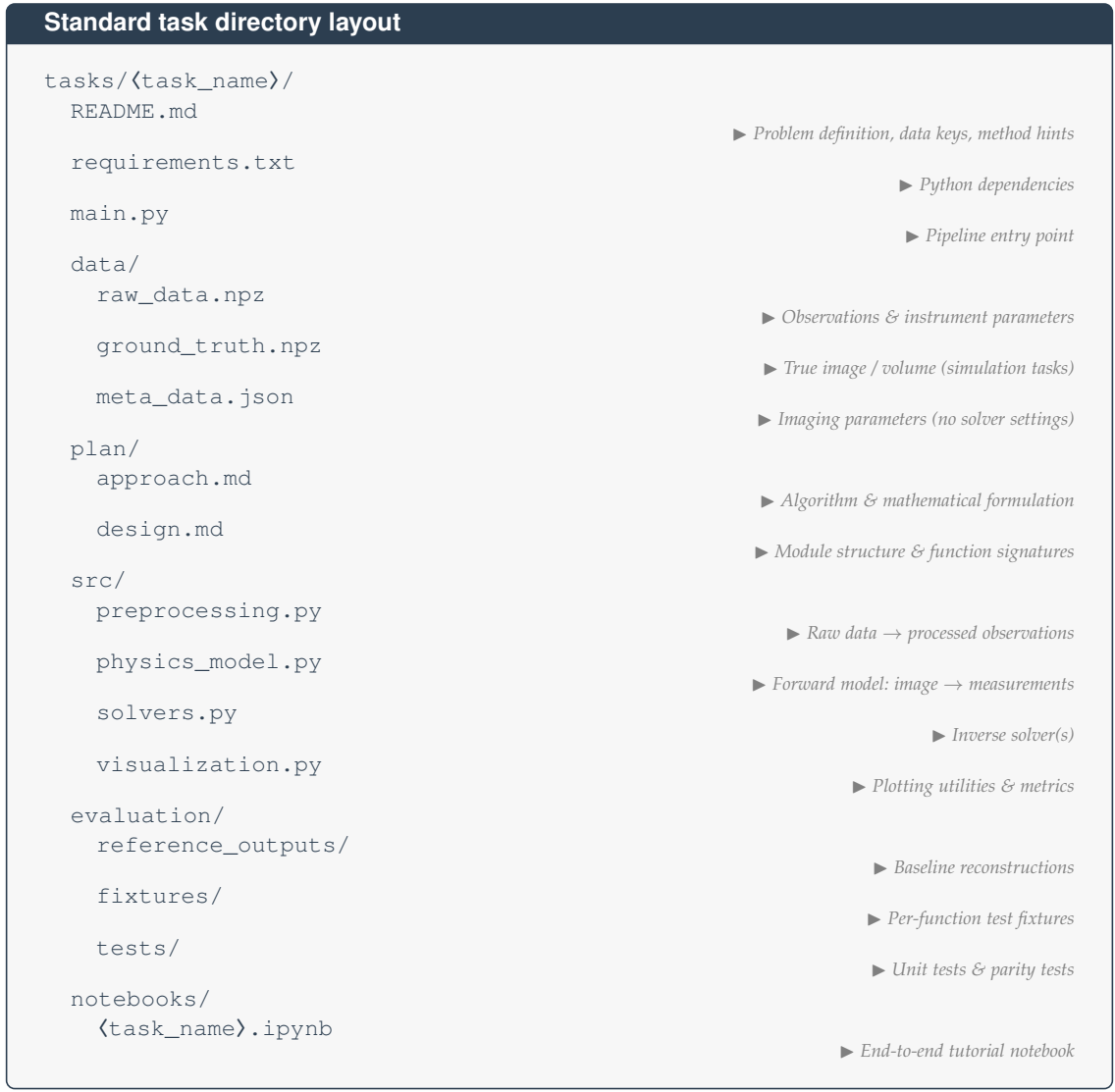

\centering
\begin{tcolorbox}[
  enhanced,
  width=0.82\linewidth,
  colback=dirtreebg,
  colframe=dirtreetxt,
  fontupper=\small\ttfamily\color{dirtreetxt},
  title={\sffamily\bfseries Standard task directory layout},
  boxrule=0.6pt]
\newcommand{\I}{\hspace{0.35cm}}%
\newcommand{\desc}[1]{{\color{gray}\scriptsize$\blacktriangleright$\;\textrm{\itshape #1}}}%
\renewcommand{\arraystretch}{1.15}%
\begin{tabularx}{\linewidth}{@{} l @{\quad} >{\raggedleft\arraybackslash}X @{}}
tasks/\textlangle task\_name\textrangle/ & \\
\I README.md          & \desc{Problem definition, data keys, method hints} \\
\I requirements.txt   & \desc{Python dependencies} \\
\I main.py            & \desc{Pipeline entry point} \\
\I data/              & \\
\I\I raw\_data.npz    & \desc{Observations \& instrument parameters} \\
\I\I ground\_truth.npz & \desc{True image / volume (simulation tasks)} \\
\I\I meta\_data.json  & \desc{Imaging parameters (no solver settings)} \\
\I plan/              & \\
\I\I approach.md      & \desc{Algorithm \& mathematical formulation} \\
\I\I design.md        & \desc{Module structure \& function signatures} \\
\I src/               & \\
\I\I preprocessing.py & \desc{Raw data $\to$ processed observations} \\
\I\I physics\_model.py & \desc{Forward model: image $\to$ measurements} \\
\I\I solvers.py       & \desc{Inverse solver(s)} \\
\I\I visualization.py & \desc{Plotting utilities \& metrics} \\
\I evaluation/        & \\
\I\I reference\_outputs/ & \desc{Baseline reconstructions} \\
\I\I fixtures/        & \desc{Per-function test fixtures} \\
\I\I tests/           & \desc{Unit tests \& parity tests} \\
\I notebooks/         & \\
\I\I \textlangle task\_name\textrangle.ipynb & \desc{End-to-end tutorial notebook} \\
\end{tabularx}
\end{tcolorbox}
\caption{Standard directory layout shared by all 57 tasks.
The \texttt{plan/} directory contains the two planning documents
that define the algorithmic contract an agent must satisfy;
\texttt{src/} holds the reference implementation;
\texttt{evaluation/} provides ground-truth outputs and tests.}
\label{fig:dirtree}
\end{figure}

Every task in Imaging-101 follows a standardized directory layout
(Figure~\ref{fig:dirtree}).
Below we illustrate the three \emph{planning documents}
(\texttt{README.md}, \texttt{plan/approach.md}, and
\texttt{plan/design.md})
using the \texttt{eht\_black\_hole\_original} task, which
reconstructs the simulated radio image (geometric model) of M87* from gain-corrupted
interferometric data.

\paragraph{README.md.}
The README defines the physical problem, the measurement model,
the data layout, and a \emph{Method Hints} section that names
the algorithm family without prescribing implementation details.
Figure~\ref{fig:readme_example} shows the key sections of the
EHT task README.

\begin{figure*}[htbp]
\begin{readmebox}
\small
\textbf{\# EHT Black Hole Imaging with Closure Quantities}

\smallskip
\textit{%
Recover the radio image of a supermassive black hole from
gain-corrupted interferometric data using closure phases and
log closure amplitudes, gain-invariant observables that bypass
the need for antenna-based calibration.}

\smallskip
\textbf{Domain:} Astronomy \quad
\textbf{Keywords:} radio interferometry, compressed sensing \quad
\textbf{Difficulty:} Hard

\medskip\hrule\medskip

\textbf{\#\# Problem Description} \par\smallskip
The van Cittert--Zernike theorem relates the measured visibility to the sky brightness:
$V(u,v) = \iint I(l,m)\,e^{-2\pi i(ul+vm)}\,dl\,dm$.
Discretized on an $N{\times}N$ pixel grid with the DFT matrix $\mathbf{A}$:
\[
  \mathbf{y} = \mathbf{A}\,\mathbf{x},
  \qquad
  A_{m,n} = P(u_m,v_m)\,
  \exp\!\bigl[{+}2\pi i\,(u_m l_n + v_m m_n)\bigr]
\]

\smallskip
\begin{tabular}{@{}llll@{}}
\toprule
Symbol & Description & Size \\
\midrule
$\mathbf{x}$        & Sky brightness image (non-negative) & $N^2{=}4096$ \\
$\mathbf{A}$        & DFT measurement matrix (triangle pulse) & $M{\times}N^2$ \\
$\mathbf{y}_\text{corr}$ & Gain-corrupted complex visibilities  & $M{=}421$ \\
\bottomrule
\end{tabular}

\medskip\hrule\medskip

\textbf{\#\# Data Description} \enspace (\texttt{data/raw\_data.npz}) \par\smallskip
\begin{tabular}{@{}lllp{5.8cm}@{}}
\toprule
Key & Shape & Dtype & Description \\
\midrule
\texttt{vis\_corrupt}       & (421,)   & complex128 & Gain-corrupted visibilities (Jy) \\
\texttt{uv\_coords}         & (421,2)  & float64    & Baseline $(u,v)$ positions \\
\texttt{sigma\_vis}          & (421,)   & float64    & Per-baseline thermal noise $\sigma$ \\
\texttt{cp\_values\_deg}     & (269,)   & float64    & Closure phases from calibrated data \\
\texttt{lca\_values}         & (233,)   & float64    & Log closure amplitudes \\
$\cdots$ & & & \textit{(16 additional keys omitted for brevity)} \\
\bottomrule
\end{tabular}

\medskip\hrule\medskip

\textbf{\#\# Method Hints} \par\smallskip
Reconstruct using \textbf{Regularized Maximum Likelihood (RML)} imaging
directly from closure quantities, following Chael et al.\ (2018).
The objective minimizes a weighted sum of closure chi-squared data terms
and image regularizers (Gull--Skilling entropy, TV) subject to
non-negativity.
Use L-BFGS-B with multi-round optimization (3 rounds of 300 iterations).

\end{readmebox}
\caption{\crfix{Key sections of the \texttt{README.md} for the
\texttt{eht\_black\_hole\_original} task: problem description, data-key table,
and Method Hints.}}
\label{fig:readme_example}
\end{figure*}

\paragraph{plan/approach.md.}
The approach file specifies the complete algorithm: mathematical
formulation, step-by-step solution strategy, and expected
quantitative outcomes.
For the EHT task it states the closure-phase and
log-closure-amplitude chi-squared objectives, the Gull--Skilling
and TV regularizers, and the multi-round L-BFGS-B optimization
schedule.

\begin{figure*}[htbp]
\begin{approachbox}
\small

\textbf{\# Approach}
\textit{Recover a $64{\times}64$ radio image from 421 gain-corrupted
visibilities using closure quantities as gain-invariant observables.}

\medskip\hrule\medskip

\textbf{Mathematical formulation.}\enspace
Forward model: $\mathbf{y}=\mathbf{A}\mathbf{x}$, $\mathbf{A}\in\mathbb{C}^{421\times4096}$.
Station gains corrupt visibilities:
$y_{ij}^{\text{corr}}=g_i\,g_j^*\,y_{ij}^{\text{true}}$.
Closure quantities cancel the gains:
\begin{align*}
  \text{Closure phase:}\quad &
    \phi_C = \arg(V_{ij}\,V_{jk}\,V_{ki}), \\
  \text{Log closure amplitude:}\quad &
    \log\!\mathrm{CA} = \log|V_{ij}|+\log|V_{kl}|-\log|V_{ik}|-\log|V_{jl}|.
\end{align*}

\textbf{Solution strategy.}\enspace
Three RML solvers of increasing gain-robustness:
\begin{enumerate}[nosep,leftmargin=1.2em]
  \item \textbf{Visibility RML}: fits corrupted visibilities directly;
        catastrophically fails under gain corruption.
  \item \textbf{Amplitude + Closure Phase}: partially robust (amplitudes
        remain gain-sensitive).
  \item \textbf{Closure-only RML} (main result, Chael 2018), both data
        terms are fully gain-invariant:\par\smallskip
        $\min_{\mathbf{x}\ge0}
          \;\alpha_\text{CP}\bigl(\chi^2_\text{CP}-1\bigr)
          +\alpha_\text{CA}\bigl(\chi^2_\text{CA}-1\bigr)
          +\alpha_\text{GS}\,S_\text{GS}(\mathbf{x})
          +\alpha_\text{TV}\,\mathrm{TV}(\mathbf{x})$
\end{enumerate}
Optimization: L-BFGS-B with positivity, 3 rounds $\times$ 300 iterations.

\medskip\hrule\medskip

\textbf{Expected results.}\par\smallskip
\begin{tabular}{@{}lcc@{}}
\toprule
Method & Calibrated NRMSE\,/\,NCC & Corrupted NRMSE\,/\,NCC \\
\midrule
Visibility RML       & $\sim$0.26\,/\,0.97 & $\sim$5.0\,/\,0.01 \\
Amplitude + CP       & $\sim$0.70\,/\,0.76 & $\sim$1.7\,/\,0.41 \\
\textbf{Closure-only}& $\sim$0.82\,/\,0.75 & $\sim$\textbf{0.85}\,/\,\textbf{0.72} \\
\bottomrule
\end{tabular}\par\smallskip
\textit{Key insight:} Closure-only imaging sacrifices some fidelity on
perfectly calibrated data but provides robust reconstructions
under realistic gain corruption, the typical EHT regime.

\end{approachbox}
\caption{\crfix{Key sections of \texttt{plan/approach.md} for the
\texttt{eht\_black\_hole\_original} task: mathematical formulation, solution
strategy, and expected results.}}
\label{fig:approach_example}
\end{figure*}

\paragraph{plan/design.md.}
The design file lists every source module with typed function
signatures and a data-flow diagram, forming the \emph{interface
contract} the agent must satisfy.

\onecolumn
\begin{designbox}
\small

\textbf{\# Code Design}\par\medskip

\textbf{File structure.}\par\smallskip
\begin{tabular}{@{}ll@{}}
\texttt{main.py}                 & Pipeline orchestration \\
\texttt{src/preprocessing.py}    & Data loading, closure quantity computation \\
\texttt{src/physics\_model.py}   & DFT forward model + closure $\chi^2$ terms \\
\texttt{src/solvers.py}          & RML solvers (3 variants) \\
\texttt{src/visualization.py}    & Plotting and metrics \\
\texttt{src/generate\_data.py}   & Synthetic data generation \\
\end{tabular}

\medskip\hrule\medskip

\textbf{Key function signatures} (abridged).\par\smallskip
\begin{lstlisting}[
  language=Python,
  basicstyle=\ttfamily\scriptsize,
  keywordstyle=\bfseries\color{blue!70!black},
  commentstyle=\itshape\color{gray},
  columns=fullflexible,
  keepspaces=true,
  xleftmargin=0pt,
  aboveskip=2pt, belowskip=2pt]
# preprocessing.py
def load_observation(data_dir="data") -> dict: ...
def find_triangles(station_ids, n_stations) -> ndarray: ...
def compute_closure_phases(vis, station_ids, tri) -> ndarray: ...
def prepare_data(data_dir="data") -> (obs, closure, meta): ...

# physics_model.py
class ClosureForwardModel:
    def forward(self, image) -> ndarray:  # y = A @ x
    def chisq_cphase_from_uv(imvec, N, psize, uv1, uv2, uv3,
                              clphase_deg, sigma_deg) -> float: ...
    def chisq_logcamp_from_uv(imvec, N, psize, uv1, uv2, uv3,
                               uv4, log_clamp, sigma) -> float: ...

# solvers.py
class ClosureRMLSolver:
    def reconstruct(self, model, obs_data, x0=None) -> ndarray: ...
\end{lstlisting}

\medskip\hrule\medskip

\textbf{Data-flow diagram.}\par\smallskip
\begin{tikzpicture}[
  node distance=0.55cm and 0.3cm,
  every node/.style={font=\scriptsize\ttfamily, align=center},
  block/.style={draw, rounded corners=2pt, fill=designbg,
                minimum height=0.55cm, minimum width=3.0cm,
                font=\scriptsize\sffamily},
  arrow/.style={->, >=stealth, thick, designframe},
]
  \node[block] (data) {raw\_data.npz + meta\_data};
  \node[block, below=of data] (pre)  {preprocessing.py};
  \node[block, below=of pre]  (phys) {physics\_model.py};
  \node[block, below=of phys] (sol)  {solvers.py};
  \node[block, below=of sol]  (vis)  {visualization.py};
  \node[block, below=of vis]  (out)  {output/reconstruction.npy};

  \draw[arrow] (data) -- node[right, font=\scriptsize\sffamily]
    {obs, closure, meta} (pre);
  \draw[arrow] (pre)  -- node[right, font=\scriptsize\sffamily]
    {ClosureForwardModel} (phys);
  \draw[arrow] (phys) -- node[right, font=\scriptsize\sffamily]
    {3 methods $\times$ 2 data states} (sol);
  \draw[arrow] (sol)  -- node[right, font=\scriptsize\sffamily]
    {comparison figures} (vis);
  \draw[arrow] (vis)  -- (out);
\end{tikzpicture}

\end{designbox}
\noindent\begin{minipage}{\linewidth}
\captionsetup{hypcap=false}

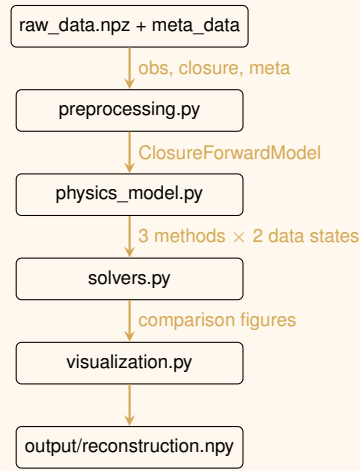
\captionof{figure}{\crfix{Key sections of \texttt{plan/design.md} for the
\texttt{eht\_black\_hole\_original} task: file structure, key function
signatures, and the data-flow diagram.}}
\label{fig:design_example}
\end{minipage}

\medskip\noindent
Together, these three documents fully specify the reconstruction
problem (README), the solution algorithm (approach), and the
implementation blueprint (design).
An agent that faithfully follows this contract can produce a
working pipeline without reading the reference source code in
\texttt{src/}.

\newpage
\section{Evaluation Implementation}
\label{app:implementation}

\subsection{\crfix{Evaluation Inputs and Outputs per Track}}
\label{app:eval_io}

\crfix{To make the evaluation interface explicit (cf.\
Fig.~\ref{fig:main framework of our work}B), Table~\ref{tab:eval_io} lists, for
each track, the artifacts the agent receives, what it must produce, what is
withheld, and how its output is scored. In every track the agent works in an
isolated sandbox; in the end-to-end track the sandbox builds the per-task Python
environment from \texttt{requirements.txt}.}

\begin{table}[ht]
\centering
\footnotesize
\setlength{\tabcolsep}{4pt}
\renewcommand{\arraystretch}{1.15}
\caption{\crfix{Inputs provided, expected output, withheld artifacts, and scoring
for the three evaluation tracks. README files include the Method Hints section.}}
\label{tab:eval_io}
\begin{tabularx}{\linewidth}{@{}l >{\RaggedRight\arraybackslash}X >{\RaggedRight\arraybackslash}X >{\RaggedRight\arraybackslash}X >{\RaggedRight\arraybackslash}X@{}}
\toprule
\textbf{Track} & \textbf{Provided to agent} & \textbf{Agent produces} & \textbf{Withheld} & \textbf{Scoring} \\
\midrule
Planning
  & \texttt{README.md}; \texttt{meta\_data.json}
  & \texttt{approach.md}; \texttt{design.md}
  & reference plan; \texttt{src/}; tests; fixtures; ground truth
  & LLM difference report + two-expert binary pass/fail on 3 dimensions \\
\addlinespace[2pt]
Function-level
  & \texttt{README.md}; reference \texttt{approach.md}\,+\,\texttt{design.md}; target function signature \& docstring; paired unit-test file
  & target function implementation
  & reference function body; test fixtures (expected values); ground truth
  & per-function numerical parity (\texttt{pytest}) \\
\addlinespace[2pt]
End-to-end
  & \texttt{README.md}; \texttt{data/} (\texttt{raw\_data.npz}, \texttt{meta\_data.json}); \texttt{requirements.txt}
  & full \texttt{src/} pipeline + \texttt{output.npz}
  & reference \texttt{src/}; reference plan; hidden \texttt{ground\_truth} / \texttt{baseline\_reference}
  & NCC/NRMSE vs.\ hidden reference \\
\bottomrule
\end{tabularx}
\end{table}

\subsection{Planning Evaluation}
\label{app:planning_eval}

\paragraph{Plan and architecture prompts}
The plan and architecture evaluations are conducted as two separate
stages in \texttt{tests/test\_plan\_architect.py}.
In the planner stage, the \texttt{PlannerAgent} receives only the task
description file specified by \texttt{task\_description\_path}
(\texttt{README.md} in our configuration) through the
\texttt{task\_desc} field, and produces a single planning output saved as
\texttt{plan\_test/plan/<model>/<task>\_plan.md}.
In the architect stage, the \texttt{ArchitectAgent} is evaluated
separately using the same task description together with the task's
reference \texttt{plan/approach.md}; it outputs one architecture file,
saved as \texttt{plan\_test/architect/<model>/<task>\_architect.md}.

\crfix{Both planning stages run as a single-agent ReAct loop on top of the shared
system preamble in Figure~\ref{fig:plan_general_prompt}, which is prepended to every
turn and also governs the function-level evaluation (Section~\ref{app:function_eval}).
On top of this preamble, the \texttt{PlannerAgent} receives the plan-generation
instruction in Figure~\ref{fig:plan_approach_prompt} and writes \texttt{plan/approach.md},
while the \texttt{ArchitectAgent} receives the design-generation instruction in
Figure~\ref{fig:plan_design_prompt} and writes \texttt{plan/design.md}. In both
instructions, the braced fields (\texttt{\{readme\}}, \texttt{\{meta\_data\}},
\texttt{\{approach\}}) are substituted with the task's \texttt{README.md}, data
specification, and reference approach document, respectively.}

\begin{figure}[htbp]
\begin{tcolorbox}[
  enhanced, breakable,
  colback=promptbg, colframe=promptframe,
  fonttitle=\bfseries\sffamily,
  title={Shared single-agent system preamble (planning and function-level)},
  label=lst:plan_general_prompt]
\begin{lstlisting}[
  basicstyle=\ttfamily\scriptsize,
  keywordstyle=\bfseries\color{blue!70!black},
  commentstyle=\itshape\color{gray},
  columns=fullflexible, keepspaces=true,
  xleftmargin=0pt, aboveskip=2pt, belowskip=2pt]
You are a computational imaging expert implementing a
reconstruction pipeline. You work inside a Linux container
at /workspace. Available files: data/ (observations),
README.md (problem description).

You can take these actions:

1) WRITE_FILE - create or overwrite a file
   Action: WRITE_FILE
   Path: <file path relative to /workspace>
   Content:
   <file content - include the ENTIRE file, not a snippet>
   END_CONTENT
2) RUN - execute a shell command
   Action: RUN
   Command: <shell command>
3) READ_FILE - read a file
   Action: READ_FILE
   Path: <file path>
4) DONE - signal that you have finished

Respond in this exact format every turn:
Thought: <your reasoning>
Action: <one of WRITE_FILE | RUN | READ_FILE | DONE>
<action arguments as shown above>

Important rules:
- Write ONE action per response. Wait for the Observation
  before continuing.
- Always write complete files, never partial patches.
- After writing code, run it to verify correctness.
- If errors occur, read the error, fix the code, and re-run.
- Do NOT use invented actions like CHECK, TEST, or ANALYZE.
  To inspect data, write a script and RUN it.
\end{lstlisting}
\end{tcolorbox}
\caption{\crfix{Shared single-agent ReAct system preamble prepended to every turn of
both the planning and function-level evaluations. It fixes the container layout, the
four available actions (\textsc{Write\_File}, \textsc{Run}, \textsc{Read\_File},
\textsc{Done}), the strict \emph{Thought/Action} response format, and the
one-action-per-turn discipline.}}
\label{fig:plan_general_prompt}
\end{figure}

\begin{figure}[htbp]
\begin{tcolorbox}[
  enhanced, breakable,
  colback=promptbg, colframe=promptframe,
  fonttitle=\bfseries\sffamily,
  title={Planner instruction prompt (plan/approach.md)},
  label=lst:plan_approach_prompt]
\begin{lstlisting}[
  basicstyle=\ttfamily\scriptsize,
  keywordstyle=\bfseries\color{blue!70!black},
  commentstyle=\itshape\color{gray},
  columns=fullflexible, keepspaces=true,
  xleftmargin=0pt, aboveskip=2pt, belowskip=2pt]
Read the problem description and data specification below,
then write a solution approach document to plan/approach.md.

The document should include:
1. Problem statement (what we recover, from what
   measurements)
2. Mathematical formulation (forward model equation)
3. Solution strategy (step-by-step algorithmic approach)
4. Expected results (which methods, expected quality)

== Problem Description ==
{readme}

== Data Specification (meta_data) ==
{meta_data}

Begin by reading any additional data files if needed, then
write plan/approach.md.
\end{lstlisting}
\end{tcolorbox}
\caption{\crfix{Instruction prompt given to the \texttt{PlannerAgent} (appended after the
shared preamble of Figure~\ref{fig:plan_general_prompt}). It elicits a four-part solution
approach (problem statement, forward-model formulation, solution strategy, and expected
results) written to \texttt{plan/approach.md}.}}
\label{fig:plan_approach_prompt}
\end{figure}

\begin{figure}[htbp]
\begin{tcolorbox}[
  enhanced, breakable,
  colback=promptbg, colframe=promptframe,
  fonttitle=\bfseries\sffamily,
  title={Architect instruction prompt (plan/design.md)},
  label=lst:plan_design_prompt]
\begin{lstlisting}[
  basicstyle=\ttfamily\scriptsize,
  keywordstyle=\bfseries\color{blue!70!black},
  commentstyle=\itshape\color{gray},
  columns=fullflexible, keepspaces=true,
  xleftmargin=0pt, aboveskip=2pt, belowskip=2pt]
Based on the problem description and the solution approach
below, write a code design document to plan/design.md.

The document should include:
1. File structure (which src/*.py files)
2. Function signatures with full type annotations and
   docstrings
3. Class definitions where appropriate
4. Data flow diagram (text-based)

== Problem Description ==
{readme}

== Solution Approach ==
{approach}

Write plan/design.md with complete function signatures for
every function that needs to be implemented.
\end{lstlisting}
\end{tcolorbox}
\caption{\crfix{Instruction prompt given to the \texttt{ArchitectAgent} (appended after the
shared preamble of Figure~\ref{fig:plan_general_prompt}). Conditioned on the README and the
reference \texttt{plan/approach.md}, it elicits a code-design document (file structure,
fully typed function signatures with docstrings, class definitions, and a text data-flow
diagram) written to \texttt{plan/design.md}.}}
\label{fig:plan_design_prompt}
\end{figure}

\paragraph{LLM-as-judge evaluation}
\crfix{Plan quality is assessed by a two-stage human--LLM pipeline. First, an LLM judge compares each candidate plan against the
reference \texttt{plan/approach.md} and emits a structured \emph{difference
report} that lists agreements and discrepancies along three dimensions:
preprocessing, forward physics modeling, and inverse solver. Second, two domain
experts from the construction pool, blinded to model identity, independently read
the report and the candidate plan and assign a binary pass/fail judgment per
dimension; a dimension is marked correct only when both experts agree, and any
disagreement is counted as a failure. A plan passes \emph{Overall} only if all
three dimensions pass.}
\crfix{Rubric. A dimension is PASS if the plan
meets the criterion below, else FAIL.
\begin{itemize}[nosep,leftmargin=1.2em]
  \item \textbf{Preprocessing:} includes every measurement-to-model
  conversion needed to put the raw data into the form the forward operator
  expects (unit and scale conversions, coordinate/domain transforms, calibration
  or normalization), with no physically required step omitted or applied with the
  wrong convention.
  \item \textbf{Forward physics modeling:} specifies a forward operator
  (and adjoint where required) consistent with the modality's measurement
  equation, with correct operator form, signs, and normalization.
  \item \textbf{Inverse solver:} selects a solver matched to the
  structure of the ill-posed problem (a physics-coupled or appropriately
  regularized method) rather than a generic optimizer applied without regard to
  that structure; a reasonable alternative an expert would accept as capable of
  yielding a correct reconstruction also passes.
\end{itemize}
}

\begin{promptbox}{LLM-judge difference-report prompt}
\footnotesize
\crfix{Compare the difference between the plan checklist and the plan generated by the model. A plan is judged correct only if it contains all the points in the checklist correctly.}
\end{promptbox}
\crfix{To successfully assess the plan, we curated completed checklists for these three dimensions. Table~\ref{tab:preproc_checklist} lists the curated pre-processing checklist extracted from the reference plans.}



\begingroup\scriptsize\setlength{\tabcolsep}{3pt}
\begin{longtable}{>{\raggedright\arraybackslash}p{3.6cm} >{\raggedright\arraybackslash}p{12.8cm}}
\caption{Ground-truth pre-processing checklist: the 32 tasks whose reference plan prescribes genuine data pre-processing, with the required steps. Formulas and parameter values are quoted from the reference plans. \crfix{We restrict this checklist to these 32 of the 57 benchmark tasks because the remaining tasks operate directly on the provided measurements and require no pre-processing step, so there is nothing to score along this dimension.}}\label{tab:preproc_checklist}\\
\toprule
\textbf{Task} & \textbf{Ground-truth pre-processing steps} \\
\midrule
\endfirsthead
\multicolumn{2}{c}{\tablename\ \thetable{} (continued)}\\
\toprule
\textbf{Task} & \textbf{Ground-truth pre-processing steps} \\
\midrule
\endhead
\midrule
\multicolumn{2}{r}{\emph{Continued on next page}}\\
\endfoot
\bottomrule
\endlastfoot
\tsk{confocal-nlos-fk} & 1) Amplitude correction + depth compensation: $\Psi=\sqrt{|\tau|\cdot t^2}$, $t=\mathrm{linspace}(0,1,M)$; 2) zero-pad $\Psi$ to $(2M,2N,2N)$ before 3-D FFT \\ \midrule
\tsk{ct\_fan\_beam} & 1) Parker short-scan weighting $w(\beta,\gamma)$ ($\sin^2$ ramp conditions, $\epsilon=\gamma_{\max}$); 2) cosine/distance pre-weighting $w(\gamma)=D_{\mathrm{sd}}/\sqrt{D_{\mathrm{sd}}^2+\gamma_{\mathrm{iso}}^2}$; 3) ramp-filter construction ($h[0]=1/(8d^2)$, odd-$n$ $-1/(2\pi d n)^2$, Hann cutoff 0.3) --- all sinogram-domain, GT frames as FBP steps \\ \midrule
\tsk{ct\_poisson\_lowdose} & 1) post-log sinogram as input; 2) PWLS weights $W=\mathrm{diag}(I_i)$ from photon counts, normalized to mean 1 \\ \midrule
\tsk{differentiable\_deflectometry} & 1) load calibration + centre-crop $2048^2 \rightarrow 768^2$; 2) lens-mount origin triangulation across 2 cameras; 3) 4-step phase shift $\psi=\arctan((I_3-I_1)/(I_0-I_2))$, periods $T\in\{70,100,110\}$ px; 4) unwrap (skimage) + DC offset removal via integer-period search; 5) convert to mm (0.115 mm/px) + average across periods; 6) subtract reference (no-lens) intersections $\rightarrow$ displacement $\hat{p}$; 7) validity mask from fringe amplitude/modulation (largest connected component) \\ \midrule
\tsk{diffusion\_mri\_dti} & 1) clamp negative/zero signals before log; 2) log transform (linearization $y=\ln S$); 3) tissue masking for fit/metrics \\ \midrule
\tsk{eht\_black\_hole\_UQ} & 1) compute closure phases + log closure amplitudes from visibilities (gain-immune data products) with $\sigma$-weighted losses \\ \midrule
\tsk{eht\_black\_hole\_feature\_extraction\_dynamic} & 1) per-frame closure phase + log closure amplitude computation (gain-invariant, avoids station calibration) \\
\end{longtable}
\endgroup

\normalsize

\crfix{Tables~\ref{tab:fm_name_index} and \ref{tab:fm_formulations} list, for each of the 57 tasks, the ground-truth forward model and its full formulation.}

\begingroup\scriptsize\setlength{\tabcolsep}{3pt}
\begin{longtable}{>{\raggedright\arraybackslash}p{4.2cm} >{\raggedright\arraybackslash}p{12.2cm}}
\caption{Ground-truth forward-model name index for all 57 tasks.}\label{tab:fm_name_index}\\
\toprule
\textbf{Task} & \textbf{Forward model}\\
\midrule
\endfirsthead
\multicolumn{2}{c}{\tablename\ \thetable{} (continued)}\\
\toprule
\textbf{Task} & \textbf{Forward model}\\
\midrule
\endhead
\midrule \multicolumn{2}{r}{\emph{Continued on next page}}\\
\endfoot
\bottomrule
\endlastfoot
\tsk{cars\_spectroscopy} & G-matrix CARS susceptibility (MEG relaxation, collisional narrowing) + Kataoka pump \& slit convolutions\\ \midrule
\tsk{confocal-nlos-fk} & Stolt f-k migration wave model (confocal NLOS as seismic backpropagation)\\ \midrule
\tsk{conventional\_ptychography} & Ptychographic exit wave $\psi=P\cdot O$, far-field/AS intensity\\ \midrule
\tsk{ct\_dual\_energy} & Polychromatic Beer-Lambert two-material model, Poisson counts\\ \midrule
\tsk{ct\_fan\_beam} & Fan-beam Radon transform, flat-detector geometry\\ \midrule
\tsk{ct\_poisson\_lowdose} & Parallel-beam Radon (explicit sparse matrix), Poisson counts $\rightarrow$ PWLS\\ \midrule
\tsk{ct\_sparse\_view} & Discrete Radon transform (skimage \texttt{radon}, circle=True), 30/180 views\\ \midrule
\tsk{differentiable\_deflectometry} & Differentiable ray tracer (Newton aspheric intersection + vector Snell + screen hit)\\
\end{longtable}
\endgroup

\begingroup\scriptsize\setlength{\tabcolsep}{3pt}
\begin{longtable}{>{\raggedright\arraybackslash}p{3.4cm} >{\raggedright\arraybackslash}p{13.0cm}}
\caption{Ground-truth forward-model formulations extracted from the reference plans.}\label{tab:fm_formulations}\\
\toprule
\textbf{Task} & \textbf{Formulation}\\
\midrule
\endfirsthead
\multicolumn{2}{c}{\tablename\ \thetable{} (continued)}\\
\toprule
\textbf{Task} & \textbf{Formulation}\\
\midrule
\endhead
\midrule \multicolumn{2}{r}{\emph{Continued on next page}}\\
\endfoot
\bottomrule
\endlastfoot
\tsk{cars\_spectroscopy} & Per vibrational level $v\in\{0,1\}$, branch $b\in\{Q,O,S\}$: line positions $\nu_j$ from Dunham constants; relaxation matrix $\Gamma$ from MEG law (N$_2$--N$_2$; off-diagonal = rotational energy transfer, diagonal = detailed balance); G-matrix $K=\mathrm{diag}(\nu_j)+i\Gamma$; eigendecompose $K\rightarrow\chi_{\mathrm{RS}}(\nu)$ as sum of complex Lorentzian-like terms weighted by population differences and Raman cross-sections. Add $\chi_{\mathrm{NR}}$ (scaled by mole fraction, pressure, $T$); $I\propto|\chi_{\mathrm{total}}|^2$; convolve Gaussian pump linewidth (Kataoka approx.); convolve asymmetric-Gaussian slit; downsample to experimental grid.\\ \midrule
\tsk{confocal-nlos-fk} & $T=\mathrm{fftshift}(\mathrm{FFT3D}(\Psi_{\mathrm{padded}}))$; Stolt dispersion $k_{z,\mathrm{new}}=\sqrt{\mathrm{scale}^2(k_x^2+k_y^2)+k_f^2}$, $\mathrm{scale}=(N\cdot\mathrm{range\_m})/(M\cdot\mathrm{width}\cdot 4)$, $\mathrm{range\_m}=M\,c\cdot\mathrm{bin\_resolution}$, $\mathrm{width}=\mathrm{wall\_size}/2$; bilinear interpolation (real/imag separately); Jacobian $|k_f|/\max(k_{z,\mathrm{new}},\varepsilon)$; zero $k_f\le 0$; $\mathrm{vol}=|\mathrm{iFFT3D}(\mathrm{ifftshift}(T))|^2$.\\ \midrule
\tsk{conventional\_ptychography} & $\psi_j(r)=P(r)\cdot O(r-r_j)$; $I_j(q)=|D[\psi_j](q)|^2$, $D$ = Angular Spectrum or Fraunhofer propagator.\\ \midrule
\tsk{ct\_dual\_energy} & $\nu_m(a)=\sum_E I_{0,m}(E)\exp\!\big(-\sum_k a_k\mu_k(E)\big)\Delta E$ for the two basis materials; Poisson NLL $L(a)=\sum_m[\nu_m(a)-g_m\ln\nu_m(a)]$.\\ \midrule
\tsk{ct\_fan\_beam} & $t_{\mathrm{proj}}=t\,(D_{\mathrm{sd}}+D_{\mathrm{dd}})/(D_{\mathrm{sd}}-s)$, $t=x\cos\beta+y\sin\beta$, $s=-x\sin\beta+y\cos\beta$; FBP $f=\int(1/U^2)\,Q(\beta,t_{\mathrm{proj}})\,d\beta$, $U=D_{\mathrm{sd}}-s$; iterative variant $\min\tfrac12\|Ax-b\|^2+\lambda\,\mathrm{TV}$.\\
\end{longtable}
\endgroup

\crfix{Tables~\ref{tab:inv_index} and \ref{tab:inv_entries} report, for the inverse-solver stage, the ground-truth loss/solver index for all 57 tasks, the full per-task loss formulations and solver specifics.}

\begingroup\scriptsize\setlength{\tabcolsep}{3pt}
\begin{longtable}{>{\raggedright\arraybackslash}p{3.4cm} >{\raggedright\arraybackslash}p{6.2cm} >{\raggedright\arraybackslash}p{6.6cm}}
\caption{Ground-truth inverse-solver index for all 57 tasks: loss/objective type and chosen solver. N/A marks tasks whose reference method has no explicit objective (direct/analytic methods, projection phase retrieval, plug-and-play implicit priors, pretrained inference, generative sampling).}\label{tab:inv_index}\\
\toprule
\textbf{Task} & \textbf{Loss} & \textbf{Solver}\\ \midrule
\endfirsthead
\multicolumn{3}{c}{\tablename\ \thetable{} (continued)}\\
\toprule
\textbf{Task} & \textbf{Loss} & \textbf{Solver}\\ \midrule
\endhead
\midrule \multicolumn{3}{r}{\emph{Continued on next page}}\\
\endfoot
\bottomrule
\endlastfoot
\tsk{cars\_spectroscopy} & NLS $\sum(I_{\mathrm{meas}}-I_{\mathrm{model}})^2$, box bounds & \texttt{scipy} \texttt{least\_squares} TRF, \texttt{max\_nfev}=100 \\ \midrule
\tsk{confocal-nlos-fk} & N/A (direct migration) & Stolt f-k migration (FFT $\rightarrow$ remap $\rightarrow$ iFFT) \\ \midrule
\tsk{conventional\_ptychography} & N/A (projection phase retrieval) & mPIE: ePIE + momentum every $T{=}50$, $\eta=0.7$ \\ \midrule
\tsk{ct\_dual\_energy} & Poisson NLL per bin & Gauss--Newton w/ analytic $2\times2$ Hessian, 20 iters $\rightarrow$ FBP \\ \midrule
\tsk{ct\_fan\_beam} & $\tfrac12\|Ax-b\|^2+\lambda\,\mathrm{TV}$ ($\lambda=0.005$) / N/A for FBP & Parker-weighted fan FBP + Chambolle--Pock PDHG (150 it) \\ \midrule
\tsk{ct\_poisson\_lowdose} & PWLS $+\,\beta\cdot$q-GGMRF ($\beta=0.15$), $x\ge0$ & Projected FISTA, 250 it, power-iteration step \\ \midrule
\tsk{ct\_sparse\_view} & $\tfrac12\|Af-g\|^2+\lambda\,\mathrm{TV}_{\mathrm{iso}}$ ($\lambda=0.01$), $f\ge0$ & Chambolle--Pock PDHG (semi-implicit data term), 300 it, FBP init \\ \midrule
\tsk{differentiable\_deflectometry} & $\sum\|p(u;\theta)-\hat p(u)\|^2$, 8 params & Levenberg--Marquardt, $R=\operatorname{diag}(J^{\mathsf T}J)$, autodiff Jacobian \\
\end{longtable}
\endgroup

\begingroup\scriptsize\setlength{\tabcolsep}{3pt}
\begin{longtable}{>{\raggedright\arraybackslash}p{3.2cm} >{\raggedright\arraybackslash}p{13.2cm}}
\caption{Ground-truth inverse-solver checklist: full loss formulations and solver specifics extracted from the reference plans.}\label{tab:inv_entries}\\
\toprule
\textbf{Task} & \textbf{Loss / Solver}\\ \midrule
\endfirsthead
\multicolumn{2}{c}{\tablename\ \thetable{} (continued)}\\
\toprule
\textbf{Task} & \textbf{Loss / Solver}\\ \midrule
\endhead
\midrule \multicolumn{2}{r}{\emph{Continued on next page}}\\
\endfoot
\bottomrule
\endlastfoot
\tsk{cars\_spectroscopy} & \emph{Loss:} $\sum_i\bigl(I_{\mathrm{meas}}(\nu_i)-I_{\mathrm{model}}(\nu_i;T,x_{\mathrm{mol}},\nu_{\mathrm{shift}},\mathrm{slit})\bigr)^2$, no regularization; bounds $T\in[300,3500]$ K, $x_{\mathrm{mol}}\in[0.1,1.0]$, $\nu_{\mathrm{shift}}\in[-5,5]$; stretch fixed at 1.0. \par \emph{Solver:} \texttt{scipy.optimize.least\_squares}, trust-region reflective (TRF --- load-bearing for bounds), \texttt{max\_nfev}=100; init $T=2000$ K (deliberately offset from true 2400 K).\\ \midrule
\tsk{confocal-nlos-fk} & \emph{Loss:} N/A --- direct Stolt migration, no objective, no free parameters. \par \emph{Solver:} Stolt f-k: pre-weight $\sqrt{|\tau|t^2}$ $\rightarrow$ pad $(2M,2N,2N)$ $\rightarrow$ FFT3D+fftshift $\rightarrow$ remap $k_z=\sqrt{\mathrm{scale}^2(k_x^2+k_y^2)+k_f^2}$, bilinear (order $=1$) interpolation of real/imag separately $\rightarrow$ Jacobian $|k_f|/\max(k_z,\varepsilon)$, zero $k_f\le0$ $\rightarrow$ iFFT3D, $|\cdot|^2$, unpad.\\ \midrule
\tsk{conventional\_ptychography} & \emph{Loss:} N/A --- projection-type phase retrieval; intensity constraint $\tilde\psi=\sqrt{I}\,\Psi/|\Psi|$; monitored error $E<0.05$. \par \emph{Solver:} mPIE (Maiden 2017): ePIE updates normalized by $\max(|P|^2)$/$\max(|O|^2)$; momentum on full object every $T=50$ iters, $v=\eta v+\Delta O$ with $\eta=0.7$; L2 regularization (aleph $=10^{-2}$) on even rounds only; 7 rounds $\times$ 50 iters.\\ \midrule
\tsk{ct\_dual\_energy} & \emph{Loss:} per-bin Poisson NLL $L(a)=\sum_m[\nu_m(a)-g_m\ln\nu_m(a)]$, non-negativity clamp, no regularization. \par \emph{Solver:} Gauss--Newton/Newton per bin with exact analytic gradient and $2\times2$ Hessian (analytic inversion --- load-bearing), 20 iters, init $a=10^{-6}$; then iradon FBP on material sinograms (divide by pixel size 0.1 cm).\\
\end{longtable}
\endgroup

\paragraph{Worked examples}
To make the pass/fail rubric concrete, we trace three judgments produced by
exactly the procedure above. Each plan was written by the \emph{same} model
(GPT-5.4) and both were scored by the two domain experts blind to model
identity; each plan is decided dimension by dimension against the curated
checklists in Tables~\ref{tab:preproc_checklist}, \ref{tab:fm_name_index}--%
\ref{tab:fm_formulations}, and \ref{tab:inv_index}--\ref{tab:inv_entries}, and a
plan is accepted (\emph{Overall} PASS) only when all three dimensions pass and
the two experts agree on each. Table~\ref{tab:plan_judgment_accept} shows an
accepted plan; Tables~\ref{tab:plan_judgment_reject}
and~\ref{tab:plan_judgment_reject_fpm} show two rejected plans that fail in
different ways: one that identifies the correct method but omits a required
pre-processing step, and one that misidentifies the core physics (both the
forward model and the inverse solver).

\begin{table}[H]
\centering\scriptsize\setlength{\tabcolsep}{3pt}
\caption{\textbf{Accepted} plan judgment: GPT-5.4 on \tsk{ct\_poisson\_lowdose}
(low-dose CT). Each dimension is judged against the curated checklist
(Tables~\ref{tab:preproc_checklist}, \ref{tab:fm_name_index}, and
\ref{tab:inv_index}); both experts agreed on every dimension.}
\label{tab:plan_judgment_accept}
\begin{tabular}{>{\raggedright\arraybackslash}p{2.1cm} >{\raggedright\arraybackslash}p{4.9cm} >{\raggedright\arraybackslash}p{6.6cm} >{\centering\arraybackslash}p{1.3cm}}
\toprule
\textbf{Dimension} & \textbf{Ground-truth requirement (checklist)} & \textbf{What the candidate plan specified} & \textbf{Verdict}\\
\midrule
Pre-processing & Post-log sinogram as input; PWLS weights $W=\mathrm{diag}(I_i)$ from photon counts, mean-normalized. & Adopts the post-log model $y_i=-\log(I_i/I_0)$ and uses the supplied photon-count weights $W=\mathrm{diag}(I_i)$ in the data term. & \textbf{PASS}\\
\midrule
Forward model & Parallel-beam Radon system matrix with a matched adjoint. & Discrete parallel-beam Radon operator $A$ with exact transpose $A^{\top}$, reused in the data-fidelity gradient $A^{\top}W(Ax-y)$. & \textbf{PASS}\\
\midrule
Inverse solver & PWLS objective $+\,\beta\,$q-GGMRF prior with a non-negativity constraint, minimized by a gradient method. & PWLS $+$ q-GGMRF ($p{=}1.2$, $q{=}2.0$, $T{=}1.0$) with $x\ge0$, minimized by projected (FISTA) gradient descent. & \textbf{PASS}\\
\midrule
\multicolumn{4}{>{\raggedright\arraybackslash}p{15.4cm}}{\textbf{Overall: ACCEPTED.} All three dimensions pass and the two experts agree on each, so the plan is accepted.}\\
\bottomrule
\end{tabular}
\end{table}

\begin{table}[H]
\centering\scriptsize\setlength{\tabcolsep}{3pt}
\caption{\textbf{Rejected} plan judgment: GPT-5.4 on \tsk{exoplanet\_imaging}
(high-contrast angular differential imaging). The plan identifies the correct
method (KLIP$+$ADI) and reproduces the forward model and inverse solver, but
omits a required pre-processing step, so it fails that dimension and is rejected
overall.}
\label{tab:plan_judgment_reject}
\begin{tabular}{>{\raggedright\arraybackslash}p{2.1cm} >{\raggedright\arraybackslash}p{4.9cm} >{\raggedright\arraybackslash}p{6.6cm} >{\centering\arraybackslash}p{1.3cm}}
\toprule
\textbf{Dimension} & \textbf{Ground-truth requirement (checklist)} & \textbf{What the candidate plan specified} & \textbf{Verdict}\\
\midrule
Pre-processing & Mean-subtract the flattened frames; replace NaNs/bad pixels; apply the inner-working-angle mask at $4$\,px. & Mean-centers the frame matrix and builds the inner-working-angle radial mask, but specifies \emph{no} NaN/bad-pixel replacement. & \textbf{FAIL}\\
\midrule
Forward model & ADI model: detector-fixed quasi-static speckles plus a companion rotating with the parallactic angle $\theta_k$. & States $T_k=I_\psi+A(\mathcal{R}_{-\theta_k}x)$ with parallactic-angle derotation; matches the reference forward model. & \textbf{PASS}\\
\midrule
Inverse solver & KLIP low-rank speckle subtraction: SVD top-$K$, per-frame projection-subtraction, derotation, temporal combine. & Full-frame KLIP via SVD top-$K$ ($K\in\{5,\dots,20\}$), per-frame subtraction, derotation by $-\theta_k$, mean combine. & \textbf{PASS}\\
\midrule
\multicolumn{4}{>{\raggedright\arraybackslash}p{15.4cm}}{\textbf{Overall: REJECTED.} The plan names the correct method and reproduces the forward model and solver, but omits the mandatory NaN-replacement step. Because unreplaced bad pixels propagate through the covariance/SVD and corrupt the entire KL basis, both experts mark pre-processing FAIL; since \emph{Overall} requires all three dimensions to pass, the plan is rejected. This case shows why the audit is step-level: naming the right algorithm is necessary but not sufficient.}\\
\bottomrule
\end{tabular}
\end{table}

\begin{table}[H]
\centering\scriptsize\setlength{\tabcolsep}{3pt}
\caption{\textbf{Rejected} plan judgment, core-physics failure: GPT-5.4 on
\tsk{fourier\_ptychography} (Fourier ptychographic microscopy). In contrast to
Table~\ref{tab:plan_judgment_reject}, here \emph{both} core-physics dimensions
fail; this task is in fact failed by all seven evaluated models on both the
forward-model and inverse-solver dimensions. Fourier ptychography is not part of
the pre-processing checklist (the task operates directly on the raw
low-resolution intensity stack), so that dimension is not scored.}
\label{tab:plan_judgment_reject_fpm}
\begin{tabular}{>{\raggedright\arraybackslash}p{2.1cm} >{\raggedright\arraybackslash}p{4.9cm} >{\raggedright\arraybackslash}p{6.6cm} >{\centering\arraybackslash}p{1.3cm}}
\toprule
\textbf{Dimension} & \textbf{Ground-truth requirement (checklist)} & \textbf{What the candidate plan specified} & \textbf{Verdict}\\
\midrule
Pre-processing & Not applicable: the task uses the raw low-resolution intensity stack directly and is not in the pre-processing checklist. & No pre-processing step required. & \textbf{N/A}\\
\midrule
Forward model & FPM pupil-bandpass shifted-spectrum intensity model $I_j=|\mathrm{IFT}\{\tilde{P}(q)\,\tilde{O}(q-q_j)\}|^2$, with the pupil $\tilde{P}$ an \emph{unknown} (aberrated) function recovered jointly with the object. & Writes the correct intensity equation but fixes $\tilde{P}$ to an \emph{ideal circular aperture} (pupil update disabled by default), so the unknown aberrated pupil is never modeled. & \textbf{FAIL}\\
\midrule
Inverse solver & qNewton/PIE update (explicitly \emph{not} ePIE) with mandatory joint pupil recovery and a bright-field-first, NA-sorted LED ordering. & ePIE-style spectral updates with optional/disabled pupil recovery and an arbitrary low-to-high or randomized LED order. & \textbf{FAIL}\\
\midrule
\multicolumn{4}{>{\raggedright\arraybackslash}p{15.4cm}}{\textbf{Overall: REJECTED.} The plan names the right method (``Fourier ptychography phase retrieval'') and even writes the correct forward intensity equation, yet adopts ePIE in place of the reference qNewton/PIE solver and fixes the pupil instead of recovering it---and ePIE is known to concentrate updates in the bright-field region and cap the achievable super-resolution, exactly the failure the reference solver avoids. Both core-physics dimensions therefore fail (as they do for all seven models on this task), so the plan is rejected. This complements the previous example: a plan can fail not only by omitting a step but by misidentifying the core physics, so naming the right method family is not sufficient.}\\
\bottomrule
\end{tabular}
\end{table}

\definecolor{funcbg}{HTML}{F5F0FF}
\definecolor{funcframe}{HTML}{7B5EA7}
\definecolor{reactbg}{HTML}{FFFDF0}
\definecolor{reactframe}{HTML}{C49B2C}
\definecolor{codebg}{HTML}{F8F8F8}

\subsection{Function-Level Evaluation}
\label{app:function_eval}

The function-level evaluation measures whether an LLM agent can implement
a single source module (e.g.\ \texttt{preprocessing.py} or
\texttt{physics\_model.py}) given the full task specification and a suite
of unit tests.
Unlike end-to-end evaluation , the agent
does not need to orchestrate the complete pipeline; all non-target
modules are pre-seeded with reference implementations so that test
failures are isolated to the target module alone.

\paragraph{Code generation prompt.}
The agent receives a structured prompt assembled from five context
blocks:

\begin{enumerate}[nosep,leftmargin=1.2em]
  \item \textbf{Problem Description}: the task README describing the
        physical problem, measurement model, and data layout.
  \item \textbf{Solution Approach}: the algorithmic strategy from
        \texttt{plan/approach.md} (mathematical formulation, solver
        schedule, expected quality).
  \item \textbf{Code Design}: the full set of typed function
        signatures from \texttt{plan/design.md}, establishing the
        interface contract.
  \item \textbf{Target Function}: the specific source file to
        implement (e.g.\ \texttt{preprocessing.py}).
  \item \textbf{Unit Tests}: the verbatim content of
        \texttt{evaluation/tests/test\_\{module\}.py}, defining the
        acceptance criteria.
\end{enumerate}

\noindent
\crfix{The agent runs under the same shared single-agent ReAct system preamble used for the
planning evaluation (Figure~\ref{fig:plan_general_prompt}), which defines the four available
actions: \textsc{Write\_File}, \textsc{Run}, \textsc{Read\_File}, and \textsc{Done}, 
and requires a strict \emph{Thought/Action} response on every turn.} The five context blocks
above are then assembled into the instruction prompt of Figure~\ref{fig:func_prompt}, where the
braced fields (\texttt{\{readme\}}, \texttt{\{approach\}}, \texttt{\{design\}},
\texttt{\{target\_function\}}, \texttt{\{test\_file\_content\}}) are substituted with the
corresponding task artifacts.

\begin{figure}[htbp]
\begin{tcolorbox}[
  enhanced, breakable,
  colback=funcbg, colframe=funcframe,
  fonttitle=\bfseries\sffamily,
  title={Function-level instruction prompt},
  label=lst:func_prompt]
\begin{lstlisting}[
  basicstyle=\ttfamily\scriptsize,
  keywordstyle=\bfseries\color{blue!70!black},
  commentstyle=\itshape\color{gray},
  columns=fullflexible, keepspaces=true,
  xleftmargin=0pt, aboveskip=2pt, belowskip=2pt]
Implement the function specified below. Context is provided
so you understand the overall pipeline, but you only need to
implement the target function.

== Problem Description ==
{readme}

== Solution Approach ==
{approach}

== Code Design (full signatures) ==
{design}

== Target Function ==
{target_function}

== Unit Tests (your implementation will be tested against these) ==
{test_file_content}

Steps:
1. Read any relevant data/fixture files to understand inputs
   and outputs.
2. Write the source file containing the target function.
3. Run the unit tests to verify.
4. Fix any failures, then signal DONE.
\end{lstlisting}
\end{tcolorbox}
\caption{\crfix{Function-level instruction prompt (appended after the shared system preamble
of Figure~\ref{fig:plan_general_prompt}). The agent is asked to implement a single target
module given five context blocks (the problem description, solution approach, full typed
code design, the target source file, and the unit tests it will be graded against) and to
iterate write/run/fix until the tests pass.}}
\label{fig:func_prompt}
\end{figure}

\paragraph{Dependency isolation.}
Before the agent loop begins, the harness copies reference
implementations of every \texttt{src/*.py} module \emph{except} the
target into the sandbox.
For example, when the target is \texttt{physics\_model.py}, the files
\texttt{preprocessing.py}, \texttt{solvers.py}, and
\texttt{visualization.py} are seeded with their ground-truth versions.
This ensures that cross-module imports succeed (e.g.\
\texttt{test\_physics\_model.py} importing from
\texttt{src.preprocessing}) and that every test failure is
attributable solely to the agent's implementation.

\begin{figure}[htbp]
\centering
\begin{tikzpicture}[
  node distance=0.8cm and 1.6cm,
  every node/.style={font=\small\sffamily, align=center},
  block/.style={draw, rounded corners=3pt, fill=funcbg,
                minimum height=0.7cm, minimum width=3.6cm,
                font=\small\sffamily},
  decision/.style={draw, diamond, aspect=2.2, fill=reactbg,
                   inner sep=1pt, font=\small\sffamily},
  arrow/.style={-{Stealth[length=5pt]}, thick, funcframe},
]
  \node[block] (llm)   {LLM Call\\(message history)};
  \node[block, below=of llm] (parse) {Parse Thought\\+ Action};
  \node[decision, below=1.0cm of parse] (done) {Action\\= DONE?};
  \node[block, right=2.0cm of done] (exec)
    {Execute in Sandbox\\(write / run / read)};
  \node[block, above=of exec] (obs)
    {Append Observation\\to History};
  \node[decision, below=1.0cm of done] (gate)
    {pytest\\ran?};
  \node[block, below=0.8cm of gate] (accept)
    {Accept DONE\\$\to$ Score};
  \node[block, left=2.0cm of gate] (reject)
    {Reject: force\\test execution};

  \draw[arrow] (llm) -- (parse);
  \draw[arrow] (parse) -- (done);
  \draw[arrow] (done) -- node[above] {no} (exec);
  \draw[arrow] (exec) -- (obs);
  \draw[arrow] (obs) -- (llm);
  \draw[arrow] (done) -- node[right] {yes} (gate);
  \draw[arrow] (gate) -- node[right] {yes} (accept);
  \draw[arrow] (gate) -- node[above] {no} (reject);
  \draw[arrow] (reject) |- (obs);
\end{tikzpicture}
\caption{The ReAct agent loop for function-level evaluation.
The agent iterates between LLM reasoning and sandbox execution.
A gate check prevents premature termination: the agent must execute
\texttt{pytest} at least once before \textsc{Done} is accepted.}
\label{fig:react_loop}
\end{figure}
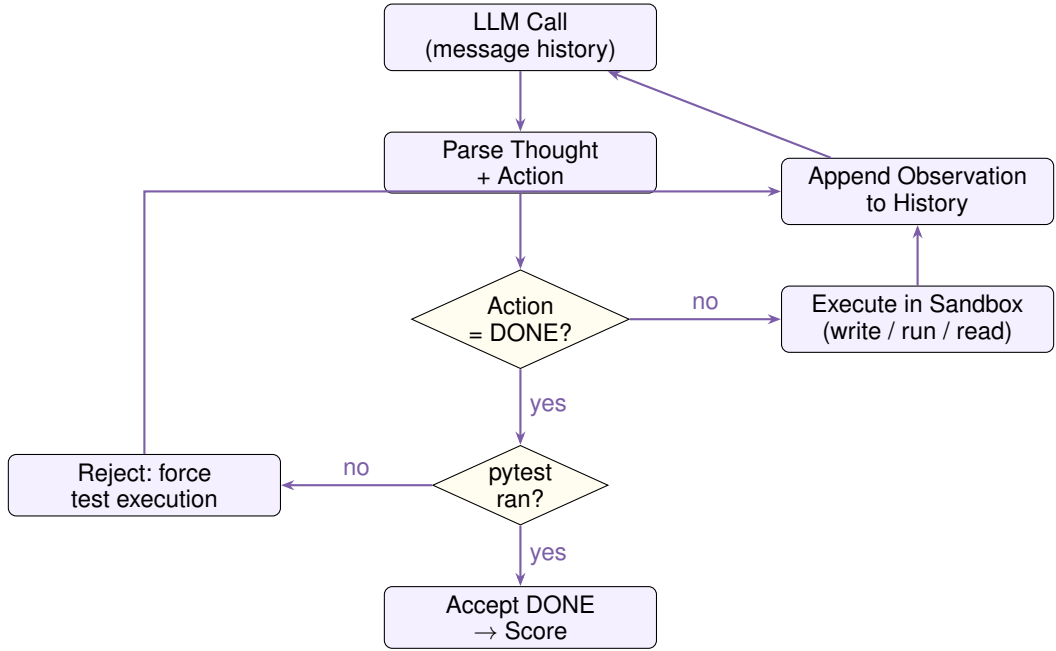

\paragraph{ReAct reflection loop.}
Code generation proceeds via the ReAct (Reasoning + Acting) loop
depicted in Figure~\ref{fig:react_loop}.
The loop runs for at most $K$ iterations
(default $K{=}10$) and proceeds as follows:

\begin{enumerate}[nosep,leftmargin=1.2em]
  \item Call the LLM with the current message history.
  \item Parse the \emph{Thought} (chain-of-thought reasoning) and
        \emph{Action} (one of the four primitives) from the response.
  \item Execute the action inside the sandboxed workspace (write a
        file, run a shell command, or read a file) and collect the
        observation.
  \item Append the observation to the message history.
  \item If the agent signals \textsc{Done}, apply a \emph{gate check}:
        verify that \texttt{pytest} was actually executed at least once.
        If not, reject the signal and instruct the agent to run the
        test suite before finishing.
  \item Repeat until \textsc{Done} is accepted or $K$ iterations are
        exhausted.
\end{enumerate}

\paragraph{Context management.}
Three mechanisms keep the conversation within the LLM's context budget
as the agent iterates:

\begin{itemize}[nosep,leftmargin=1.2em]
  \item \textbf{Output truncation.}
        Command outputs and file reads are capped at 8\,K characters.
        When exceeded, the first 3.5\,K and last 3.5\,K characters are
        retained with an elision marker in between.

  \item \textbf{Context compaction.}
        When the total conversation length exceeds 60\,K characters, an
        auxiliary LLM call summarizes the middle messages into a
        structured digest with five sections: \emph{Current State},
        \emph{Key Findings}, \emph{Error History}, \emph{What Works},
        and \emph{What Remains}.
        The summary replaces the original middle messages, preserving
        the system prompt and initial task prompt verbatim.

  \item \textbf{Sliding window.}
        The system prompt, initial user prompt, and (if present) the
        compact summary are always retained.
        The most recent messages fill the remaining budget, with a hard
        cap of 90\,K total characters (${\approx}$22\,K tokens).
        If the budget is still exceeded, the largest non-head messages
        are individually truncated.
\end{itemize}

\paragraph{Automatic scoring.}
After the agent loop terminates (either by \textsc{Done} or iteration
exhaustion), the harness invokes \texttt{pytest} inside the sandbox:

\begin{tcolorbox}[
  enhanced,
  colback=codebg, colframe=gray!60,
  boxrule=0.4pt,
  left=4pt, right=4pt, top=2pt, bottom=2pt]
\ttfamily\small
python -m pytest evaluation/tests/test\_\{module\}.py -v --tb=short
\end{tcolorbox}

\noindent
The output is parsed to extract per-test \textsc{Passed} / \textsc{Failed} /
\textsc{Error} counts.
The primary metric is the \emph{test pass rate}:
\[
  \texttt{test\_pass\_rate}
  \;=\;
  \frac{\text{tests passed}}{\text{tests total}}.
\]
This provides a fine-grained, deterministic signal: unlike
reconstruction-quality metrics (NRMSE, NCC) that require subjective
thresholds, each unit test yields an unambiguous binary verdict.
The test suite covers both correctness (output values match reference
fixtures within tolerance) and interface compliance (function signatures,
return types, array shapes).

\medskip\noindent
Table~\ref{tab:func_summary} summarizes the key parameters of the
function-level evaluation pipeline.

\begin{table}[htbp]
\centering
\small
\caption{Function-level evaluation pipeline parameters.}
\label{tab:func_summary}
\begin{tabular}{@{}ll@{}}
\toprule
\textbf{Parameter} & \textbf{Value} \\
\midrule
Max agent iterations          & \crfix{10} (default) \\
Output truncation             & 8\,K chars (3.5\,K head + 3.5\,K tail) \\
Compaction trigger            & 60\,K chars total \\
Sliding-window hard cap       & 90\,K chars (${\approx}$22\,K tokens) \\
Available actions             & \textsc{Write\_File}, \textsc{Run}, \textsc{Read\_File}, \textsc{Done} \\
DONE gate (function mode)     & \texttt{pytest} must have run $\ge$1 time \\
Primary metric                & test\_pass\_rate $\in [0, 1]$ \\
Dependency isolation          & Non-target \texttt{src/*.py} seeded with reference \\
\bottomrule
\end{tabular}
\end{table}

\subsection{End-to-End Multi-Agent Coding Workflow}
\label{app:e2e_eval}

\paragraph{Overview.}
Our end-to-end evaluation is implemented as a five-role multi-agent coding loop consisting of a
\texttt{Planner}, \texttt{Critic}, \texttt{Architect}, \texttt{Coder}, and \texttt{Judge}.
For each task, the harness constructs a task-specific sandbox, copies only the observable inputs under
\texttt{data/}, moves hidden references (\texttt{ground\_truth.npz} or \texttt{baseline\_reference.npz})
to the sandbox root so that the solver cannot read them directly, and generates auxiliary metadata files
such as \texttt{data/data\_info.json}, \texttt{data/gt\_keys.json}, and \texttt{data/output\_keys.json}.
The agents then synthesize a self-contained \texttt{solver.py}, execute it in the per-task Python
environment defined by \texttt{requirements.txt}, and save predictions to \texttt{output.npz}.
The final prediction is evaluated against the hidden reference using the task-specific boundaries in
\texttt{evaluation/metrics.json}.

\paragraph{Prompt structure.}
The prompt presented to the multi-agent system is assembled from:
(1)~the task description \texttt{README.md};
(2)~\texttt{data/meta\_data.json};
(3)~a machine-generated \texttt{data\_layout} summary listing all accessible files in \texttt{data/},
together with exact \texttt{.npz} key names, shapes, and dtypes when available;
(4)~the package list from the task-specific environment; and
(5)~role-specific system prompts for planning, code-structure design, implementation, and failure diagnosis.

\onecolumn
\begin{promptbox}{Planner system prompt}
\footnotesize

You are a Principal Scientist in Computational Imaging and Inverse Problems.

Your goal is to formulate a rigorous mathematical and algorithmic plan to solve the user's task.

\textbf{Guidelines.}

1. \textbf{Mathematical modeling.} Explicitly define the forward model $y = A(x) + n$. What is $A$? What is the noise $n$?

2. \textbf{Method selection.} Choose the most robust solver.
\begin{itemize}
\item For linear problems with constraints: consider ADMM, primal--dual, or FISTA.
\item For nonlinear / complex priors: consider deep unrolling (algorithm unrolling) or end-to-end CNNs (UNet/ResNet).
\item \emph{Constraint}: prefer stability and standard implementation over experimental papers.
\end{itemize}

3. \textbf{Dimensionality awareness.} Mentally check the input and output shapes. If the input is $(B,C,H,W)$, ensure the operations preserve or transform dimensions correctly.

4. \textbf{Self-contained.} The solution must only use files from the \texttt{data/} directory for input data and \texttt{data/meta\_data.json} for physical parameters. For \texttt{.npz} files, load with \texttt{np.load()} and access the correct key. No external files (\texttt{.tif}, \texttt{.yaml}, \texttt{.h5}, \texttt{.csv}, \texttt{.mat}) should be assumed. The data layout and available packages will be provided in the context.

5. \textbf{Simplicity first.} Prefer well-understood classical algorithms that are straightforward to implement in $<200$ lines of Python. Avoid complex deep-learning architectures unless explicitly needed.

6. \textbf{Hyperparameter scale awareness.}
\begin{itemize}
\item Check \texttt{meta\_data.json} for output scale hints.
\item \emph{Optimizer selection}: choose the simplest optimizer that fits the problem. For differentiable physics forward models with autograd support, prefer simple gradient descent. For classical problems with analytic gradients, consider L-BFGS-B. Justify your choice.
\item \emph{Learning rate}: calibrate \texttt{lr} by checking that \texttt{lr * max(|grad|)} produces reasonable per-step changes (typically \texttt{1--10\%} of expected output scale). Do NOT default to \texttt{lr=1e-3} without justification --- physics problems often need much larger \texttt{lr}.
\item \emph{Positivity constraints}: for strictly positive unknowns, consider reparameterization (e.g., log-space) or simple clamping after each step.
\item \emph{Regularization policy}: default to minimal regularization. Only add regularization if the task description explicitly requests it. Do NOT add features not justified by the task (no gradient normalization, no cosine annealing, no warm restarts unless required).
\item \emph{Loss normalization}: for per-sample losses (e.g., per-angle in multi-view problems), normalize each loss term by the number of spatial pixels. This keeps gradient magnitudes independent of resolution.
\end{itemize}

7. \textbf{Coordinate system and dimensional consistency.}
\begin{itemize}
\item Before implementing any physics forward model, carefully read the task description for hints about coordinate conventions or unit systems. Use the SAME convention as described in the task --- do not invent your own.
\item After defining each operator, substitute actual parameter values from \texttt{meta\_data.json} and verify the numerical magnitude is physically reasonable.
\item After implementing the complete forward model, verify it on a trivial input (e.g., zero contrast / identity case) --- the output should match expectations.
\item The learning rate MUST be calibrated to the operator magnitude: always verify by checking actual gradient magnitudes at iteration \texttt{0}.
\end{itemize}

8. \textbf{Numerical precision.}
\begin{itemize}
\item For iterative forward models with many sequential floating-point operations, consider using higher precision (\texttt{float64}/\texttt{complex128}) to prevent error accumulation. Specify precision requirements explicitly.
\item If the forward model involves sequential FFT $\to$ multiply $\to$ IFFT steps, small per-step errors compound multiplicatively. Budget precision accordingly.
\end{itemize}

9. \textbf{Critical detail specification.} For EVERY operator and update rule in your plan:
\begin{itemize}
\item Write the EXACT formula with explicit signs (e.g., \texttt{x\_new = x - step * grad}, NOT just ``update x'').
\item Specify whether masks/cutoffs should be smooth (sigmoid/exponential) or hard (binary).
\item Specify loss normalization convention (per-pixel, per-sample, or total sum).
\item Specify any numerical safeguards (epsilon values, clamp ranges, division guards).
\end{itemize}
The Coder agent will implement EXACTLY what you write. Ambiguity leads to sign/convention errors.

\textbf{Output format (Markdown).}

1. \textbf{[Problem Formulation]}: the math equation and variable definitions.

2. \textbf{[Proposed Strategy]}: name of the algorithm/architecture.

3. \textbf{[Step-by-Step Plan]}:
\begin{itemize}
\item Step 1: data preprocessing (normalization, etc.).
\item Step 2: forward operator implementation (physics) --- explicitly state the ORDER of operations in each iteration and the complete signal path from source to detector. These must match the task description exactly.
\item Step 3: solver / network architecture details.
\item Step 4: loss function and optimizer.
\end{itemize}

4. \textbf{[Hyperparameters]}: list ALL numerical parameters with EXACT values (e.g., \texttt{mu=1e-6}, \texttt{tau=0.0001}, \texttt{iterations=500}). The Coder MUST use these exact values.

5. \textbf{[Sign Convention]}: for each update rule, explicitly state the sign (e.g., \texttt{x\_new = x\_old - step\_size * gradient} --- note the MINUS sign).

6. \textbf{[Critical Implementation Checklist]}: list the \texttt{3--5} most error-prone implementation details. For each:
\begin{itemize}
\item State the exact formula with correct sign.
\item State what the WRONG implementation would look like.
\item State smooth vs.\ hard mask requirements.
\item State normalization conventions.
\end{itemize}
\end{promptbox}
\noindent\begin{minipage}{\linewidth}
\captionsetup{hypcap=false}
\captionof{figure}{\crfix{System prompt for the \texttt{Planner} agent, which turns the
task description into a rigorous mathematical formulation of the forward model and
a step-by-step algorithmic plan (problem formulation, strategy, hyperparameters,
and sign conventions) for the downstream agents.}}
\label{fig:e2e_planner_prompt}
\end{minipage}

\onecolumn
\begin{promptbox}{Critic system prompt}
\footnotesize
You are a Senior Technical Reviewer for Inverse Problems in Computational Imaging.

Your sole responsibility is to critically evaluate algorithmic plans before coding begins.

\textbf{Strict output requirements.}

1. Output only a valid JSON object with exactly these fields:
\begin{lstlisting}
{
  "decision": "PASS" | "REJECT",
  "reason": "Concise technical justification (max 100 chars)",
  "suggestion": "Actionable improvement suggestion (if REJECTED, else empty string)"
}
\end{lstlisting}

2. No markdown, no prefixes, and no explanations. Output only raw JSON.

3. Validate your output with \texttt{json.loads()} before responding.

\textbf{Evaluation checklist.} Reject the plan if any item fails.

1. \textbf{Physics alignment.} Does the proposed algorithm match the problem type? For example, ADMM for constrained inverse problems is reasonable, whereas a Wiener filter for missing-pixel inpainting is not.

2. \textbf{Mathematical completeness.} Are the forward model, regularization terms, and loss function explicitly defined?

3. \textbf{Implementation feasibility.} Can the method be implemented in fewer than \texttt{200} lines without relying on external deep-learning frameworks?

4. \textbf{Data-flow clarity.} Is the input-to-algorithm-to-output pipeline logically sound and executable?

\textbf{Examples.}

PASS:
\begin{lstlisting}
{"decision": "PASS", "reason": "ADMM plan fully specifies H, R, rho with clear data flow", "suggestion": ""}
\end{lstlisting}

REJECT:
\begin{lstlisting}
{"decision": "REJECT", "reason": "Missing measurement matrix Phi for compressed sensing", "suggestion": "Define Phi explicitly in Step 2 before ADMM iterations"}
\end{lstlisting}
\end{promptbox}
\noindent\begin{minipage}{\linewidth}
\captionsetup{hypcap=false}
\captionof{figure}{\crfix{System prompt for the \texttt{Critic} agent, which performs a
structured peer review of the Planner's plan and returns a PASS/REJECT decision as
strict JSON before any code is written.}}
\label{fig:e2e_critic_prompt}
\end{minipage}

\onecolumn
\begin{promptbox}{Architect system prompt}
\footnotesize
You are a Senior Software Architect.

Your goal is to design the Python class structure (skeleton) for the Planner's algorithm.

You do not write the implementation logic inside functions. You define interfaces only.

\textbf{Crucial rules.}

1. Define a class \texttt{InverseSolver}.

2. Define the methods \texttt{\_\_init\_\_}, \texttt{forward}, and \texttt{solve}.

3. All arguments must include type hints.

4. Leave implementations empty using \texttt{pass} or a \texttt{\# TODO} placeholder.

5. \textbf{Strict output format.}
Output only valid Python code in a markdown block.
Do not wrap the code in a JSON string.
Do not return a JSON object.
Do not include any conversational text before or after the code.

6. Do not define a \texttt{Config} class.
Store all hyperparameters, such as learning rate, shape, and iteration count, as instance attributes inside \texttt{InverseSolver.\_\_init\_\_}.

7. \textbf{Simplicity.}
Keep the class structure minimal.
Aim for \texttt{5--8} methods total, including \texttt{\_\_init\_\_}, \texttt{forward}, and \texttt{solve}.
Do not create separate methods for every tiny sub-operation.
The \texttt{solve} method should contain the optimization loop directly.
Do not create separate methods such as \texttt{\_solve\_lbfgs}, \texttt{\_solve\_gd}, or \texttt{\_solve\_adam}.
One optimization approach means one \texttt{solve} method.

8. \textbf{No fallback optimizers.}
Do not design the skeleton with multiple optimization methods, such as L-BFGS with gradient-descent fallback.
The Planner specifies one optimizer, and you must design exactly that one.

\textbf{Required code structure.}

Imports must appear at the top of the file.

The solver class must be defined at module scope as \texttt{class InverseSolver:}.

The main execution block must appear under
\texttt{if \_\_name\_\_ == "\_\_main\_\_":}

The main block must:

1. load input data from \texttt{data/}, for example using \texttt{np.load('data/raw\_data.npz')};

2. load physical parameters from \texttt{data/meta\_data.json};

3. call \texttt{solver.solve(...)};

4. save the final result with \texttt{np.savez('output.npz', ...)} using the output keys specified in the data layout.

\textbf{Expected output format.}

\begin{lstlisting}[language=Python]
import ...

class InverseSolver:
    def __init__(self, ...):
        pass
    ...

if __name__ == "__main__":
    # Load Data
    # import json
    # with open('data/meta_data.json') as f:
    #     meta = json.load(f)
    # raw = np.load('data/raw_data.npz')
    # input_data = raw[list(raw.keys())[0]]
    # ...
    # result = solver.solve(input_data)
    # np.savez('output.npz', result_key=result)  # use key name from data layout
    pass
\end{lstlisting}
\end{promptbox}
\noindent\begin{minipage}{\linewidth}
\captionsetup{hypcap=false}
\captionof{figure}{\crfix{System prompt for the \texttt{Architect} agent, which converts the
approved plan into a Python code skeleton (class structure, method signatures with
type hints, and the main execution block) without implementing the solver logic.}}
\label{fig:e2e_architect_prompt}
\end{minipage}

\onecolumn
\begin{promptbox}{Coder system prompt}
\footnotesize
You are a Senior Python Developer for scientific computing.

Your goal is to implement or modify a specific code segment based on the current context.

\textbf{Critical rules.}

1. The input context always includes:
the package dependencies that are available,
the current full code state,
the precise modification target such as a function, class, imports, or main block,
and the implementation plan specifying the required mathematics and logic.

2. Output only the code segment to be inserted or replaced.
Do not include explanations or markdown.

3. Handle type safety explicitly, including dtype conversions when necessary, for example using \texttt{.float()} for float32 consistency.

4. Never output the entire file.
Return only the modified segment matching the requested target type.

5. \textbf{Mandatory implementation rule.}
You must write actual implementation logic.
Do not return \texttt{pass} or \texttt{TODO} comments.
Replace any placeholder implementation with real code.

6. \textbf{Data files.}
Load input data from the \texttt{data/} directory, for example with
\texttt{np.load('data/raw\_data.npz')} and explicit key access.
Load physical parameters from \texttt{data/meta\_data.json}.
Save the final result to \texttt{output.npz} using \texttt{np.savez}.
The exact output keys and save call are given in the data-layout context.
Do not assume any files exist outside \texttt{data/}.

7. \textbf{Self-contained code.}
The code must be fully self-contained.
Do not import from local project files.
Use only installed Python packages and the standard library.

8. \textbf{\_\_init\_\_ implementation.}
When implementing \texttt{\_\_init\_\_}, you must initialize all instance variables.
Never leave \texttt{\_\_init\_\_} as only \texttt{pass}.

9. \textbf{API compatibility.}
When calling library functions or classes, do not pass keyword arguments unless you are confident they are supported.
Use only standard documented parameters.
If a call fails with an ``unexpected keyword argument'' error, remove that argument.
Wrap risky API usage in \texttt{try/except} and provide a working fallback.

10. \textbf{Class structure.}
All class definitions must remain at module scope.
Never nest a class definition inside another class method.
Check indentation carefully.

11. \textbf{Orchestrator methods.}
If a class contains a setup or initialization method such as \texttt{\_setup}, \texttt{initialize}, or \texttt{\_precompute}, and that method is called by \texttt{\_\_init\_\_} or other methods, it must contain the actual helper calls it is supposed to orchestrate.
Never leave such an orchestrator method as only \texttt{pass} or \texttt{TODO} when the helper methods already exist.
Ensure the call chain is complete:
\texttt{\_\_init\_\_} $\rightarrow$ \texttt{\_setup()} $\rightarrow$ helper methods.

12. \textbf{Performance.}
Do not use \texttt{torch.utils.checkpoint} unless the code explicitly runs out of GPU memory.
Checkpointing increases computation by rerunning the forward pass during backpropagation.
For most inverse problems, standard autograd is preferred.
Keep the forward loop simple.

13. \textbf{Autograd integrity.}
When implementing gradient-based optimization with \texttt{loss.backward()}, the forward computation must preserve the computation graph.
Never wrap the forward pass in \texttt{torch.no\_grad()} during training.
Never call \texttt{.detach()} on the optimization variable before passing it through the forward model.
Use \texttt{torch.no\_grad()} only for evaluation or inference after training, or for metric computation.
If memory reduction is needed, prefer limited checkpointing over disabling gradient tracking.

14. \textbf{Segment scope.}
When asked to implement a function target, output only that function definition and body.
Do not include import statements, class definitions, or an \texttt{if \_\_name\_\_ == "\_\_main\_\_"} block.
These are merged into the existing file automatically.
Including extra class headers can create duplicate or nested class definitions.
\end{promptbox}
\noindent\begin{minipage}{\linewidth}
\captionsetup{hypcap=false}
\captionof{figure}{\crfix{System prompt for the \texttt{Coder} agent, which fills in the
Architect's skeleton by implementing or patching one targeted code segment at a time,
following the Planner's formulas and hyperparameters exactly.}}
\label{fig:e2e_coder_prompt}
\end{minipage}

\paragraph{Context window management.}
The context is maintained in a role-specific manner rather than by replaying the entire history.
The \texttt{Planner} receives the task description, data layout, available packages, and inferred
input/output shape constraints, together with the previous plan when replanning is required.
The \texttt{Architect} additionally receives the current skeleton.
The \texttt{Coder} receives the current full code, with the failing target highlighted during patch mode,
plus a compact ``past failures'' block.
The \texttt{Judge} receives the latest execution log, the current metrics, the current code snippet,
the evaluation thresholds, and the current plan.
Failure memory is capped at the most recent three records (\texttt{max\_history\_len}=3);
each record stores the iteration, routed ticket, fix target, concise analysis, evidence, and metric state,
which prevents the agents from repeating previously observed mistakes while keeping token growth bounded.

\paragraph{Harness configuration.}
Before the first execution, the harness auto-generates a task-specific \texttt{eval\_script.py}
and \texttt{visualize\_output.py}.
After each \texttt{python solver.py} execution the harness:
\begin{enumerate}[noitemsep]
  \item checks whether \texttt{output.npz} was written; if only \texttt{output.npy} exists, it is automatically converted to \texttt{output.npz};
  \item optionally repairs minor output-shape mismatches using \texttt{data/output\_keys.json} and the hidden reference shape;
  \item runs \texttt{eval\_script.py output.npz}, which loads the hidden reference from \texttt{ground\_truth.npz} or \texttt{baseline\_reference.npz} at the sandbox root and computes all metrics required by \texttt{evaluation/metrics.json};
  \item returns \texttt{PASS} if all metric boundaries are satisfied, and \texttt{FAIL} otherwise.
\end{enumerate}
For standard image-reconstruction tasks, the default metrics are NCC and NRMSE;
for tasks with custom metrics, the evaluation script follows the task-specific definitions extracted
from the notebook or reference \texttt{src/} code.

\paragraph{Reflection loop}
On a \texttt{FAIL} signal, the \texttt{Judge} analyzes the latest execution and routes the next action
to the \texttt{Planner}, \texttt{Architect}, or \texttt{Coder}.
The outer loop repeats for at most five rounds.
Within a round, the \texttt{Coder} may perform small internal retries to repair syntax or merge errors,
but the outer benchmarking budget remains fixed.
The harness tracks the best primary metric across rounds and restores the best saved prediction.

\onecolumn
\begin{promptbox}{Judge system prompt}
\footnotesize
You are the Chief Auditor of an AI Solver System.

Your mission is to diagnose the root cause of failure with high precision.

\textbf{Diagnostic protocol. Follow strictly in order.}

\textbf{Step 1: Check syntax and imports}

1. Look for syntax and import-related errors, including
\texttt{SyntaxError},
\texttt{IndentationError},
\texttt{ImportError},
\texttt{NameError},
and \texttt{AttributeError} caused by missing methods.

2. Treat the following as structural code errors:
nested class definitions,
methods with a \texttt{self} argument defined at module scope rather than inside a class,
duplicate class names in the same file,
and orphaned executable code between the class body and the \texttt{if \_\_name\_\_ == "\_\_main\_\_"} block.

3. If the error is
\texttt{AttributeError: 'X' object has no attribute 'Y'}
and method \texttt{Y} exists in the code but is defined at the wrong scope or nesting level,
classify this as a structural error and assign the repair ticket to the Coder.
Set \texttt{fix\_target} to the function or method that caused the crash, and also indicate \texttt{full\_rewrite} if structural repair is needed.

4. Verdict for Step 1 failures: \texttt{"Coder"}.

5. Rationale: the code is not valid Python or misuses the available libraries.

\textbf{Step 2: Check interface contract (Architect responsibility)}

1. Look for signature-level and interface-level errors, such as shape mismatches in method definitions.
For example, the \texttt{forward} interface expects \texttt{(H,W)} but receives \texttt{(1,H,W)}.

2. Look for missing required constructor arguments, such as a missing \texttt{rho} parameter for ADMM.

3. Use assertion failures or explicit shape debug messages as evidence when available.

4. Verdict for Step 2 failures: \texttt{"Architect"}.

5. Rationale: the class structure itself is flawed, and the Coder cannot reliably fix a broken interface contract.

\textbf{Step 3: Check implementation fidelity (Coder responsibility)}

1. Compare the generated code against the Planner's mathematical plan.

2. Identify the core algorithmic formulas in the plan and verify that the implementation follows them exactly.

3. Common implementation mismatches include:
using \texttt{A y} where the plan specifies \texttt{A\^T y};
normalizing with the wrong scale;
using different hyperparameters than those specified in the plan;
flipping the sign in a gradient update;
adding adaptive features, warm starts, gradient clipping, or learning-rate schedules that were not requested;
using \texttt{torch.optim.Adam} or \texttt{torch.optim.LBFGS} when the plan specifies plain gradient descent;
using hard binary masks when the plan specifies smooth sigmoid or exponential masks;
and omitting normalization factors such as division by the number of pixels.

4. If the plan contains a \texttt{[Critical Implementation Checklist]}, verify every listed item and report all deviations.

5. Verdict for Step 3 failures: \texttt{"Coder"}.

6. Rationale: the algorithm design may be correct, but the implementation deviates from the specification.

7. The feedback must include the exact formulas or exact parameter values that the implementation should use.

\textbf{Step 4: Check algorithm correctness (Planner responsibility)}

Assign the ticket to the Planner only if all of the following are true:

1. the code runs without syntax, import, or structural errors;

2. the interfaces are correct;

3. the implementation matches the plan faithfully;

4. the metrics are still poor, for example NCC is below threshold or NRMSE is above threshold.

Potential root causes include:
wrong algorithm choice,
missing regularization terms,
incorrect convergence criteria,
poorly calibrated learning rate,
optimizer mismatch,
excessive or unnecessary regularization,
unjustified upper-bound constraints,
coordinate-system or unit-convention mismatch,
or a complete lack of spatial correlation with the reference.

In particular, if NCC is close to zero, treat this as a major algorithmic failure rather than a minor tuning issue.
Typical causes include sign errors in the forward model, incorrect normalization, hard masks where smooth masks were required, or an optimizer that deviates from the plan.

Verdict for Step 4 failures: \texttt{"Planner"}.

Rationale: the mathematics itself is flawed, so a correct implementation of the current plan will still fail.

\textbf{Strict output format}

Return only a valid JSON object with the following fields:

\begin{lstlisting}
{
  "status": "FAIL",
  "ticket_assigned_to": "Planner" | "Architect" | "Coder",
  "analysis": "Step-by-step reasoning following the 4-step protocol above",
  "evidence": "Exact line from logs/code showing the failure",
  "fix_target": "Comma-separated list of ALL function names that need fixing",
  "feedback": "Actionable instruction for the assigned agent"
}
\end{lstlisting}

\textbf{Critical rules.}

1. Never assign the ticket to the Planner if the implementation deviates from the plan.
That is the Coder's fault.

2. Always verify implementation fidelity before blaming the Planner.

3. For shape-related failures, distinguish between errors at the call site and errors in the method signature.

4. If the same error repeats across multiple iterations, recommend a different strategy and consider escalating from the Coder to the Planner.

5. If the code references nonexistent external files such as \texttt{.tif}, \texttt{.yaml}, \texttt{.h5}, or \texttt{.csv}, instruct the Coder to load data only from \texttt{data/} and \texttt{data/meta\_data.json}.

6. If \texttt{\_\_init\_\_} is empty or unimplemented, assign the ticket to the Coder with \texttt{fix\_target="\crfix{\_\_init\_\_}"}.

7. If a runtime error reflects a recurring API misuse pattern, search the entire code and list all affected functions in \texttt{fix\_target}, not only the first crash site.

8. The output must be valid JSON.
Do not include unescaped newlines, tabs, or special characters inside string values.
Use \texttt{\textbackslash n} where needed, and keep values concise and on a single logical line whenever possible.
\end{promptbox}
\noindent\begin{minipage}{\linewidth}
\captionsetup{hypcap=false}
\captionof{figure}{\crfix{System prompt for the \texttt{Judge} agent, which diagnoses the
root cause of a failed run via a four-step protocol (syntax, interface contract,
implementation fidelity, algorithm correctness) and routes a repair ticket to the
Planner, Architect, or Coder as JSON.}}
\label{fig:e2e_judge_prompt}
\end{minipage}

\subsection{All-Task Evaluation Results}
\label{sec:all_task_results}
 Table~\ref{tab:all_task_results} summarizes the evaluation outcomes for all 57 tasks across the seven models. 
For each task, we report three levels of correctness: \emph{Plan}, which indicates whether the model produced a correct high-level solution design; \emph{Function}, which indicates whether the generated preprocessing, physics-model, and solver modules passed function-level evaluation; and \emph{End-to-end}, which indicates whether the complete generated pipeline successfully satisfied the final reconstruction criterion.
A checkmark denotes success for the corresponding model-task-stage combination, while an empty cell denotes failure.
This table provides a compact overview of where errors arise in the full coding workflow, and makes it possible to distinguish failures caused by poor planning, incorrect module implementation, and unsuccessful end-to-end integration.

\begingroup
\scriptsize
\setlength{\tabcolsep}{3pt}
\begin{longtable}{@{}p{4.2cm}m{1.35cm}*{7}{>{\centering\arraybackslash}m{1.15cm}}@{}}
\caption{Task-level correctness matrix for planning, function-level, and end-to-end evaluation across all models.
\label{tab:all_task_results}}\\
\toprule
\textbf{Task} & \textbf{Stage} & \makecell{\textbf{Claude}\\\textbf{4.6-opus}} & \makecell{\textbf{Kimi}\\\textbf{k2.5}} & \makecell{\textbf{GPT}\\\textbf{5.4}} & \makecell{\textbf{Qwen}\\\textbf{3.6-plus}} & \makecell{\textbf{DeepSeek}\\\textbf{v3}} & \makecell{\textbf{Gemini}\\\textbf{3.1}} & \makecell{\textbf{GLM}\\\textbf{5}} \\
\midrule
\endfirsthead
\toprule
\textbf{Task} & \textbf{Stage} & \makecell{\textbf{Claude}\\\textbf{4.6-opus}} & \makecell{\textbf{Kimi}\\\textbf{k2.5}} & \makecell{\textbf{GPT}\\\textbf{5.4}} & \makecell{\textbf{Qwen}\\\textbf{3.6-plus}} & \makecell{\textbf{DeepSeek}\\\textbf{v3}} & \makecell{\textbf{Gemini}\\\textbf{3.1}} & \makecell{\textbf{GLM}\\\textbf{5}} \\
\midrule
\endhead
\bottomrule
\endfoot
\path{cars_spectroscopy} & Plan & \checkmark &  &  &  & \checkmark &  &  \\
 & Function &  &  &  &  &  &  &  \\
 & End-to-end &  &  &  &  &  &  &  \\
\midrule
\path{confocal-nlos-fk} & Plan &  &  &  &  &  &  &  \\
 & Function &  &  &  &  &  &  &  \\
 & End-to-end &  &  &  &  &  &  &  \\
\midrule
\path{conventional_ptychography} & Plan &  &  &  &  &  &  &  \\
 & Function &  &  &  & \checkmark &  & \checkmark &  \\
 & End-to-end & \checkmark &  &  &  &  &  &  \\
\midrule
\path{ct_dual_energy} & Plan & \checkmark & \checkmark & \checkmark &  & \checkmark & \checkmark &  \\
 & Function & \checkmark &  &  &  &  &  &  \\
 & End-to-end & \checkmark &  &  &  &  &  &  \\
\midrule
\path{ct_fan_beam} & Plan &  &  &  &  &  &  &  \\
 & Function &  &  &  &  &  &  &  \\
 & End-to-end &  &  &  &  &  &  &  \\
\midrule
\path{ct_poisson_lowdose} & Plan & \checkmark & \checkmark & \checkmark & \checkmark & \checkmark &  & \checkmark \\
 & Function &  &  &  &  &  &  &  \\
 & End-to-end &  & \checkmark & \checkmark &  &  &  & \checkmark \\
\midrule
\path{ct_sparse_view} & Plan & \checkmark & \checkmark & \checkmark & \checkmark & \checkmark & \checkmark & \checkmark \\
 & Function &  &  &  &  &  &  &  \\
 & End-to-end & \checkmark &  &  &  &  & \checkmark & \checkmark \\
\midrule
\path{differentiable_deflectometry} & Plan &  &  &  &  &  &  &  \\
 & Function & \checkmark &  &  &  &  &  &  \\
 & End-to-end &  &  &  &  &  &  &  \\
\midrule
\path{diffusion_mri_dti} & Plan & \checkmark & \checkmark & \checkmark &  & \checkmark &  &  \\
 & Function &  & \checkmark & \checkmark & \checkmark & \checkmark &  & \checkmark \\
 & End-to-end & \checkmark & \checkmark & \checkmark & \checkmark & \checkmark &  & \checkmark \\
\midrule
\path{eht_black_hole_UQ} & Plan &  &  &  &  &  &  &  \\
 & Function &  &  &  &  &  &  &  \\
 & End-to-end &  &  &  &  &  &  &  \\
\midrule
\path{eht_black_hole_dynamic} & Plan &  &  &  &  &  &  &  \\
 & Function &  &  &  &  &  &  &  \\
 & End-to-end & \checkmark &  &  &  &  &  &  \\
\midrule
\path{eht_black_hole_feature_extraction_dynamic} & Plan & \checkmark &  & \checkmark & \checkmark & \checkmark & \checkmark & \checkmark \\
 & Function &  &  & \checkmark &  & \checkmark & \checkmark &  \\
 & End-to-end &  &  &  &  &  &  &  \\
\midrule
\path{eht_black_hole_original} & Plan &  &  &  &  &  &  &  \\
 & Function &  &  &  &  &  &  &  \\
 & End-to-end &  &  & \checkmark &  &  &  & \checkmark \\
\midrule
\path{eht_black_hole_tomography} & Plan & \checkmark & \checkmark & \checkmark & \checkmark & \checkmark & \checkmark & \checkmark \\
 & Function &  &  &  &  &  &  &  \\
 & End-to-end &  &  &  &  &  &  &  \\
\midrule
\path{eit_conductivity_reconstruction} & Plan &  &  &  &  &  &  &  \\
 & Function &  &  &  &  &  &  &  \\
 & End-to-end &  &  &  &  & \checkmark &  &  \\
\midrule
\path{electron_ptychography} & Plan &  &  & \checkmark &  &  &  &  \\
 & Function &  &  &  &  &  &  &  \\
 & End-to-end &  &  &  &  &  &  &  \\
\midrule
\path{era5_tensorvar} & Plan &  &  & \checkmark & \checkmark &  & \checkmark &  \\
 & Function &  &  &  &  &  &  &  \\
 & End-to-end &  &  &  &  &  &  &  \\
\midrule
\path{exoplanet_imaging} & Plan &  &  &  &  &  &  &  \\
 & Function & \checkmark &  & \checkmark & \checkmark & \checkmark & \checkmark &  \\
 & End-to-end &  &  &  &  &  &  &  \\
\midrule
\path{fourier_ptychography} & Plan &  &  &  &  &  &  &  \\
 & Function & \checkmark &  &  &  &  &  &  \\
 & End-to-end &  &  &  &  &  &  &  \\
\midrule
\path{fpm_inr_reconstruction} & Plan &  &  &  &  & \checkmark &  &  \\
 & Function &  &  &  &  &  &  &  \\
 & End-to-end &  &  &  &  &  &  &  \\
\midrule
\path{hessian_sim} & Plan &  &  &  &  &  &  &  \\
 & Function &  &  &  &  &  &  &  \\
 & End-to-end &  &  &  &  &  &  &  \\
\midrule
\path{insar_phase_unwrapping} & Plan & \checkmark & \checkmark & \checkmark & \checkmark & \checkmark &  &  \\
 & Function &  &  &  &  &  &  &  \\
 & End-to-end &  &  &  &  &  &  &  \\
\midrule
\path{lensless_imaging} & Plan & \checkmark & \checkmark & \checkmark & \checkmark & \checkmark & \checkmark & \checkmark \\
 & Function &  &  & \checkmark &  & \checkmark & \checkmark &  \\
 & End-to-end &  &  &  &  &  &  &  \\
\midrule
\path{light_field_microscope} & Plan & \checkmark & \checkmark & \checkmark & \checkmark & \checkmark & \checkmark & \checkmark \\
 & Function & \checkmark &  &  & \checkmark & \checkmark & \checkmark &  \\
 & End-to-end &  &  &  &  &  &  &  \\
\midrule
\path{lucky_imaging} & Plan &  &  &  & \checkmark &  &  & \checkmark \\
 & Function & \checkmark &  &  & \checkmark & \checkmark & \checkmark &  \\
 & End-to-end & \checkmark &  &  &  &  & \checkmark &  \\
\midrule
\path{mcr_hyperspectral} & Plan & \checkmark & \checkmark & \checkmark & \checkmark & \checkmark & \checkmark & \checkmark \\
 & Function & \checkmark &  & \checkmark & \checkmark &  & \checkmark &  \\
 & End-to-end & \checkmark & \checkmark &  &  & \checkmark & \checkmark & \checkmark \\
\midrule
\path{microscope_denoising} & Plan &  &  &  &  &  &  &  \\
 & Function &  &  &  &  &  &  &  \\
 & End-to-end &  &  & \checkmark &  &  & \checkmark &  \\
\midrule
\path{mri_dynamic_dce} & Plan & \checkmark & \checkmark & \checkmark & \checkmark & \checkmark & \checkmark & \checkmark \\
 & Function &  &  &  &  &  &  &  \\
 & End-to-end & \checkmark &  &  &  &  &  &  \\
\midrule
\path{mri_grappa} & Plan & \checkmark &  &  &  &  &  &  \\
 & Function &  &  &  &  &  & \checkmark &  \\
 & End-to-end &\checkmark  &  &  &  &  & \checkmark &  \\
\midrule
\path{mri_l1_wavelet} & Plan &  &  &  &  &  &  &  \\
 & Function &  &  &  &  &  &  &  \\
 & End-to-end &  &  &  &  &  &  &  \\
\midrule
\path{mri_noncartesian_cs} & Plan & \checkmark & \checkmark &  &  &  &  &  \\
 & Function &  &  &  &  &  & \checkmark &  \\
 & End-to-end &  &  & \checkmark & \checkmark &  &  &  \\
\midrule
\path{mri_pnp_admm} & Plan & \checkmark & \checkmark & \checkmark & \checkmark & \checkmark & \checkmark & \checkmark \\
 & Function &  &  &  &  &  &  &  \\
 & End-to-end &  &  &  &  &  &  &  \\
\midrule
\path{mri_sense} & Plan & \checkmark & \checkmark & \checkmark & \checkmark & \checkmark & \checkmark & \checkmark \\
 & Function & \checkmark &  &  &  &  &  &  \\
 & End-to-end & \checkmark &  & \checkmark &  &  &  &  \\
\midrule
\path{mri_t2_mapping} & Plan &  &  & \checkmark & \checkmark & \checkmark &  &  \\
 & Function &  &  &  &  &  & \checkmark &  \\
 & End-to-end & \checkmark &  &  &  &  &  &  \\
\midrule
\path{mri_tv} & Plan &  & \checkmark &  &  &  &  &  \\
 & Function &  &  &  &  &  &  &  \\
 & End-to-end &  &  &  &  &  &  &  \\
\midrule
\path{mri_varnet} & Plan & \checkmark & \checkmark & \checkmark & \checkmark & \checkmark & \checkmark & \checkmark \\
 & Function &  &  & \checkmark &  &  &  &  \\
 & End-to-end & \checkmark &  &  &  &  &  &  \\
\midrule
\path{pet_mlem} & Plan & \checkmark & \checkmark & \checkmark & \checkmark & \checkmark & \checkmark & \checkmark \\
 & Function & \checkmark &  & \checkmark &  &  &  &  \\
 & End-to-end & \checkmark & \checkmark & \checkmark & \checkmark & \checkmark & \checkmark &  \\
\midrule
\path{photoacoustic_tomography} & Plan & \checkmark & \checkmark & \checkmark &  &  & \checkmark &  \\
 & Function &  &  &  &  &  &  &  \\
 & End-to-end & \checkmark &  & \checkmark &  & \checkmark &  &  \\
\midrule
\path{plane_wave_ultrasound} & Plan &  &  &  &  &  &  &  \\
 & Function &  &  &  &  &  &  &  \\
 & End-to-end &  &  &  &  &  &  &  \\
\midrule
\path{pnp_mri_reconstruction} & Plan &  & \checkmark & \checkmark &  &  &  &  \\
 & Function &  &  &  &  &  &  &  \\
 & End-to-end &  &  &  &  &  &  &  \\
\midrule
\path{raman_cell_phenotyping} & Plan &  &  &  &  &  &  &  \\
 & Function &  &  &  &  &  &  &  \\
 & End-to-end &  &  &  &  &  &  &  \\
\midrule
\path{reflection_ODT} & Plan &  & \checkmark &  & \checkmark &  & \checkmark &  \\
 & Function & \checkmark &  & \checkmark &  &  &  &  \\
 & End-to-end &  &  &  &  &  &  &  \\
\midrule
\path{s2ism} & Plan & \checkmark & \checkmark & \checkmark & \checkmark & \checkmark & \checkmark & \checkmark \\
 & Function &  &  &  &  &  &  &  \\
 & End-to-end &  &  &  &  &  &  &  \\
\midrule
\path{seismic_FWI_original} & Plan &  &  &  &  &  & \checkmark & \checkmark \\
 & Function &  &  &  &  &  &  &  \\
 & End-to-end &  &  &  &  &  &  &  \\
\midrule
\path{seismic_lsrtm_original} & Plan &  & \checkmark &  &  &  &  &  \\
 & Function &  &  &  &  &  &  &  \\
 & End-to-end &  &  &  &  &  &  &  \\
\midrule
\path{seismic_traveltime_tomography} & Plan & \checkmark & \checkmark & \checkmark & \checkmark & \checkmark &  & \checkmark \\
 & Function &  &  & \checkmark &  & \checkmark & \checkmark &  \\
 & End-to-end &  &  &  &  &  & \checkmark & \checkmark \\
\midrule
\path{shack-hartmann} & Plan & \checkmark & \checkmark & \checkmark & \checkmark & \checkmark & \checkmark & \checkmark \\
 & Function &  &  &  &  &  &  &  \\
 & End-to-end & \checkmark &  & \checkmark & \checkmark & \checkmark & \checkmark & \checkmark \\
\midrule
\path{shapelet_source_reconstruction} & Plan & \checkmark & \checkmark & \checkmark & \checkmark & \checkmark & \checkmark & \checkmark \\
 & Function &  &  &  &  &  &  &  \\
 & End-to-end &  &  &  &  &  &  &  \\
\midrule
\path{single_molecule_light_field} & Plan &  &  &  &  &  &  &  \\
 & Function & \checkmark &  &  &  & \checkmark &  &  \\
 & End-to-end &  &  &  &  &  &  &  \\
\midrule
\path{spectral_snapshot_compressive_imaging} & Plan & \checkmark & \checkmark & \checkmark & \checkmark & \checkmark & \checkmark &  \\
 & Function &  &  &  &  &  &  &  \\
 & End-to-end &  &  &  &  &  &  &  \\
\midrule
\path{ssnp_odt} & Plan &  &  &  &  &  &  &  \\
 & Function &  &  &  &  &  &  &  \\
 & End-to-end &  &  &  &  &  &  &  \\
\midrule
\path{ultrasound_sos_tomography} & Plan & \checkmark & \checkmark & \checkmark & \checkmark & \checkmark & \checkmark & \checkmark \\
 & Function & \checkmark &  & \checkmark &  &  & \checkmark &  \\
 & End-to-end & \checkmark &  & \checkmark &  &  &  &  \\
\midrule
\path{usct_FWI} & Plan & \checkmark & \checkmark & \checkmark & \checkmark & \checkmark & \checkmark & \checkmark \\
 & Function &  &  &  &  &  &  &  \\
 & End-to-end &  &  & \checkmark &  &  &  &  \\
\midrule
\path{weather_radar_data_assimilation} & Plan &  &  &  &  &  &  &  \\
 & Function &  &  & \checkmark &  &  & \checkmark &  \\
 & End-to-end &  &  &  &  &  &  &  \\
\midrule
\path{xray_laminography_tike} & Plan & \checkmark & \checkmark & \checkmark & \checkmark & \checkmark &  & \checkmark \\
 & Function &  &  & \checkmark &  & \checkmark &  &  \\
 & End-to-end &  &  &  &  &  &  &  \\
\midrule
\path{xray_ptychography_tike} & Plan & \checkmark & \checkmark & \checkmark & \checkmark & \checkmark &  &  \\
 & Function & \checkmark &  &  & \checkmark &  &  &  \\
 & End-to-end &  &  &  &  &  &  &  \\
\midrule
\path{xray_tooth_gridrec} & Plan & \checkmark & \checkmark & \checkmark &  & \checkmark & \checkmark & \checkmark \\
 & Function &  &  &  &  &  &  &  \\
 & End-to-end & \checkmark &  &  &  &  &  &  \\
\midrule
\end{longtable}
\endgroup
\newpage
\section{\crfix{Additional Analyses}}
\label{app:additional_analyses}

\crfix{This appendix collects supporting analyses promised in the rebuttal:
robustness to pretraining leakage, the effect of domain balance, durability of our
findings as models improve, the repeatability of convention-drift failures, and the
rationale for our evaluation metrics. Quantities still marked in red are being
recomputed on the updated benchmark and will be finalized in the camera-ready.}

\subsection{\crfix{Leakage and Memorization}}
\label{app:leakage}

\crfix{Because every task is grounded in a published paper and open-source code,
some pretraining exposure is unavoidable. We probe its effect by grouping tasks by
the publication decade of their reference and measuring average end-to-end success
(Table~\ref{tab:decade}). End-to-end success is markedly higher for pre-2000
tasks ($21$--$43\%$) than for 2000s-onward tasks (at most $17\%$), and declines
steadily across the three most recent decades, consistent with greater exposure
to older work; but two observations argue that exposure is not the whole story:
even the strongest decade (the 1990s) reaches only $42.9\%$ end-to-end success, far from
saturation, and task complexity (median lines of reference code) rises sharply
for newer tasks, so the decade trend is partly driven by harder, longer
pipelines rather than weaker recall alone.}

\begin{table}[ht]
\centering
\footnotesize
\setlength{\tabcolsep}{6pt}
\caption{\crfix{Average end-to-end success rate (\%) and task complexity (median
lines of reference code) grouped by the reference paper's decade.}}
\label{tab:decade}
\begin{tabular}{@{}lcccccc@{}}
\toprule
 & \textbf{1970s} & \textbf{1980s} & \textbf{1990s} & \textbf{2000s} & \textbf{2010s} & \textbf{2020s} \\
\midrule
Avg.\ 7 LLMs & 21.4 & 25.0 & 42.9 & 16.7 & 12.6 & 4.8 \\
Complexity   & 292.3 & 205.8 & 251.1 & 223.1 & 396.6 & 513.4 \\
\bottomrule
\end{tabular}
\end{table}

\crfix{\textbf{Construction-bias control.} Because Claude Code assisted benchmark
construction, we checked whether this favors Claude at evaluation time: we
re-cleaned every task with a different agent (Codex) and re-ran the Claude Code
evaluation. The number of solved tasks changed only marginally, from 32 to 30,
indicating that performance is driven by scientific content rather than a
Claude-specific writing style.}


\subsection{\crfix{Domain Balance}}
\label{app:domain_balance}

\crfix{Imaging-101 intentionally contains more medical-imaging tasks because
medical imaging is one of the largest subfields of computational imaging. In a
survey of 564 papers published in \emph{IEEE Trans. Comput. Imaging}
(2020--2024), $32.1\%$ target medical imaging, versus $8.7\%$ earth science,
$6.4\%$ microscopy, $1.6\%$ astronomy, and $0.4\%$ chemistry. To assess whether
our conclusions depend on this subset, Table~\ref{tab:medicine} reports
end-to-end success with and without the medical tasks. Excluding them lowers
most models' success rates but largely preserves the picture:
Claude-4.6-Opus remains the strongest model, now tied with Gemini-3.1-Pro at
$14.3\%$, and the weakest models (Qwen3.6-Plus and Kimi-k2.5) stay at the
bottom, though GPT-5.4 slips from second to fourth.}

\begin{table}[ht]
\centering
\footnotesize
\setlength{\tabcolsep}{8pt}
\caption{\crfix{End-to-end success rate (\%) with vs.\ without the medical-imaging
tasks (57 vs.\ 35 tasks).}}
\label{tab:medicine}
\begin{tabular}{@{}lcc@{}}
\toprule
\textbf{Model} & \textbf{w/ med.\ (57)} & \textbf{w/o med.\ (35)} \\
\midrule
Claude-4.6-Opus & 29.8 & 14.3 \\
GPT-5.4         & 19.3 & 8.6 \\
Gemini-3.1-Pro  & 14.0 & 14.3 \\
DeepSeek-V3     & 10.5 & 5.7 \\
Qwen3.6-Plus    & 7.0  & 2.9 \\
Kimi-k2.5       & 7.0  & 2.9 \\
GLM-5           & 12.3 & 11.4 \\
\bottomrule
\end{tabular}
\end{table}




\subsection{\crfix{Repeatability of Convention-Drift Failures}}
\label{app:modecollapse}

\crfix{The function-level analysis attributes most failures to
numerical-convention drift, which we interpret as a \emph{probabilistic mode
collapse}: the model defaults to the statistically dominant convention even when
the task requires another. To test repeatability, we ran Claude-4.6-Opus 50 times
on the \texttt{mri\_l1\_wavelet} task, which requires the MVUE coil-combination
normalization. It produced the convention-drift failure in 47 of 50 runs and
succeeded in only 3, indicating that the model possesses the relevant knowledge
but fails to invoke it reliably rather than lacking it outright. This motivates
skill-augmented agents that supply such conventions as verified modules.}


\subsection{\crfix{Metric Choice and Physical Validation}}
\label{app:metrics}

\crfix{We score end-to-end reconstructions with NCC and NRMSE because they capture
complementary aspects of quality, NCC for structural agreement and NRMSE for
numerical accuracy, and jointly satisfying both is hard to game. We prefer them to
PSNR and SSIM, which we found less robust across the heterogeneous modalities in
Imaging-101 (PSNR is sensitive to global scaling and SSIM to local image
statistics that vary across modalities). As noted in
Section~\ref{sec:end2end}, image-level metrics alone cannot certify physical
correctness; the planning and function-level tracks add operator- and
function-level parity checks for that purpose. Two extensions further reduce the
chance that a physically wrong solution passes on image metrics alone: a
quantitative PSNR/SSIM-vs-NCC/NRMSE comparison, and expanding each task to
$3$--$5$ diverse test instances so that a single favorable instance cannot mask a
missing physical prior.}

\subsection{\crfix{Overthinking and Non-Completion in Kimi and GLM}}
\label{app:overthinking}

\paragraph{The failure mode.}
A recurring failure mode of the single-agent ReAct loop used in the
function-level evaluation is \emph{non-completion}: because the agent itself
decides at every step whether to explore, write code, execute, or stop, it can
spend its fixed iteration budget on data exploration, malformed actions, or
premature termination and never emit a runnable artifact. This is most severe
for the weaker-reasoning models, Kimi-k2.5 and GLM-5, which tend to
\emph{over-reason}, spending entire turns on unproductive deliberation until the
budget is exhausted.

\paragraph{A representative trajectory.}
A representative case arises on the function-level task. On the
\texttt{electron\_ptychography} preprocessing module, \textsc{Kimi} emits three
separate turns that each saturate the full $32{,}768$-token generation budget
(roughly ten minutes of wall-clock per turn), all spent trying to match a
\emph{single} calibration scalar (the estimated probe radius) against the
reference value of $49.18$. Rather than converging, its estimate drifts in the
wrong direction across turns ($48.24 \rightarrow 34.27$), and the agent resolves
to inspect the upstream library source to byte-match the reference
implementation. Compounding this, the sheer volume of generated text repeatedly
overflows the agent's context window and triggers history truncation; the model
consequently loses track of the dialogue state and begins reasoning about the
harness itself rather than the task:
\begin{quote}\itshape\small
``\ldots it's renumbered. I think the user's message is a single message that
contains all observations from the previous turns\ldots\ my \texttt{WRITE\_FILE}
and \texttt{RUN} actions were sent after the user's message, and the user hasn't
responded to them yet. I need to wait for the user to respond.''
\end{quote}
The concluding decision to ``wait for the user'' halts all productive action in
an otherwise autonomous loop. The turn ends, a freshly truncated context is
reloaded, and the cycle repeats until the iteration cap is reached, so the
module is never completed. This single trajectory exhibits both pathologies that
characterize the model's failures: unbounded reasoning that exhausts the token
budget without progress, and context-window saturation that degrades coherence.
Across the full run, \textsc{Kimi} hits the generation cap on $219$ calls
(mean completion $6{,}002$ tokens), which both inflates wall-clock cost and
drives the high rate of unfinished modules.

\paragraph{Why the other models avoid it.}
The remaining models rarely fall into this trap: even when their reconstructions
are wrong, they almost always terminate the loop with a runnable artifact, so
their failures are \emph{accuracy} failures rather than non-completion. Two
factors separate them from Kimi-k2.5 and GLM-5. First, they are better calibrated
about when to stop: instead of spending whole turns refining a single stuck
sub-goal (e.g., one calibration scalar), they commit to writing the output entry
point early, recognize diminishing returns, and follow the ReAct protocol by
reliably issuing \texttt{WRITE\_FILE}/\texttt{RUN} actions rather than stalling on
meta-reasoning about the harness. Second, their per-turn completions stay well
below the $32{,}768$-token generation cap, so they seldom overflow the context
window or trigger the history truncation that erodes Kimi's and GLM's coherence.
The pathology is thus specific to weaker-reasoning models left to self-direct a
free-form loop, rather than a property of the task or the iteration budget; this
is also why the structured pipeline below removes it for every model.

\paragraph{Why the end-to-end pipeline is immune.}
Our end-to-end multi-agent pipeline removes this failure mode by construction.
Rather than leaving each iteration as a free-form action chosen by the model, it
drives a fixed sequence of specialized agents
($\textsc{Planner}\rightarrow\textsc{Architect}\rightarrow\textsc{Coder}
\rightarrow\textsc{Execute}\rightarrow\textsc{Judge}$): the \textsc{Architect}
first emits a validated code skeleton (a fixed set of function signatures), and
the \textsc{Coder} fills \emph{every} function in it, including the entry point
that writes the output, so a syntactically complete \texttt{solver.py} is
assembled on the very first iteration. The required output contract (the exact
array key, shape, and \texttt{np.savez} call) is supplied to the agents in
advance, so no iterations are wasted discovering the expected output format.
Each subsequent iteration then executes the assembled solver and, on failure,
receives a targeted repair ticket from the \textsc{Judge}. The budget thus acts
as a bounded \emph{error-correction} loop rather than an exploration allowance:
a complete, runnable artifact is guaranteed by the end of the first cycle, and
exhausting the budget means an unrepaired defect remains, not that nothing was
produced. Reconstruction \emph{accuracy} remains an orthogonal axis governed by
the per-task evaluation thresholds.




\newpage

\end{document}